\documentclass{article}
\usepackage[table]{xcolor}

    \PassOptionsToPackage{numbers, sort&compress}{natbib}

\usepackage[final]{neurips_2022}

\usepackage[utf8]{inputenc} 
\usepackage[T1]{fontenc}    
\usepackage{booktabs}       
\usepackage{amsfonts}       
\usepackage{nicefrac}       
\usepackage{microtype}      
\usepackage{xcolor}         
\usepackage{amsmath}
\usepackage{wrapfig}

\usepackage{graphicx}
\usepackage{subfigure}
\usepackage{titlesec} 
\usepackage{booktabs}
\usepackage[font=footnotesize]{caption}
\usepackage[ruled, vlined]{algorithm2e}
\usepackage[
bookmarksopen,
bookmarksdepth=2,
breaklinks=true,
colorlinks=true,
urlcolor=blue]{hyperref}
\newcommand{\yxw}[1]{\textcolor{blue}{[{\bf yxw:} #1]}}

\title{CEIP: Combining Explicit and Implicit Priors for Reinforcement Learning with Demonstrations}

\author{%
  Kai Yan \qquad Alexander G. Schwing \qquad Yu-Xiong Wang\\
  University of Illinois Urbana-Champaign\\
  \texttt{\{kaiyan3, aschwing, yxw\}@illinois.edu} \\
  \url{https://github.com/289371298/CEIP}
}

\titleformat{\subsubsection}[runin]
  {\normalfont\normalsize\bfseries}{\thesubsubsection}{1em}{}

\begin{document}

\maketitle

\begin{abstract}
Although reinforcement learning has found widespread use in dense reward settings, training autonomous agents with sparse rewards remains challenging. To address this difficulty, prior work has shown promising results when using not only task-specific demonstrations but also task-agnostic albeit somewhat related demonstrations. In most cases, the available demonstrations are distilled into an implicit prior, commonly represented via a single deep net. Explicit priors in the form of a database that can be queried have also been shown to lead to encouraging results. To better benefit from available demonstrations, we develop a method to Combine Explicit and Implicit Priors (CEIP). CEIP exploits multiple implicit priors in the form of normalizing flows in parallel to form a single complex prior. Moreover, CEIP uses an effective explicit retrieval and push-forward mechanism to condition the implicit priors. In three challenging environments, we find the proposed CEIP method to improve upon sophisticated state-of-the-art techniques.
\end{abstract}

\section{Introduction}
\label{sec:intro}
Reinforcement learning (RL) has found widespread use across domains from robotics~\cite{xiali2020relmogen} and game AI~\cite{Silver2017MasteringCA} to recommender systems~\cite{chen2019generative}. Despite its success, reinforcement learning is also known to be sample inefficient. For instance, training a robot arm with sparse rewards to sort objects from scratch still requires many training steps if it is at all feasible~\cite{Singh2021ParrotDB}. 


To increase the sample efficiency of reinforcement learning, prior work aims to leverage demonstrations~\cite{Brys2015RLfD, pertsch2021skild, Rengarajan2022LOGO}. These demonstrations can be {\em task-specific}~\cite{Hester2018DQLfD, Brys2015RLfD}, i.e., they directly correspond to and address the task of interest. More recently, the use of {\em task-agnostic} demonstrations has also been studied~\cite{pertsch2021skild, Hakhamaneshi2022FIST, Singh2021ParrotDB, Gupta2019RelayPL}, showing that demonstrations for loosely related tasks can enhance sample efficiency of reinforcement learning agents.

To benefit from either of these two types of demonstrations, most work distills the information within the demonstrations into an {\em implicit prior}, by encoding available demonstrations in a deep net. For example, 
SKiLD~\cite{pertsch2021skild} and FIST~\cite{Hakhamaneshi2022FIST} use a variational auto-encoder (VAE) to encode the ``skills,'' i.e., action sequences, in a latent space, and train a prior conditioned on states based on demonstrations to use the skills. Differently, PARROT~\cite{Singh2021ParrotDB} adopts a state-conditional normalizing flow to encode a transformation from a latent space to the actual action space. However, the idea of using the available demonstrations as an \textit{explicit prior} has not received a lot of attention. Explicit priors enable the agent to maintain a database of demonstrations, which can be used to retrieve state-action sequences given an agent's current state. This technique has been utilized in robotics~\cite{Chaplot2020SLAM, pari2021surprising} and early attempts of reinforcement learning with demonstrations~\cite{Brys2015RLfD}. It was also implemented as a baseline in~\cite{Gupta2019RelayPL}. One notable recent exception is FIST~\cite{Hakhamaneshi2022FIST}, which queries a database of demonstrations using the current state to retrieve a likely next state. The use of an explicit prior was shown to greatly enhance the performance. However, FIST uses pure imitation learning without any RL, hence losing the chance for trial and remedy if the imitation is not good enough.

Our key insight is to leverage demonstrations both explicitly \emph{and} implicitly, thus benefiting from both worlds. To achieve this, we develop \textbf{CEIP}, a method which \textbf{c}ombines \textbf{e}xplicit and \textbf{i}mplicit \textbf{p}riors. \textbf{CEIP} leverages implicit demonstrations by learning a transformation from a latent space to the real action space via normalizing flows. More importantly, different from prior work, such as PARROT and FIST which combine all the information within a single deep net, \textbf{CEIP} selects the most useful prior by combining multiple flows \textit{in parallel} to form a single large flow. To benefit from demonstrations explicitly, \textbf{CEIP} augments the input of the normalizing flow with a likely future state, which is retrieved via a  lookup from a database of transitions. For an effective retrieval, we propose a push-forward technique which ensures the database to return future states that have not been referred to yet, encouraging the agent to complete the whole trajectory even if it fails on a single task.


We evaluate the proposed approach on three challenging environments: fetchreach~\cite{Plappert2018MultiGoalRL}, kitchen~\cite{fu2020d4rl}, and office~\cite{Singh2020cog}. In each environment, we study the use of both task-specific and task-agnostic demonstrations. We observe that integrating an explicit prior, especially with our proposed push-forward technique,  greatly improves results. Notably, the proposed approach works well on sophisticated long-horizon robotics tasks with a few, or sometimes even one task-specific demonstration. 

\section{Preliminaries}
\label{sec:pre}

\textbf{Reinforcement Learning.}  Reinforcement learning (RL) aims to train an \textit{agent} to make the `best' decision towards completing a particular task in a given environment. The environment and the task are often described as a Markov Decision Process (MDP), which is defined by a tuple $(\mathcal{S}, \mathcal{A}, T, r, \gamma)$. In timestep $t$ of the Markov process, the agent observes the current \textit{state} $s_t\in\mathcal{S}$, and executes an \textit{action} $a_t\in\mathcal{A}$ following some probability distribution, i.e., \textit{policy} $\pi(a_t|s_t)\in\Delta(\mathcal{A})$, where $\Delta(\mathcal{A})$ denotes the probability simplex over elements in space $\mathcal{A}$. Upon executing action $a_t$, the state of the agent changes to $s_{t+1}$ following the dynamics of the environment, which are governed by the \textit{transition function}   $T(s_t,a_t):\mathcal{S}\times\mathcal{A}\rightarrow \Delta(\mathcal{S})$. Meanwhile, the agent receives a \textit{reward} $r(s_t, a_t)\in\mathbb{R}$. 
The agent aims to maximize the cumulative reward $\sum_t \gamma^tr(s_t,a_t)$, where $\gamma\in[0, 1]$ is the discount factor. One complete run in an environment is called an \textit{episode}, and the corresponding state-action pairs $\tau = \{(s_1, a_1), (s_2, a_2), \dots\}$ form a \textit{trajectory} $\tau$.

\textbf{Normalizing Flows.} A normalizing flow~\cite{Kobyzev2021NFreview} is a generative model that transforms elements $z_0$ drawn from a simple distribution $p_z$, e.g., a Gaussian, to elements $a_0$ drawn from a more complex distribution $p_a$. For this transformation, a bijective function $f$ is used, i.e., $a_0=f(z_0)$. The use of a bijective function ensures that the log-likelihood of the more complex distribution at any point is tractable and that samples of such a distribution can be easily generated by taking samples from the simple distribution and pushing them through the flow. Formally, the core idea of a normalizing flow can be summarized via $p_a(a_0)=p_z(f^{-1}(a_0)) \left|\frac{\partial f^{-1}(a)}{\partial a}|_{a=a_0}\right|$, where $\left|\cdot\right|$ is the determinant (guaranteed positive by flow designs),  $a$ is a random variable with the desired more complex distribution, and $z$ is a random variable governed by a simple distribution. To efficiently compute the determinant of the Jacobian matrix of $f^{-1}$, special constraints are imposed on the form of $f$. For example, coupling flows like RealNVP~\cite{Dinh17RealNVP} and autoregressive flows~\cite{Papamakarios2017MAFDE} impose the Jacobian of $f^{-1}$ to be triangular.

\section{CEIP: Combining Explicit and Implicit Priors}
\label{sec:method}


\begin{figure}[t]
    \centering
    \includegraphics[width=0.8\linewidth]{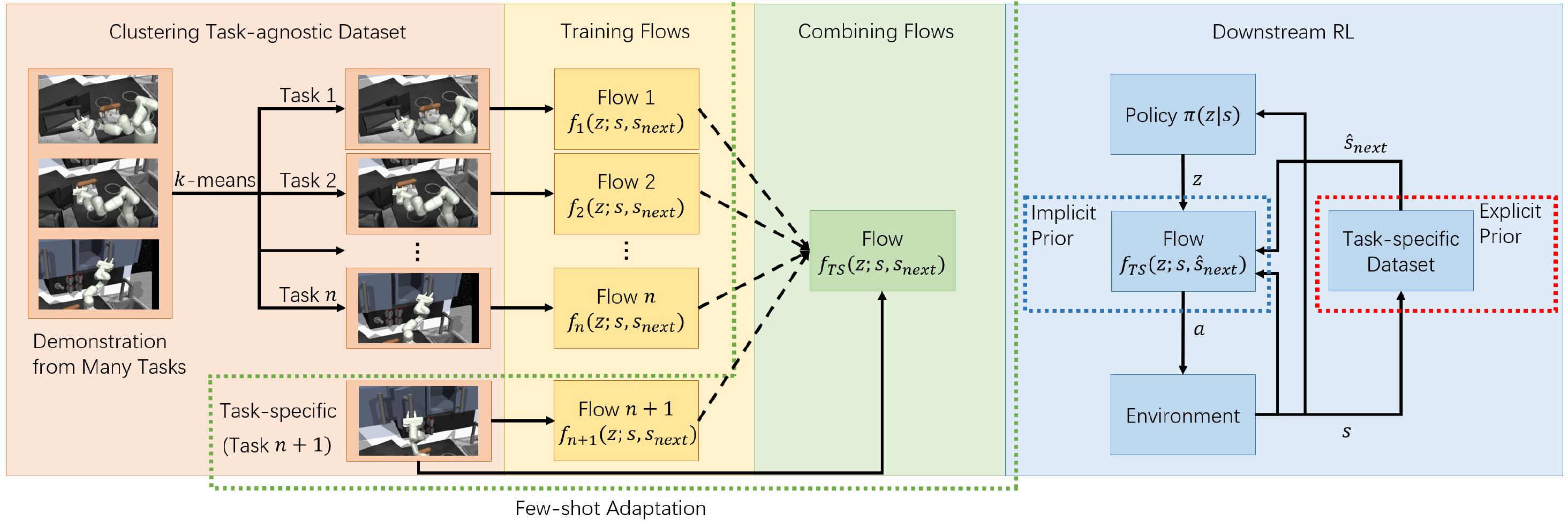}
    \caption{Overview of our proposed approach, CEIP. Our approach can be divided into three steps: a) cluster the task-agnostic dataset into different tasks, and then train one flow on each of the $n$ tasks of the task-agnostic dataset; b) train a flow on the task-specific dataset, and then train the coefficients to combine the $n+1$ flows into one large flow $f_\text{TS}$, which is the implicit prior; c) conduct reinforcement learning on the target task; for each timestep, we perform a dataset lookup in the task-specific dataset to find the state most similar to current state $s$, and return the likely \textit{next state} $\hat{s}_{\text{next}}$ in the trajectory, which is the explicit prior.}
    \label{fig:overview}
\end{figure}
\subsection{Overview}
\label{sec:overview}
As illustrated in Fig.~\ref{fig:overview}, our goal is to train an autonomous agent to solve challenging tasks despite sparse rewards, such as controlling a robot arm to complete item manipulation tasks (like turning on a switch or opening a cabinet). For this we aim to benefit from available demonstrations. Formally, we consider a task-specific dataset $D_{\text{TS}}=\{\tau^{\text{TS}}_1, \tau^{\text{TS}}_2, \dots, \tau^{\text{TS}}_m\}$, where $\tau^{\text{TS}}_i$ is the $i$-th trajectory of the task-specific dataset, and a task-agnostic dataset $D_{\text{TA}}=\{\bigcup D_i|i\in\{1,2,3,\dots,n\}\}$, where $D_i=\{\tau^i_1, \tau^i_2, \dots, \tau^i_{m_i}\}$ subsumes the demonstration trajectories for the $i$-th task in the task-agnostic dataset. Each trajectory $\tau=\{(s_1, a_1),(s_2,a_2),\dots\}$ in the dataset is a state-action pair sequence of a complete episode, where $s$ is the state, and $a$ is the action. We assume that the number of available task-specific trajectories is very small, i.e., $\sum_{i=1}^{n}m_i\gg m$, which is common in practice.  For readability, we will also refer to $D_{\text{TS}}$ using $D_{n+1}$.


Our approach leverages demonstrations implicitly by training a normalizing flow $f_\text{TS}$, which transforms the probability distribution represented by a policy $\pi(z|s)$ over a simple latent probability space $\mathcal{Z}$, i.e.,  $z\in\mathcal{Z}$, into a reasonable expert policy over the space of real-world actions $\mathcal{A}$. As before, $s$ is the current environment state. Thus, the downstream RL agent only needs to learn a policy $\pi(z|s)$ that results in a probability distribution over latent space $\mathcal{Z}$, which is subsequently mapped via the flow $f_\text{TS}$ to a real-world action $a\in\mathcal{A}$. Intuitively, the MDP in the latent space is governed by a less complex probability distribution, making it easier to train because the flow increases the exposure of more likely actions, while reducing the chance that a less-likely action is chosen. This is because the flow reduces the probability mass for less likely actions given the current state.

Task-agnostic demonstrations contain useful patterns that may be related to the task at hand. 
However, not all the task-agnostic data are always equally useful, as different task-agnostic data may require to expose different parts of the action space. Therefore, different from prior work where all data are fed into the same deep net model, we first partition the task-agnostic dataset into different groups according to task similarity so as to increase flexibility.  For this we use a classical $k$-means algorithm. We then train different flows $f_i$ on each of the groups, and finally combine the flows via learned coefficients into a single flow $f_\text{TS}$. Beneficially, this process permits to expose different parts of the action space as needed and according to perceived task similarity. 

Lastly, our approach further leverages demonstrations explicitly, by conditioning the flow not only on the current state but also on a likely next state, to better inform the agent of the state it should try to achieve with its current action. In the following, we first discuss the implicit prior of \textbf{CEIP} in Sec.~\ref{sec:implicitprior}; afterward we discuss our explicit prior in Sec.~\ref{sec:explicitprior}, and the downstream reinforcement learning with both priors in Sec.~\ref{sec:RL}.

\subsection{Implicit Prior}
\label{sec:implicitprior}

To better benefit from demonstrations implicitly, we use a 1-layer normalizing flow as the backbone of our implicit prior. It essentially corresponds to a conditioned affine transformation of a Gaussian distribution. We choose a flow-based model instead of a VAE-based one for two reasons: 1) as the dimensionality before and after the transformation via a normalizing flow remains identical and since the flow is invertible, the agent is guaranteed to have control over the whole action space. This ensures that all parts of the action space are accessible, which is not guaranteed by VAE-based methods like SKiLD or FIST; 2) normalizing flows, especially coupling flows such as RealNVP~\cite{Dinh17RealNVP}, can be easily stacked \textit{horizontally}, so that the combination of parallel flows is also a flow. Among feasible flow models, we found that the simplest 1-layer flow  suffices to achieve good results, and is even more robust in training than a more complex RealNVP. 
Next, in Sec.~\ref{sec:nf} we first introduce details regarding the normalizing flow $f_i$, before we discuss in Sec.~\ref{sec:adapt} how to combine the flows into one flow $f_\text{TS}$ applicable to the task for which  the task-specific dataset contains demonstrations.

\subsubsection{Normalizing Flow Prior.}
\label{sec:nf}


For each task $i$ in the task-agnostic dataset, i.e., for each $D_i$, we train a conditional 1-layer normalizing flow $f_i(z; u)=a$ which maps a latent space variable $z\in\mathbb{R}^q$ to an action $a\in\mathbb{R}^{q}$, where $q$ is the number of dimensions of the real-valued action vector. We let $u$ refer to a conditioning variable. In our case $u$ is either the current environment state $s$ (if no explicit prior is used) or a concatenation of the current and a likely next state $[s, s_{\text{next}}]$ (if an explicit prior is used). Concretely, the formulation of our 1-layer flow is
\begin{equation}
f_i(z; u)=a=\exp\{c_i(u)\}\odot z + d_i(u),
\end{equation}
where $c_i(u)\in\mathbb{R}^q$, $d_i(u)\in\mathbb{R}^q$ are trainable deep nets, and $\odot$ refers to the Hadamard product. The $\exp$ function is applied elementwise. When training the flow, we sample state-action pairs (without explicit prior) or transitions (with explicit prior) $(u, a)$ from the dataset $D_i$, and maximize the log-likelihood $\mathbb{E}_{(u,a)\sim D_i}\log p(a|u)$; refer to \cite{Kobyzev2021NFreview} for how to maximize this objective.


In the discussion above, we assume the decomposition of the task-agnostic dataset into tasks to be given. If such a decomposition is not provided (e.g., for the kitchen and office environments in our experiments), we perform a  $k$-means clustering to divide the task-agnostic dataset into different parts. The clustering algorithm operates on the last state of a trajectory, which is used to represent the whole trajectory. The intuition is two-fold. First, for many real-world MDPs, achieving a particular terminal state is more important than the actions taken~\cite{Seyed2019Divergence}. For example, when we control a robot to pick and place items, we want all target items to reach the right place eventually; 
however, we do not care too much about the actions taken to achieve this state. Second, among all the states, the final state is often the most informative about the task that the agent has completed. 
The number of clusters $k$ in the $k$-means algorithm is a hyperparameter, which empirically should be larger than the number of dimensions of the action space. 
Though we assume the task-agnostic dataset is partitioned into labeled clusters, our experiments show that our approach is robust and good results are achieved even without a precise ground-truth decomposition.

In addition to the clusters in the task-agnostic dataset, we train a flow $f_{n+1}(z; u)=a$ on the task-specific dataset $D_{n+1}=D_{\text{TS}}$, using the same maximum log-likelihood loss, which is optional but always available. This is not necessary when the task is relatively simple and the episodes are short (e.g., the fetchreach environment in the experiment section), but becomes particularly helpful in scenarios where some subtasks of a task sequence only appear in the task-specific dataset (e.g., the kitchen environment). 



\begin{figure}[t]
    \centering
    \includegraphics[width=0.8\linewidth]{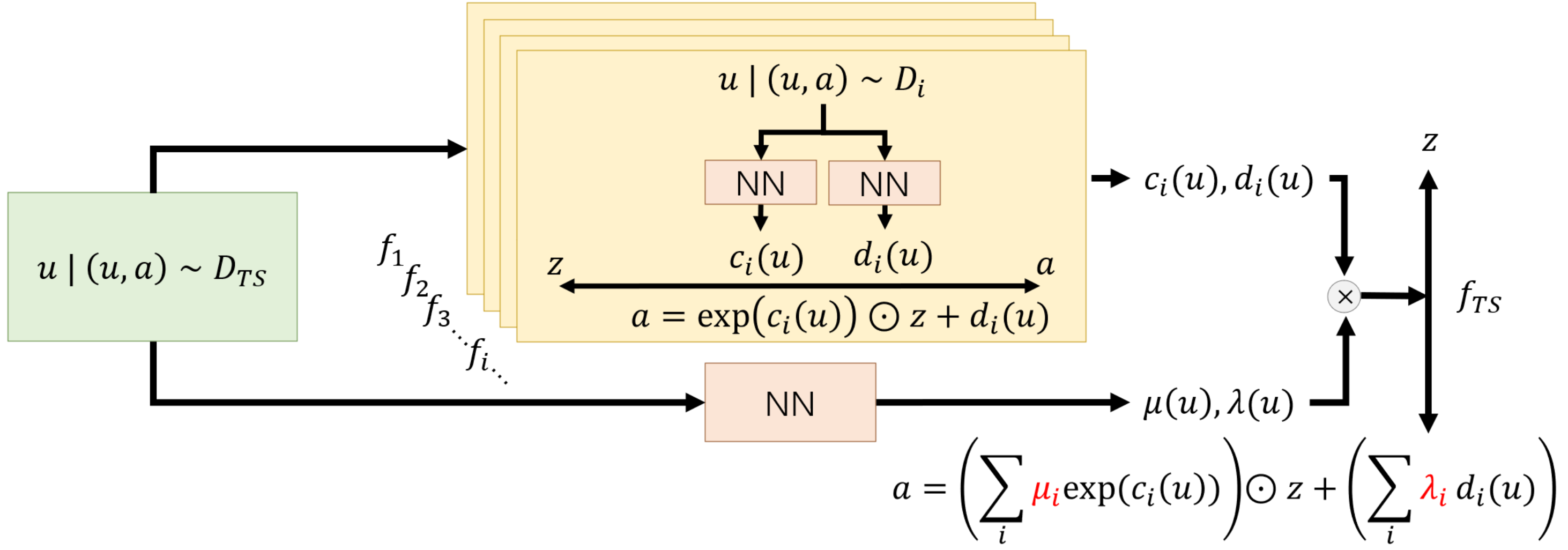}
    \caption{An illustration of how we combine different flows into one large flow for the task-specific dataset. Each red block of ``NN'' stands for a neural network. Note that $c_i(u)$ and $d_i(u)$ are vectors, while $\mu_i$ and $\lambda_i$ are the $i$-th dimension of $\mu(u)$ and $\lambda(u)$.}
    \label{fig:flowcombine}
\end{figure}

\subsubsection{Few-shot Adaptation.}
\label{sec:adapt}


The flow models discussed in Sec.~\ref{sec:nf} learn which parts of the action space to be more strongly exposed from the latent space. However, not all the flows expose useful parts of the action space for the current state. 
For example, the target task needs the agent to move its gripper upwards at a particular location, but in the task-agnostic dataset, the robot more often moves the gripper downwards to finish another task. In order to select the most useful prior, we need to tune our set of flows learned on the task-agnostic datasets to the small number of trajectories available in the task-specific dataset. To ensure that this does not lead to overfitting as only a very small number of task-specific trajectories are available, we train a set of coefficients that selects the flow that works the best for the current task. 
Concretely, given all the trained flows, we  train a set of coefficients to combine the flows $f_1$ to $f_n$ trained on the task-agnostic data, and also the flow $f_{n+1}$ trained on the task-specific data. The coefficients select from the set of available flows the most useful one. To achieve this, we use the combination flow illustrated in Fig.~\ref{fig:flowcombine} which is formally specified as follows: 
\begin{equation}
f_\text{TS}(z;u)=\left(\sum_{i=1}^{n+1}\mu_i(u)\exp\{c_i(u)\}\right)\odot z+\left(\sum_{i=1}^{n+1}\lambda_i(u)d_i(u)\right).
\end{equation}
Here, $\mu_i(u)\in\mathbb{R}$, $\lambda_i(u)\in\mathbb{R}$ are the $i$-th entry of the deep nets $\mu(u)\in\mathbb{R}^{n+1}$, $\lambda(u)\in\mathbb{R}^{n+1}$, respectively, which yield the coefficients while the deep nets $c_i$ and $d_i$ are frozen. As before, the $\exp$ function is applied elementwise. We use a softplus activation and an offset at the output of $\mu$ to force $\mu_i(u)\geq 10^{-4}$ for any $i$ for numerical stability. Note that the combined flow $f_{\text{TS}}$ consisting of multiple 1-layer flows  is also a 1-layer normalizing flow. Hence,  all the compelling properties over VAE-based architectures described at the beginning of Sec.~\ref{sec:implicitprior} remain valid. To train the combined flow, we use the same log likelihood loss $\mathbb{E}_{(u,a)\sim D_{\text{TS}}}\log p(a|u)$ as that for training single flows. Here, we optimize the deep nets $\mu(u)$ and $\lambda(u)$ which parameterize $f_{\text{TS}}$.


Obviously, the employed combination of flows can be straightforwardly extended to a more complicated flow, e.g., a RealNVP~\cite{Dinh17RealNVP} or Glow~\cite{Kingma2018GlowGF}. 
However, we found the discussed simple formulation to  work remarkably well and to be robust. 







\subsection{Explicit Prior}
\label{sec:explicitprior}






Beyond distilling information from demonstrations into deep nets which are then used as implicit priors, we find explicit use of demonstrations to also be remarkably useful. 
To benefit, we encode future state information into the input of the flow. More specifically, instead of sampling $(s,a)$-pairs from a dataset $D$ for training the flows, we consider sampling a \textit{transition} $(s, a, s_{\text{next}})$ from $D$. During training, we concatenate $s$ and $s_{\text{next}}$ before feeding it into a flow, i.e., $u=[s, s_{\text{next}}]$ instead of $u=s$. 

However, we do not know the future state $s_{\text{next}}$ when deploying the policy. To obtain an estimate, we use task-specific demonstrations as explicit priors. 
More formally, we use the trajectories within the task-specific dataset $D_{\text{TS}}$ as a database. This is manageable as we assume the task-specific dataset to be small. For each environment step of reinforcement learning with current state $s$, we perform a lookup, where $s$ is the query,  states $s_{\text{key}}$ in the trajectories are the keys, and their corresponding next state $s_{\text{next}}$  is the value. Concretely, we assume $s_{\text{next}}$ belongs to trajectory $\tau$ in the task-specific dataset $D_{\text{TS}}$, and define $\hat{s}_ {\text{next}}$ as the result of the database retrieval with respect to the given query $s$, i.e., 
\begin{equation}
\begin{aligned}
    \hat{s}_{\text{next}}&= \text{argmin}_{s_{\text{next}}|(s_{\text{key}}, a, s_{\text{next}})\in D_{\text{TS}}} [(s_{\text{key}}-s)^2+C\cdot \delta (s_{\text{next}})], \text{where}\\
    \delta(s_{\text{next}})&=\begin{cases}1\ \text{if }\exists s'_{\text{next}}\in\tau, \text{ s.t. } s'_{\text{next}}\text{ is no earlier than $s_{\text{next}}$ in $\tau$ and has been retrieved},\\0\ \text{otherwise}.\end{cases}
\label{eq:ex}
\end{aligned}
\end{equation}
In Eq.~\eqref{eq:ex}, $C$ is a constant and $\delta$ is the indicator function. We  set $u=[s, \hat{s}_{\text{next}}]$ as the condition, feed it into the trained  flow $f_{\text{TS}}$, and map the latent space element $z$ obtained from the RL policy to the real-world action $a$. The penalty term $\delta$ is a push-forward technique, which aims to push the agent to move forward 
instead of staying put, imposing monotonicity on the retrieved $\hat{s}_{\text{next}}$. Consider an agent at a particular state $s$ and a flow $f_{\text{TS}}$, conditioned on $u=[s, \hat{s}_{\text{next}}]$ which maps the chosen action $z$ to a real-world action $a$ that does not modify the environment. Without the penalty term, the agent will remain at the same state, retrieve the same likely next state, which again maps onto the action that does not change the environment. Intuitively, this term discourages 1) retrieving the same state twice, and 2) returning to earlier states in a given trajectory. In our experiments, we  set $C=1$.  
 
 
\subsection{Reinforcement Learning with Priors}
 \label{sec:RL}
Given the implicit  and explicit priors, we use RL  to train a policy $\pi(z|s)$ to accomplish the target task demonstrated in the task-specific dataset. As shown in Fig.~\ref{fig:overview}, the RL agent receives a state $s$ and provides a latent space element $z$. The conditioning variable of the flow is retrieved via the dataset lookup described in Sec.~\ref{sec:explicitprior} and the real-world action $a$ is then computed using the flow. 
Note, our approach is suitable for any RL method, i.e., the policy $\pi(z|s)$ can be trained using any RL algorithm such as  proximal policy optimization (PPO)~\cite{Schulman2017PPO} or soft-actor-critic (SAC)~\cite{Haarnoja2018SoftAO}.





















\label{others}

\section{Experiments}
In this section, we evaluate our CEIP approach on three challenging environments: fetchreach (Sec.~\ref{sec:fetchreach}), kitchen (Sec.~\ref{sec:kitchen}), and office (Sec.~\ref{sec:office}), which are all tasks that manipulate a robot arm. In each experiment, we study the following questions: 1) Can the algorithm make good use of the demonstrations compared to baselines? 
2) Are our core design decisions (e.g., state augmentation with explicit prior and the push-forward technique) indeed helpful? 




\textbf{Baselines.} We compare the proposed method to three baselines: PARROT~\cite{Singh2021ParrotDB}, SKiLD~\cite{pertsch2021skild}, and FIST~\cite{Hakhamaneshi2022FIST}. In all environments, we use reward as our criteria (higher is better). The results are averaged over $3$ runs for SKiLD (much slower to train) and $9$ runs for all other methods unless otherwise mentioned. To differentiate variants of the PARROT baseline and our method, we use suffixes. We use ``EX'' to refer to variants with explicit prior, and ``forward'' for variants with the push-forward technique. For our method, if we train a task-specific flow on $D_{\text{TS}}=D_{n+1}$, we append the abbreviation ``TS.'' For PARROT, the use of the task-specific data is indicated with ``TS'' and the use of  task-agnostic data is indicated with ``TA.''\footnote{The original PARROT in~\cite{Singh2021ParrotDB} is essentially PARROT+TA. It is straightforward to use PARROT directly on the task-specific dataset. Hence, we tried PARROT+TS and PARROT+(TS+TA) as well.} See Table~\ref{tab:abbrCEIP} and Table~\ref{tab:abbrPARROT} for precise correspondence.


\subsection{FetchReach Environment}
\label{sec:fetchreach}


\textbf{Environment Setup.} The agent needs to control a robot arm to move its gripper to a goal location in 3D space, and remain there. 
During an episode of $40$ steps, the agent receives a $10$-dimensional state about its location and outputs a $4$-dimensional action, which indicates the change of coordinates of the agent and the openness of the gripper. It will receive a reward of $0$ if it arrives and stays in the vicinity of its target. Otherwise, it will receive a reward of $-1$. This environment is a harder version of the FetchReach-v1 robotics environment in gym~\cite{Plappert2018MultiGoalRL}, where we increase the average distance of the starting point to the goal, effectively increasing the training difficulty. Moreover, to test the robustness of the algorithm, we sample a random action from a normal distribution at the beginning of each episode, which the agent executes for $x$ steps before the episode begins. We use $x\sim U[5, 20]$. For simplicity, we denote the goal generated with  azimuth $\frac{\pi d}{4}$ as ``direction $d$'' (e.g., direction $4.5$).

\textbf{Dataset Setup.} We use trajectories from directions $d\in\{0, 1,  \dots, 7\}$ as the task-agnostic data. Each task includes $40$ trajectories, and each of the trajectories has $40$ steps, i.e., $1600$ environment steps in total. The task-specific datasets contain directions $4.5, 5.5, 6.5$, and $7.5$. (The robot cannot reach the other four $.5$ directions due to physical limits.) For each task-specific dataset, we use $4$ trajectories, for a total of $160$ environment steps. 

\textbf{Experimental Setup.} For fetchreach, we use a fully-connected deep net with one hidden layer of width $32$ and ReLU~\cite{Agarap2018relu} activation as a standard ``block'' of our algorithm (each block corresponds to a red ``NN'' rectangle in Fig.~\ref{fig:flowcombine}). We have a pair of blocks for $c_i(s)$ and $d_i(s)$  for each flow $f_i$. For flow training, we train $8$ flows for $8$ directions in the task-agnostic dataset without the explicit prior. 
We use a batch size of $40$ and train for $1000$ epochs for both each flow and the combination of flows, with gradient clipping  at norm $10^{-4}$, learning rate $0.001$, and Adam optimizer~\cite{KingmaB2014Adam}. We use the model that has the best performance on the validation dataset at the end of every epoch. For each dataset, we randomly draw $80\%$ state-action pairs (or transitions in ablation) as the training set and $20\%$ state-action pairs as the validation set. The combination of flow is also a block, which outputs both $\mu(s)$ and $\lambda(s)$. See the Appendix for the implementation details of SKiLD, FIST, and PARROT. For each method with RL training, we use a soft-actor-critic (SAC)~\cite{Haarnoja2018SoftAO} with 30K environment steps,  a batch size of $256$, and $1000$ steps of initial random exploration. Unless otherwise noted,  all other RL hyperparameters  in all experiments use the default values of Stable-baselines3~\cite{Raffin2021stable-baselines3}.

\begin{figure}[t]
    \centering
    \subfigure[Direction 4.5]{ 
    \begin{minipage}[b]{0.23\linewidth}
    \includegraphics[width=\linewidth]{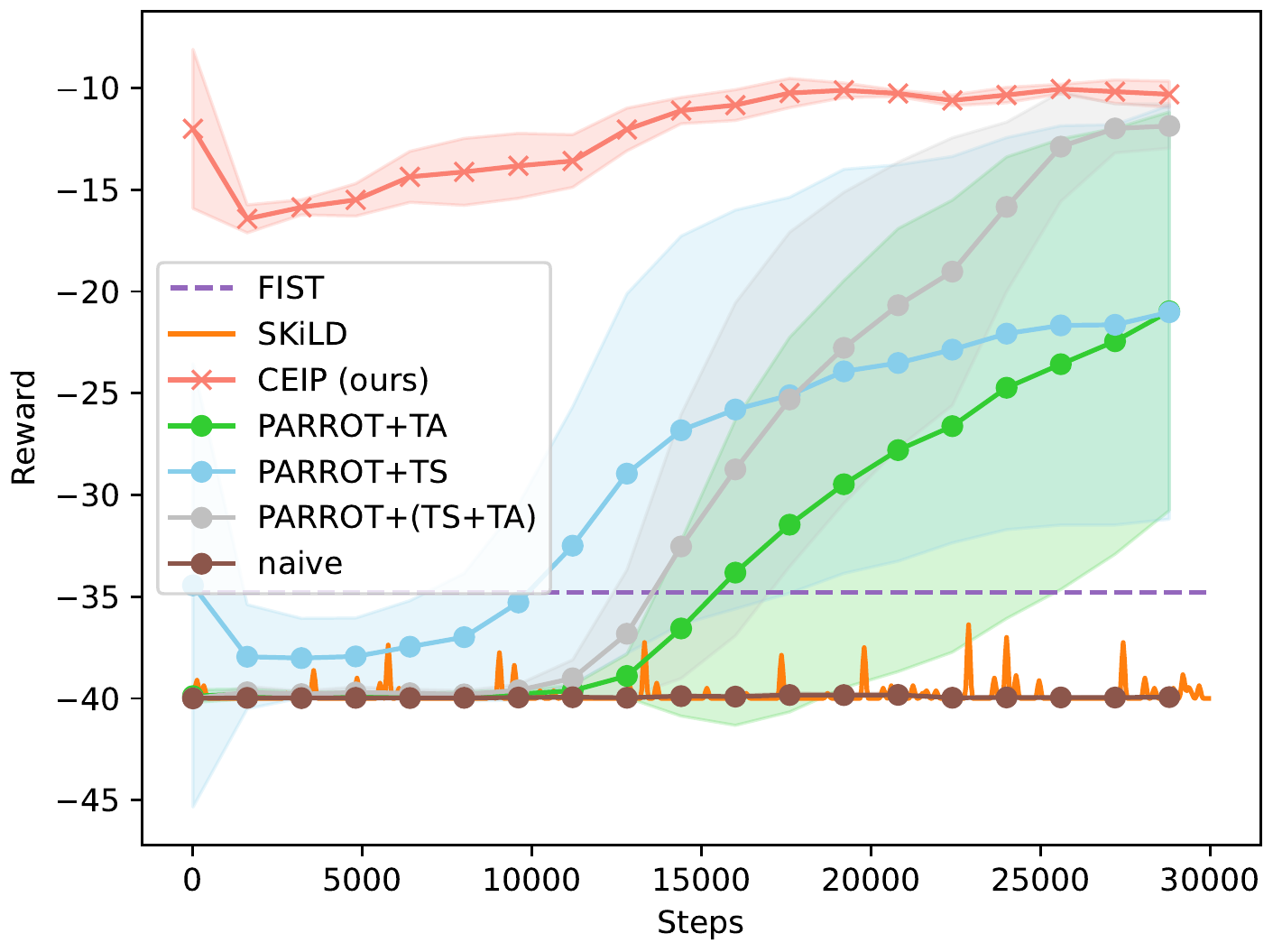}
    \end{minipage}
    }
    \subfigure[Direction 5.5]{
    \begin{minipage}[b]{0.23\linewidth}
    \includegraphics[width=\linewidth]{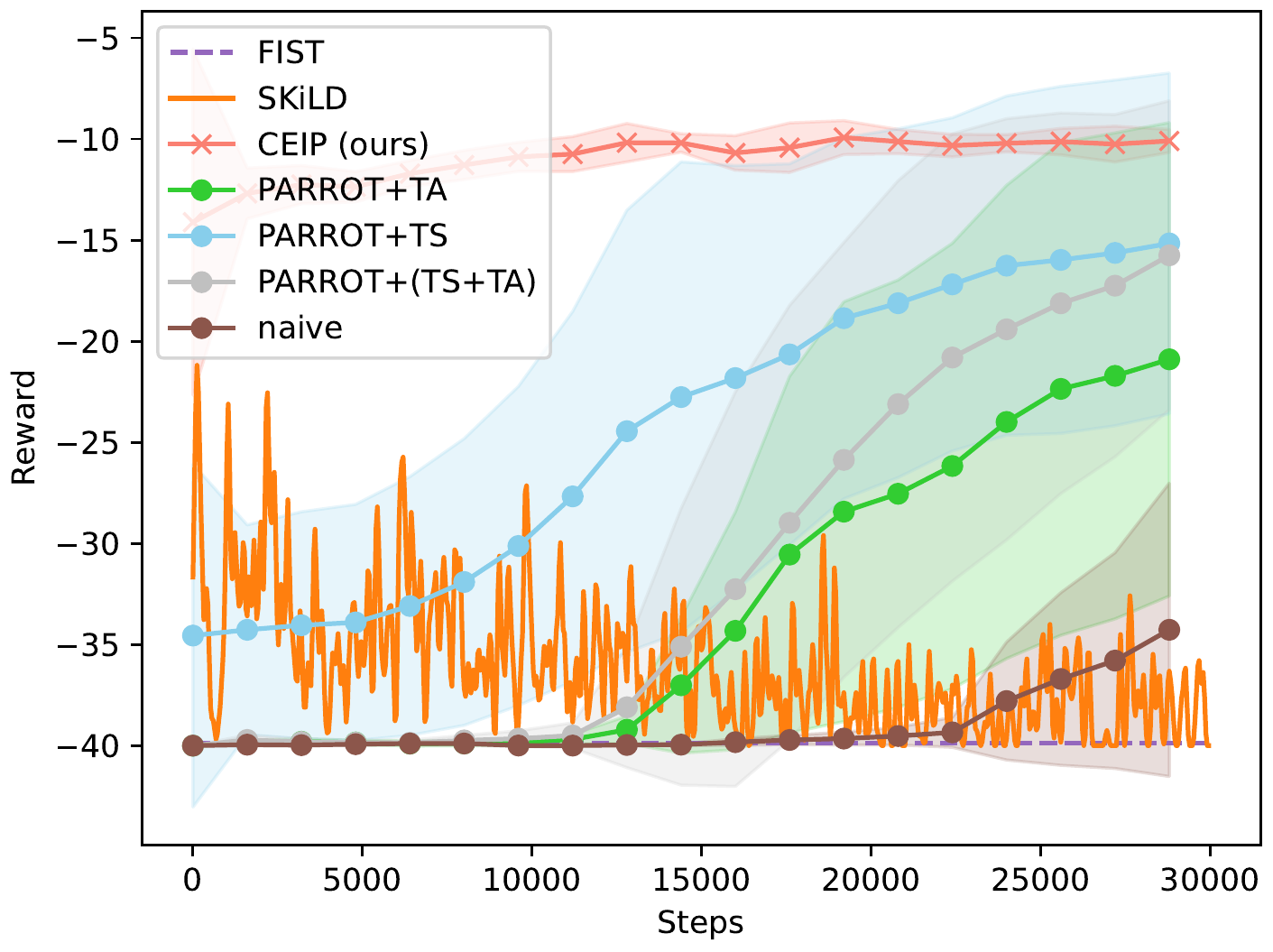}
    \end{minipage}
    }
    \subfigure[Direction 6.5]{ 
    \begin{minipage}[b]{0.23\linewidth}
    \includegraphics[width=\linewidth]{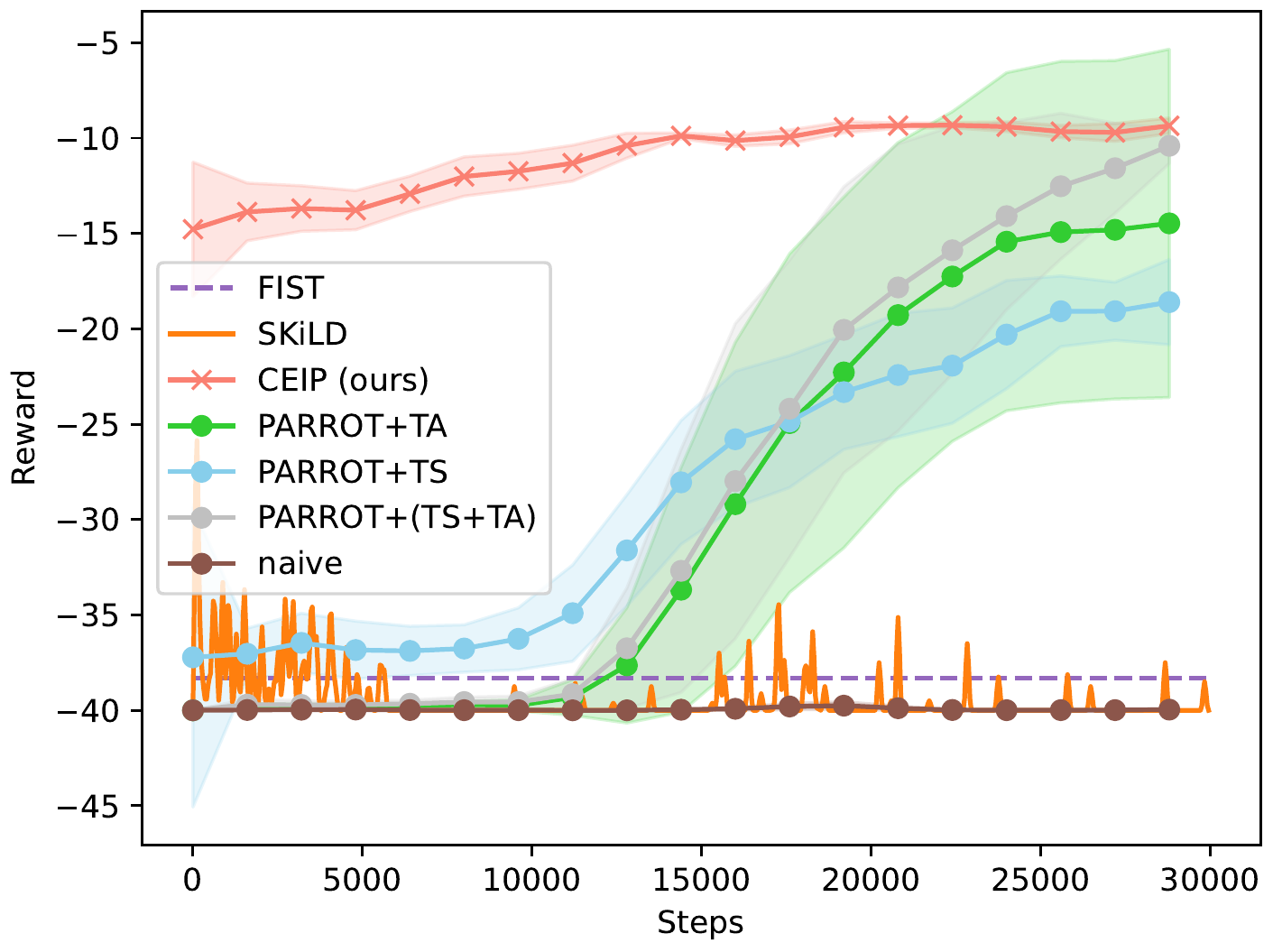}
    \end{minipage}
    }
    \subfigure[Direction 7.5]{
    \begin{minipage}[b]{0.23\linewidth}
    \includegraphics[width=\linewidth]{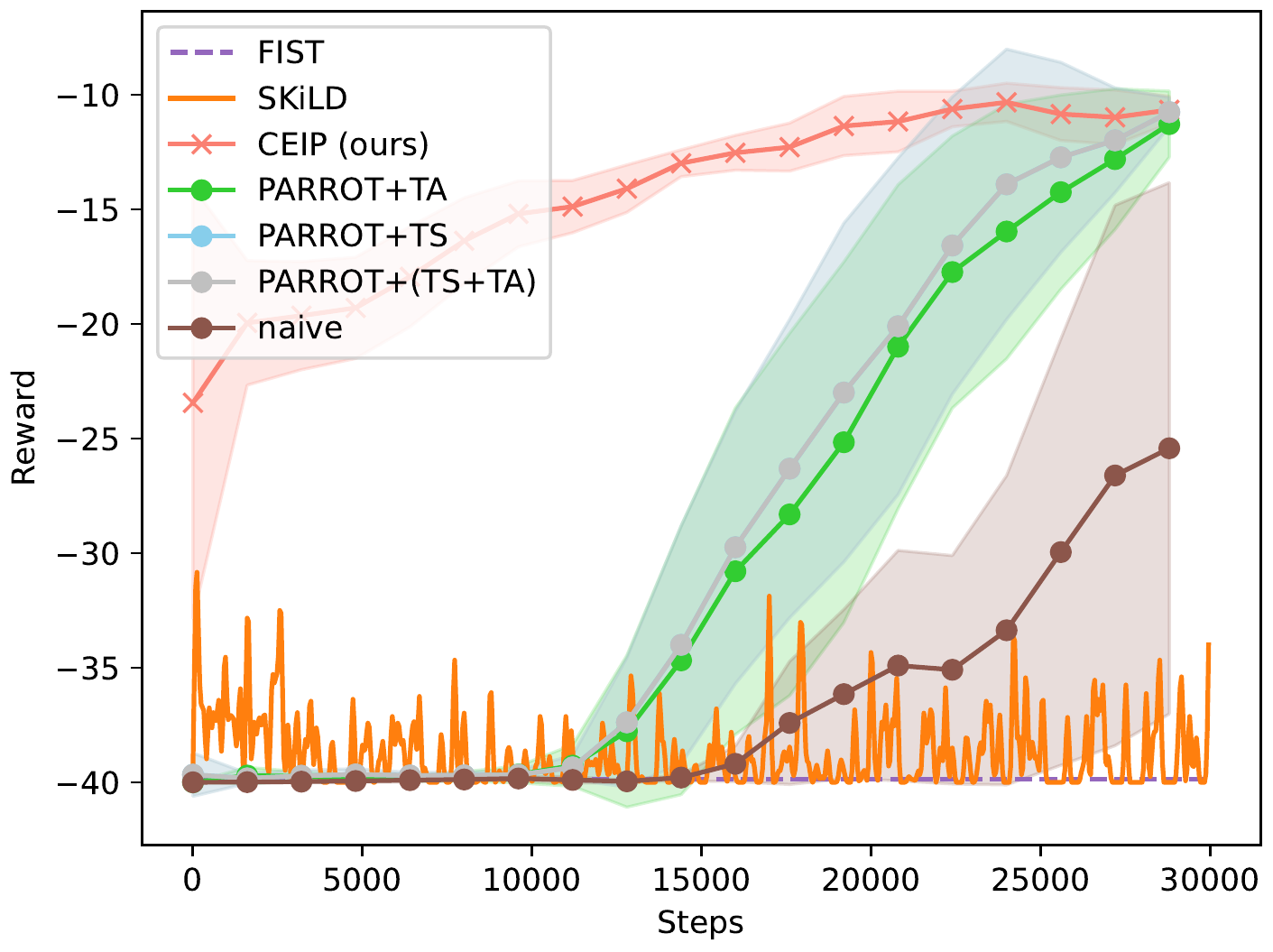}
    \end{minipage}
    }
    \vspace{-0.2cm}
    \caption{Main performance results on the fetchreach environment for different directions, where the lines are the mean reward (higher is better) and shades are the standard deviation. FIST is represented by a dashed line as it does not require RL.} 
    \label{fig:fetchreach_plot_main}
    \vspace{-0.2cm}
\end{figure}

\textbf{Main Results.}  Fig.~\ref{fig:fetchreach_plot_main} shows the results for different methods without explicit priors or task-specific single flow $f_{n+1}$. In all four tasks, our method significantly outperforms the other baselines. 
This indicates that the flow training indeed helps boost the exploration process. 
Na\"ive reinforcement learning from scratch fails in most cases, which underscores the necessity of utilizing demonstrations to aid RL exploration. As this is a simple task with only a few wildly varied trajectories, adding a flow for the task-specific dataset does not improve our method. Noteworthy, neither SKiLD nor FIST works on fetchreach. Their VAE-based architecture with each action sequence as the agent's output (``skill'') can not be trained with the little amount of wildly varied data with short horizon. 
Flow-based models like ours and PARROT, which only consider the action of the current step instead of the action sequence, work better. 

\begin{figure}[t]
    \centering
    \subfigure[Direction 4.5]{ 
    \begin{minipage}[b]{0.23\linewidth}
    \includegraphics[width=\linewidth]{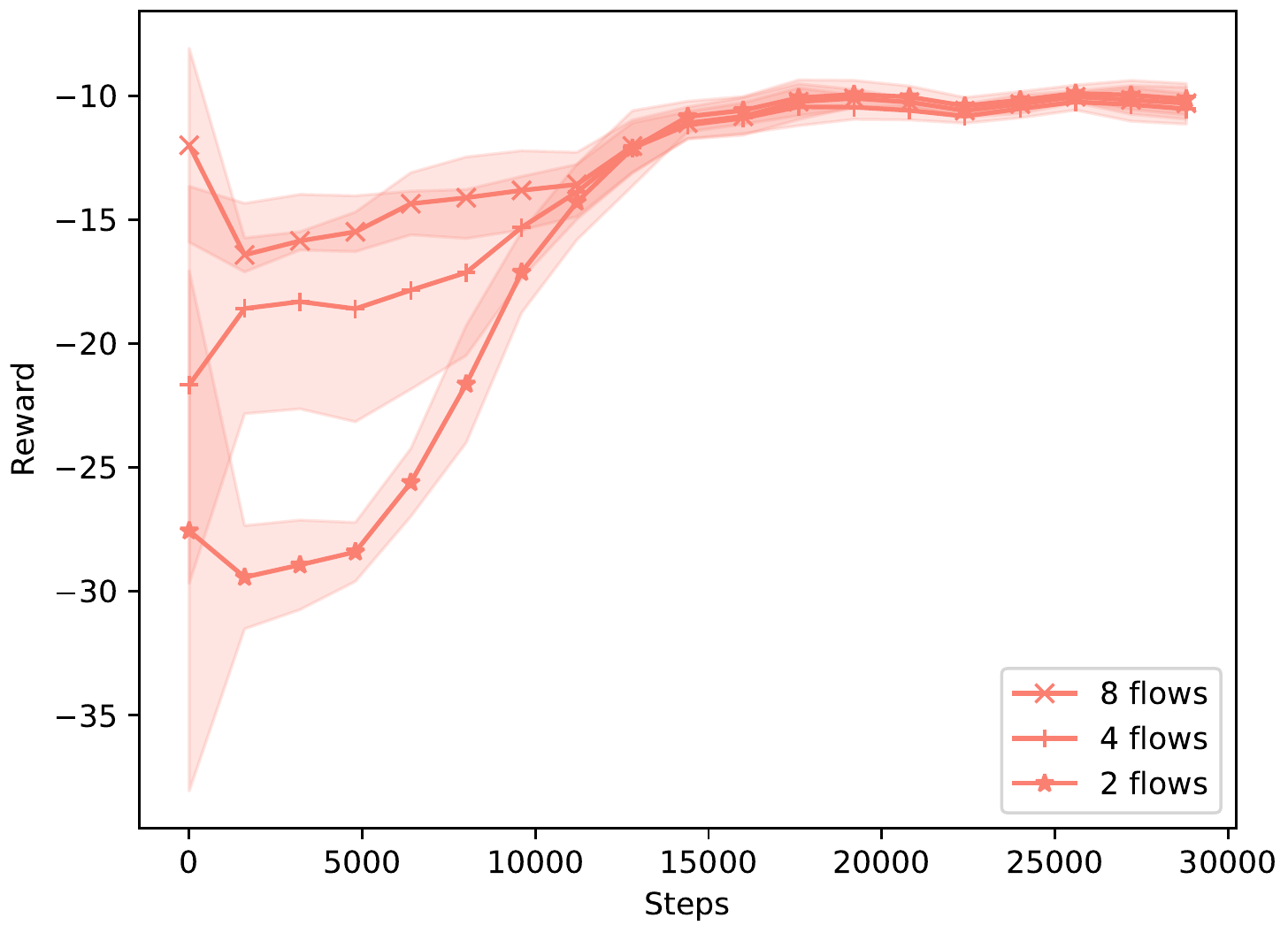}
    \end{minipage}
    }
    \subfigure[Direction 5.5]{
    \begin{minipage}[b]{0.23\linewidth}
    \includegraphics[width=\linewidth]{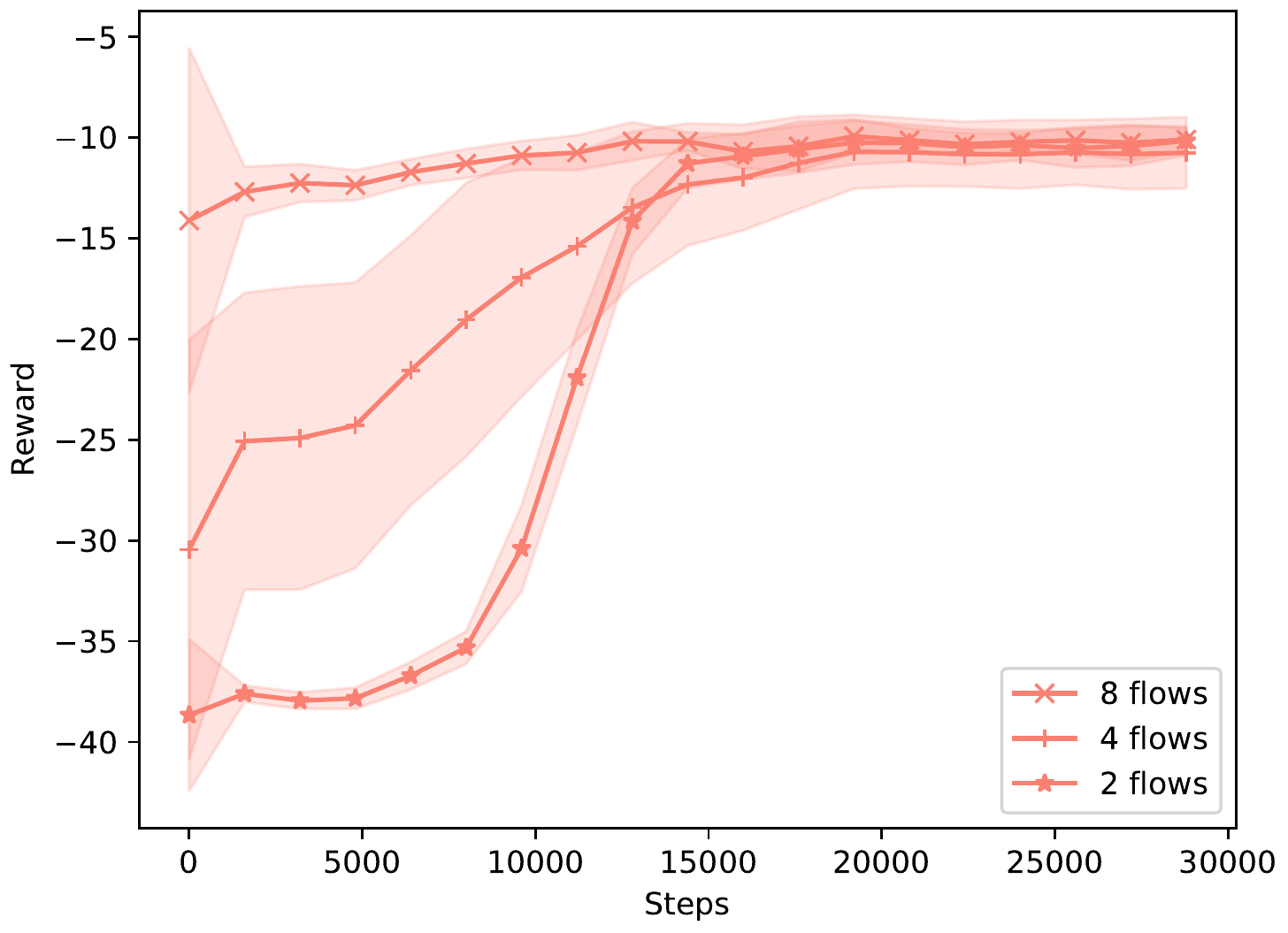}
    \end{minipage}
    }
    \subfigure[Direction 6.5]{ 
    \begin{minipage}[b]{0.23\linewidth}
    \includegraphics[width=\linewidth]{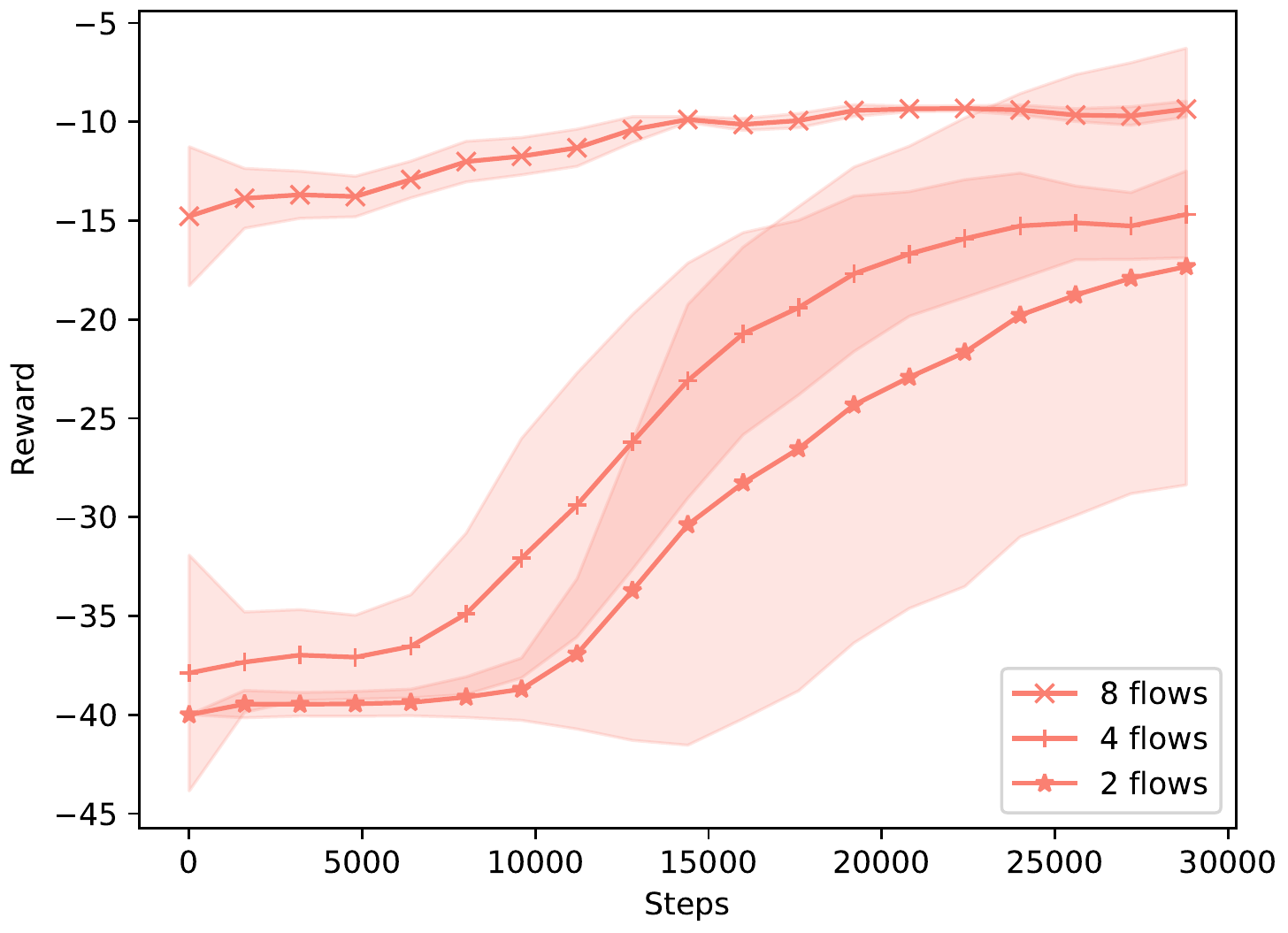}
    \end{minipage}
    }
    \subfigure[Direction 7.5]{
    \begin{minipage}[b]{0.23\linewidth}
    \includegraphics[width=\linewidth]{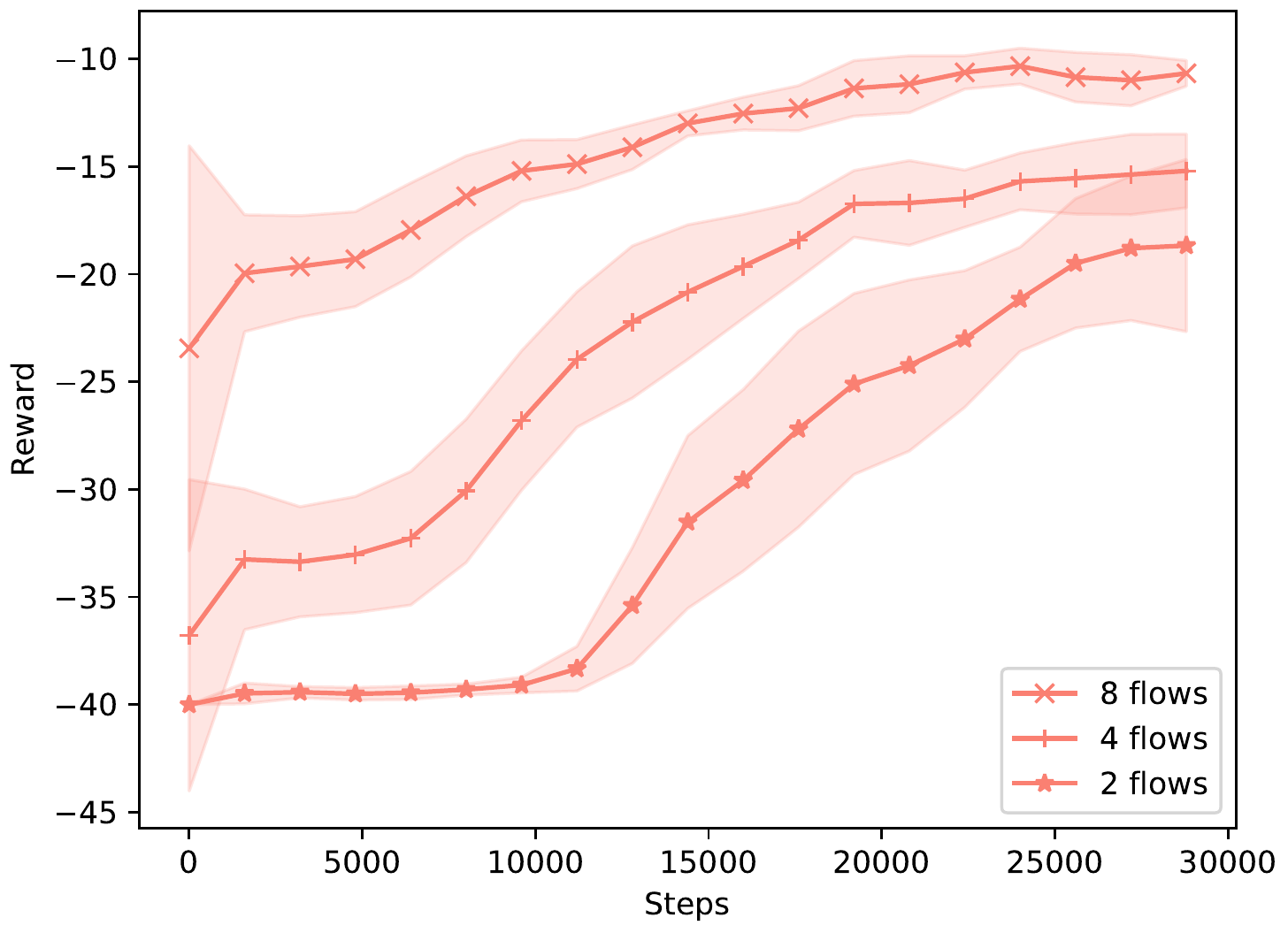}
    \end{minipage}
    }
    \caption{Ablation on the number of flows used in our method. We observe more flows to lead to better performance, likely  because  expressivity increases which helps in fitting the expert policy.}
    \label{fig:fetchreach_plot_ablation_ours_dataset}
\end{figure}

\textbf{Are more flows helpful for CEIP?} Fig.~\ref{fig:fetchreach_plot_ablation_ours_dataset} shows the performance of our method using a different number of flows, which are trained on the data of the directions that are the closest to the task-specific direction (e.g., directions $5$ and $6$ for $2$ flows with the target being direction $5.5$). The result shows that within a reasonable range, increasing the number of flows improves the expressivity and consequently results of our model. See Appendix~\ref{sec:extraexp} for more ablation studies.

\subsection{Kitchen Environment}
\label{sec:kitchen}
\textbf{Environment Setup.} We use the  kitchen environment adopted from D4RL~\cite{fu2020d4rl}, which serves as a testbed for many reinforcement learning and imitation learning approaches, like SPiRL~\cite{pertsch2020spirl}, SKiLD~\cite{pertsch2021skild}, relay policy learning~\cite{Gupta2019RelayPL}, and FIST~\cite{Hakhamaneshi2022FIST}. The agent needs to control a 7-DOF robot arm to complete a sequence of four tasks (e.g., move the kettle, or slide the cabinet) in the correct order. The agent will receive a $+1$ reward only upon finishing a task, and $0$ otherwise. The action space is $9$-dimensional and the state space is $60$-dimensional. This environment is very challenging, as high-precision control of the robot arm is needed and also a long horizon of $280$ timesteps is needed. Moreover, there is a small noise applied to each agent action, which requires the agent to be robust.

\textbf{Dataset Setup.} We use two dataset settings, which are adopted from SKiLD and FIST (denoted as \textit{Kitchen-SKiLD} and \textit{Kitchen-FIST} below). In Kitchen-SKiLD, we use $601$ teleoperated sequences that perform a variety of task sequences as the task-agnostic dataset, and use \textit{only one} trajectory for the task-specific dataset. In Kitchen-FIST, we use part of the task-agnostic dataset (about $200-300$ trajectories) that \textit{does not} contain a particular task in the task-specific dataset, and use \textit{only one} trajectory for the task-specific dataset. There are two different task-specific datasets in Kitchen-SKiLD, and four different task-specific datasets in Kitchen-FIST. The latter is significantly harder, as the agent must learn a new task from very little data. For simplicity, we denote them as ``SKiLD-A/B'' and ``FIST-A/B/C/D'' respectively. See Appendix~\ref{sec:app1}  for details on each task. 

\textbf{Experimental Setup.} We use $k$-means to partition the task-agnostic datasets into $24$ different clusters, and train $24$ flows accordingly. For each flow, we use a fully-connected network with $2$ hidden layers of width $256$ with ReLU activation as a ``block'' for our algorithm. For the combination of flows, we use a fully-connected network with $1$ hidden layer of width $64$ with ReLU activation. Each layer of the deep net (except the output layer) described above has a 1D batchnorm function. The blocks are used analogously to  the fetchreach environment. We use a batch size of $256$ for the task-agnostic dataset and a batch size of $128$ for the task-specific dataset. Other training hyperparameters are identical to the fetchreach environment. For each  RL training, we use proximal policy optimization (PPO)~\cite{Schulman2017PPO} for 200K environment steps, with update interval being $2048$ (Kitchen-SKiLD) / $4096$ (Kitchen-FIST), $60$ epochs per update, and a batch size of $64$. 

\begin{figure}[t]
{
\begingroup

\centering
\begin{minipage}[c]{0.28\linewidth}
\subfigure[Kitchen-SKiLD-A]{\includegraphics[width=\linewidth]{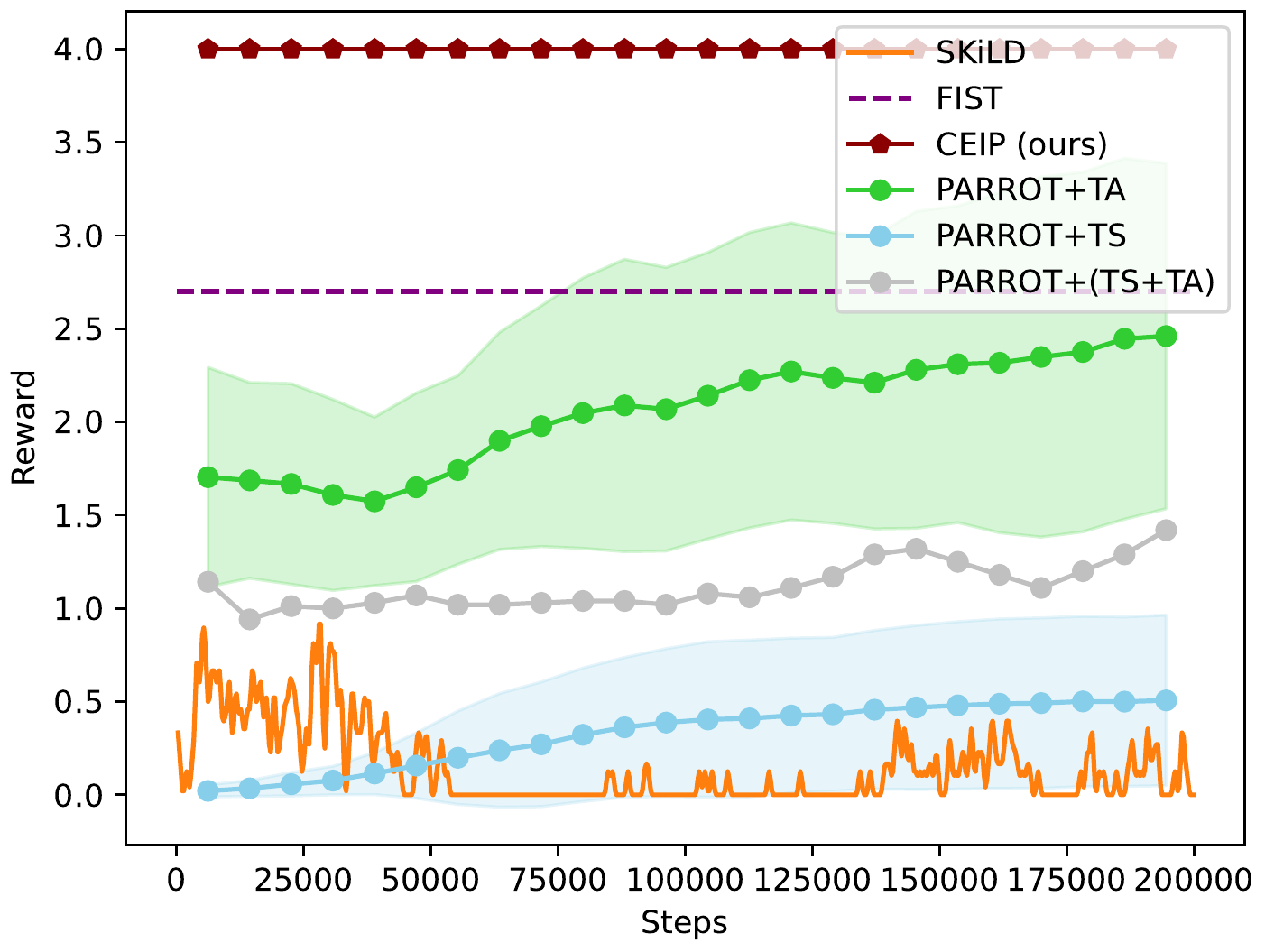}}
\subfigure[Kitchen-FIST-B]{\includegraphics[width=\linewidth]{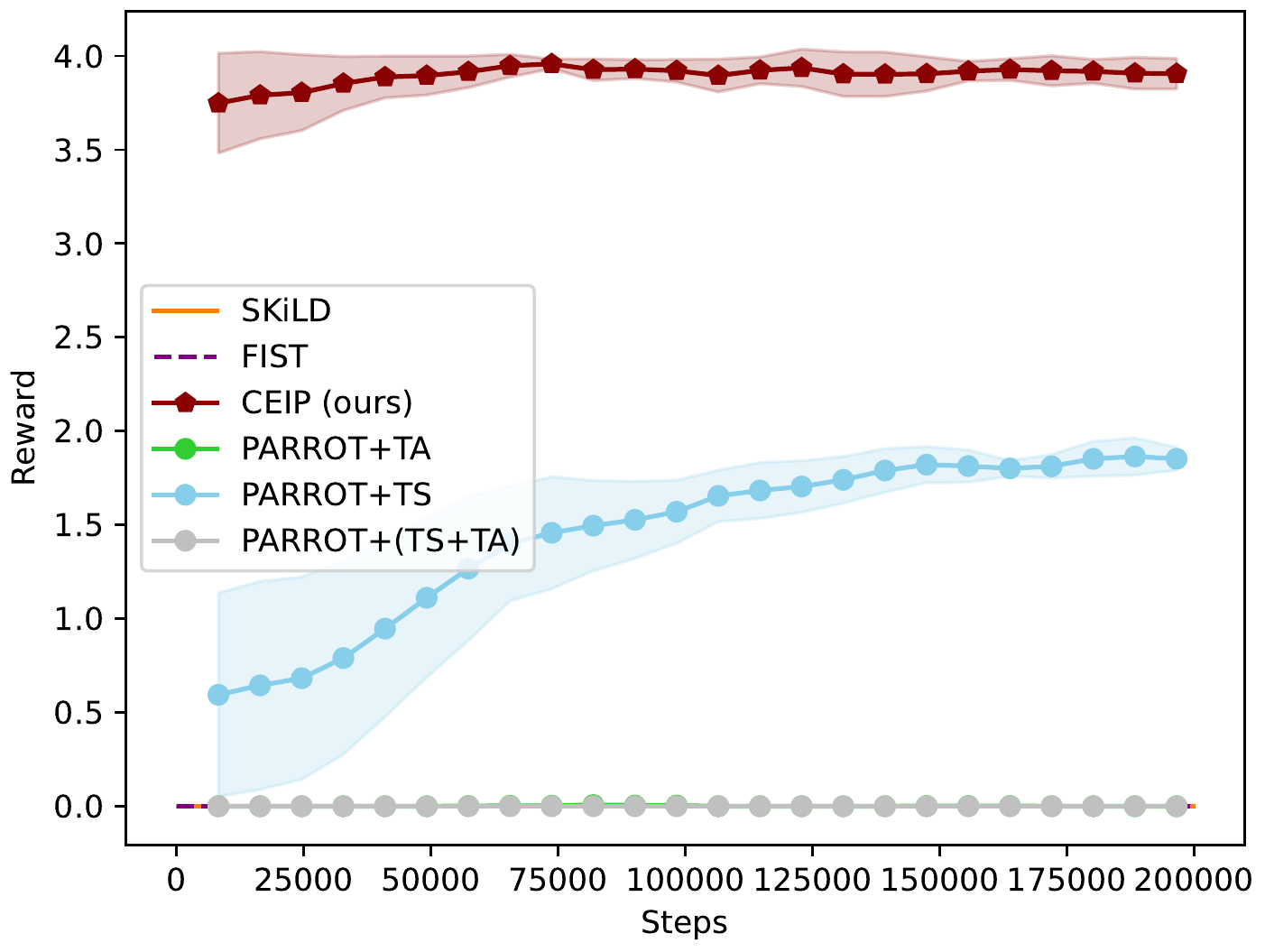}}
\end{minipage}
\begin{minipage}[c]{0.28\linewidth}
\subfigure[Kitchen-SKiLD-B]{\includegraphics[width=\linewidth]{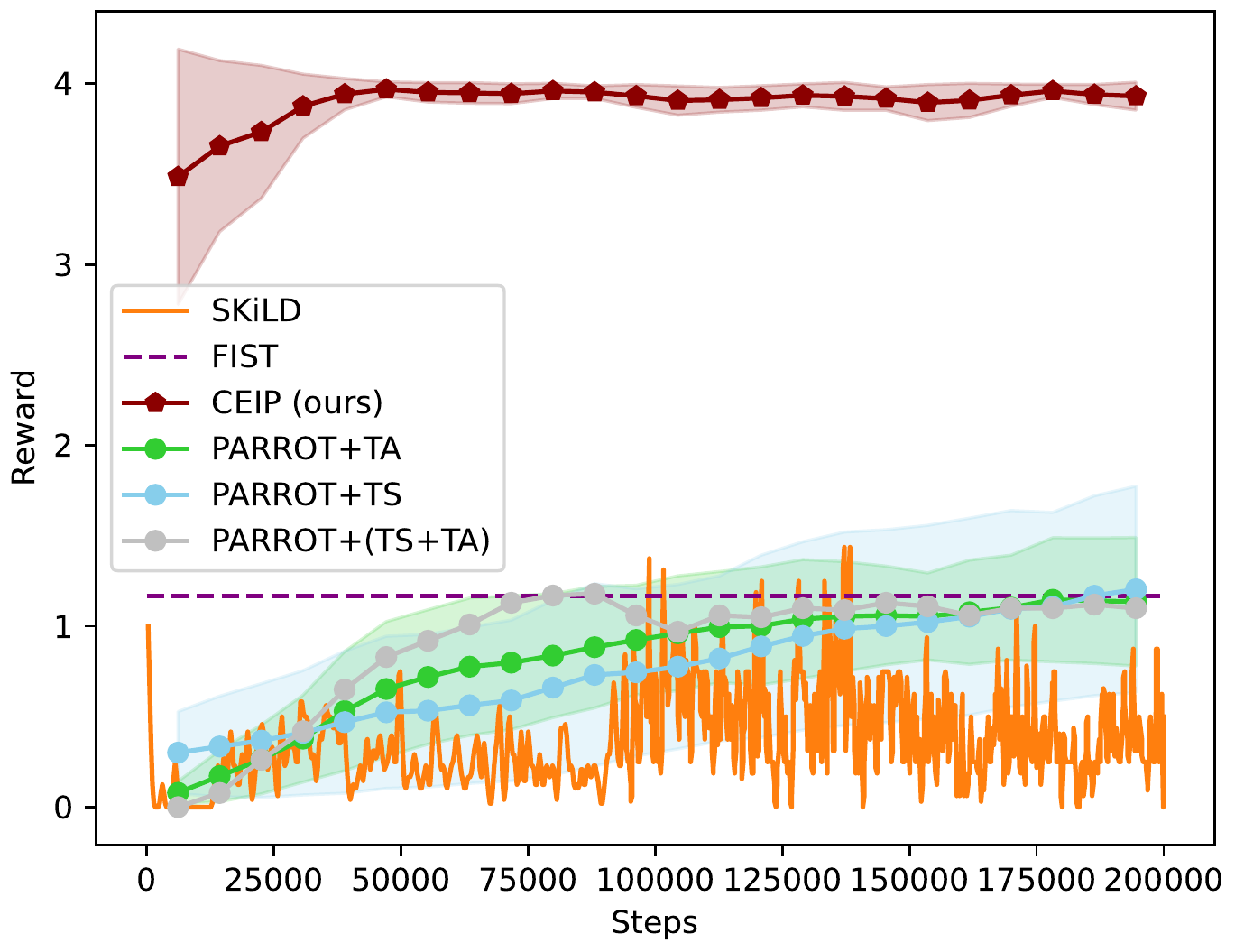}}
\subfigure[Kitchen-FIST-C]{\includegraphics[width=\linewidth]{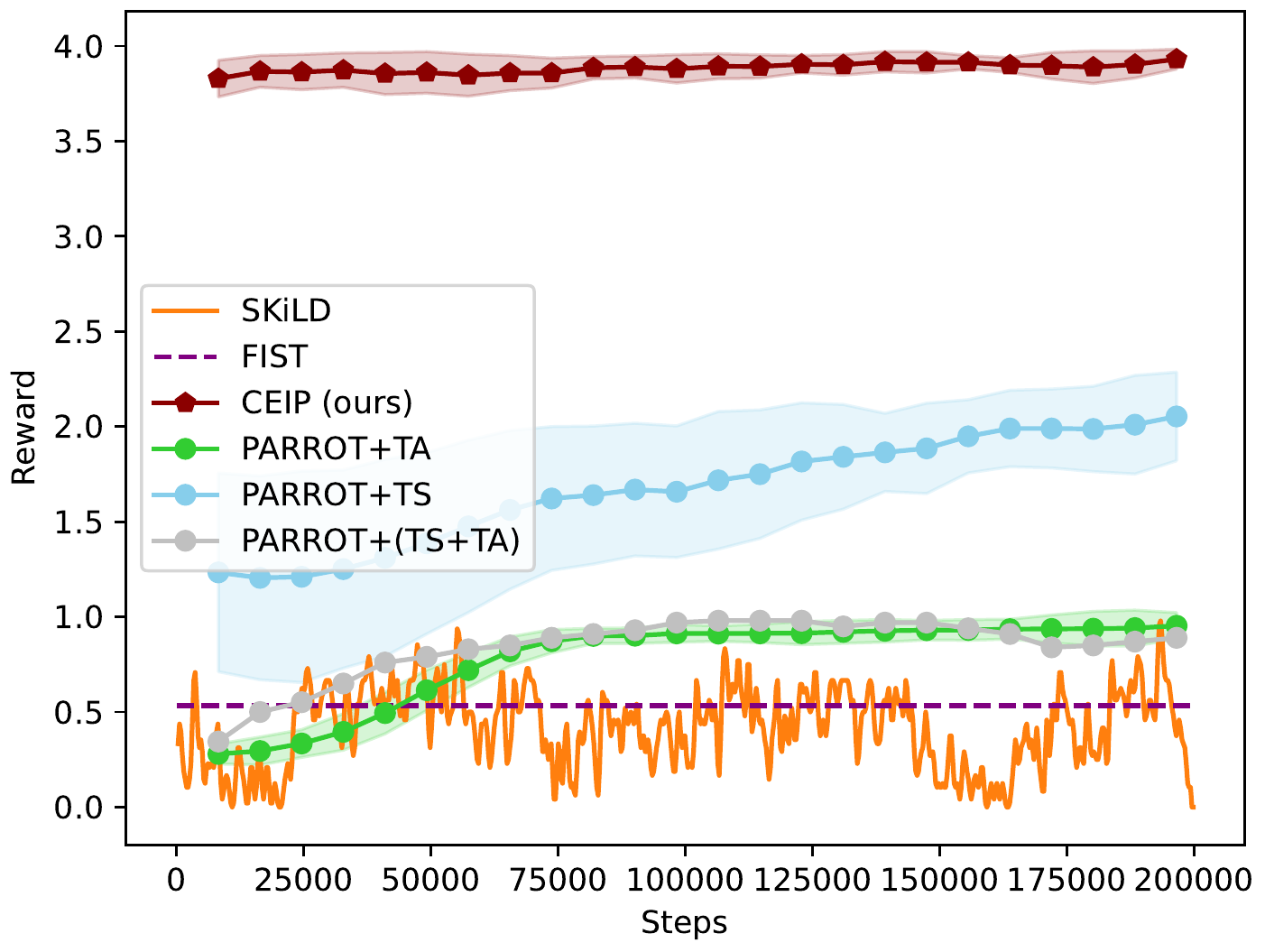}}
\end{minipage}
\begin{minipage}[c]{0.28\linewidth}
\subfigure[Kitchen-FIST-A]{\includegraphics[width=\linewidth]{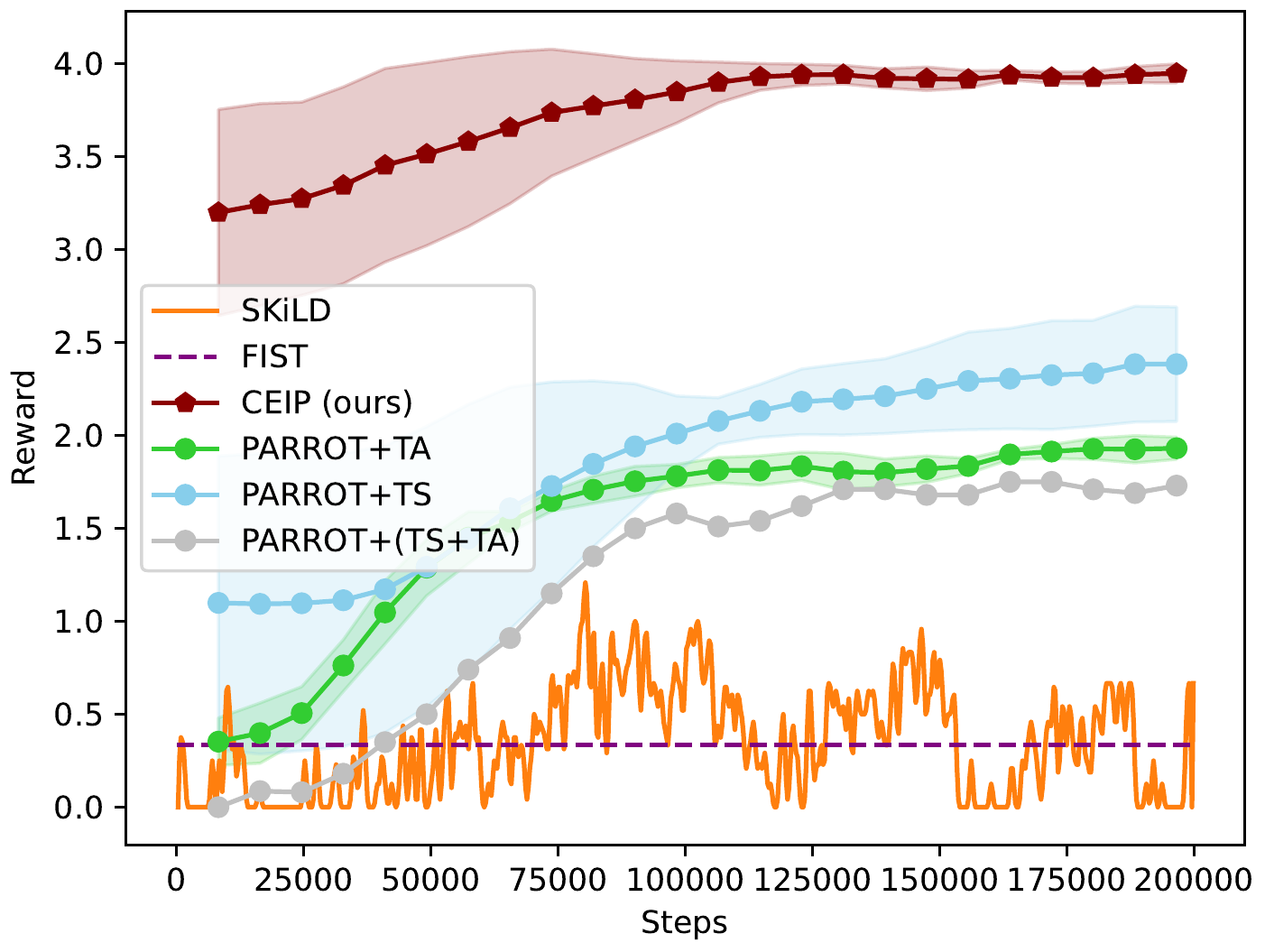}}
\subfigure[Kitchen-FIST-D]{\includegraphics[width=\linewidth]{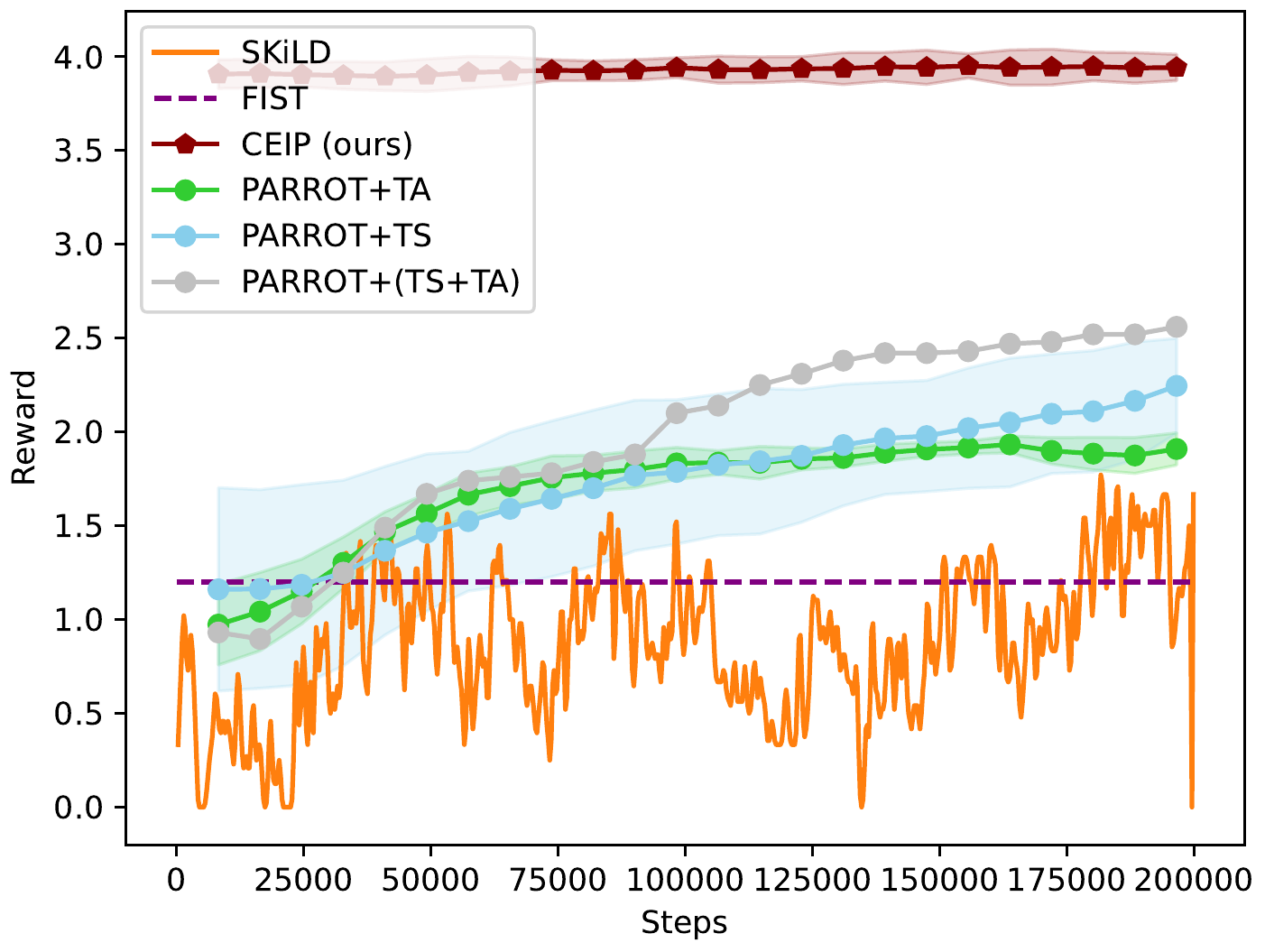}}
\end{minipage}
\vspace{-0.2cm}
\caption{Comparison on Kitchen-SKiLD and Kitchen-FIST environments.}
\label{fig:kitchen_main}

\endgroup
}
\vspace{-0.3cm}
\end{figure}

\textbf{Main Results.} Fig.~\ref{fig:kitchen_main} shows the main  results on Kitchen-SKiLD and Kitchen-FIST. Our method outperforms all other baselines in all of the $6$ settings of the task-agnostic and task-specific datasets. 
For our method, we use the task-specific single flow $f_{n+1}$, explicit prior, and the push-forward technique. We compare to the original PARROT formulation. See Appendix~\ref{sec:extraexp} for ablation studies of PARROT with explicit prior and our method without $f_{n+1}$ or explicit prior.

\begin{figure}[t]
    \centering
    \begin{minipage}[c]{0.4\linewidth}
    \centering
    \subfigure[Illustration of coefficient]{\includegraphics[height=3cm]{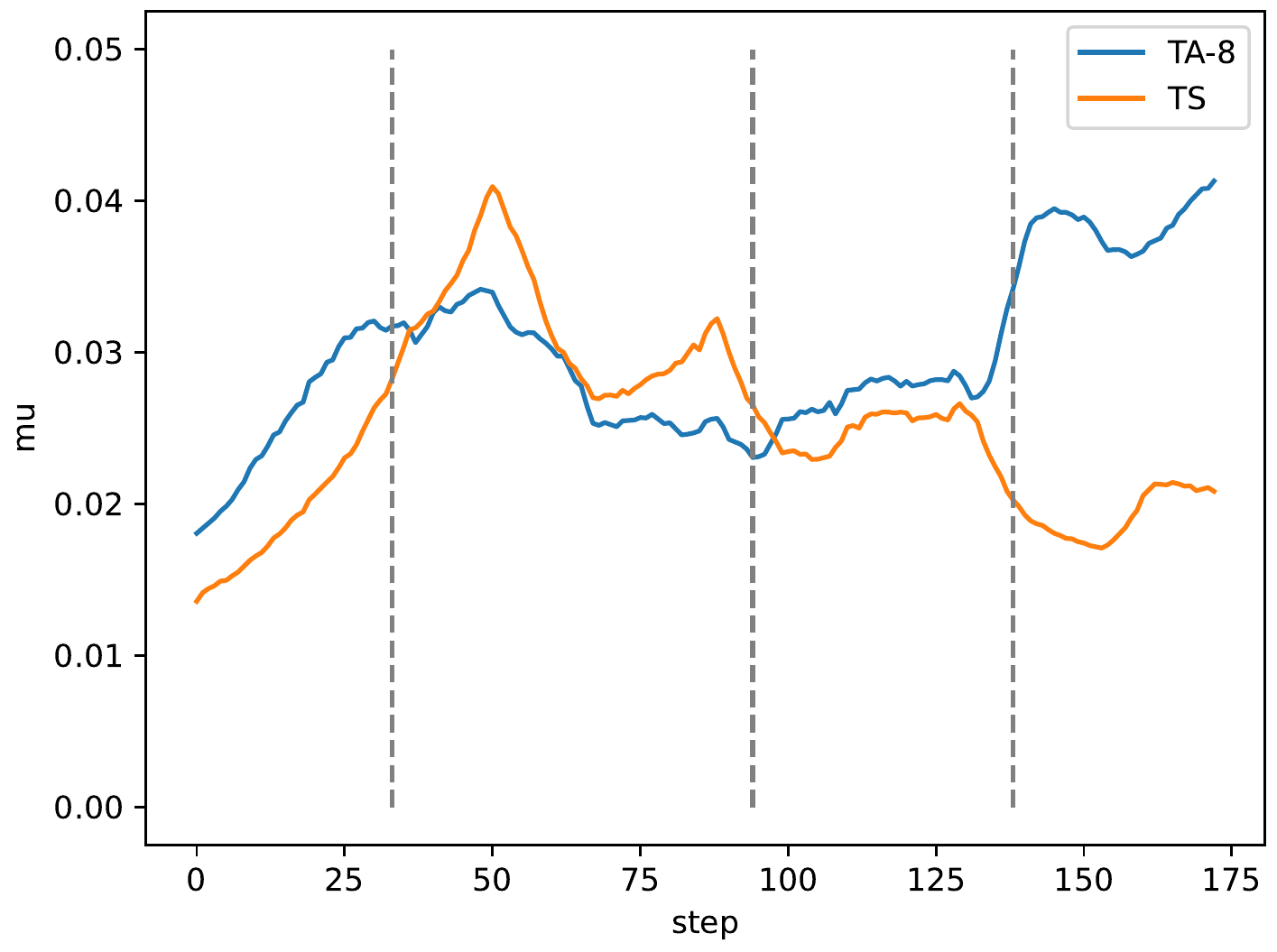}}
    \end{minipage}
    \begin{minipage}[c]{0.4\linewidth}
    \centering
    \subfigure[Average episode length]{\includegraphics[height=3cm]{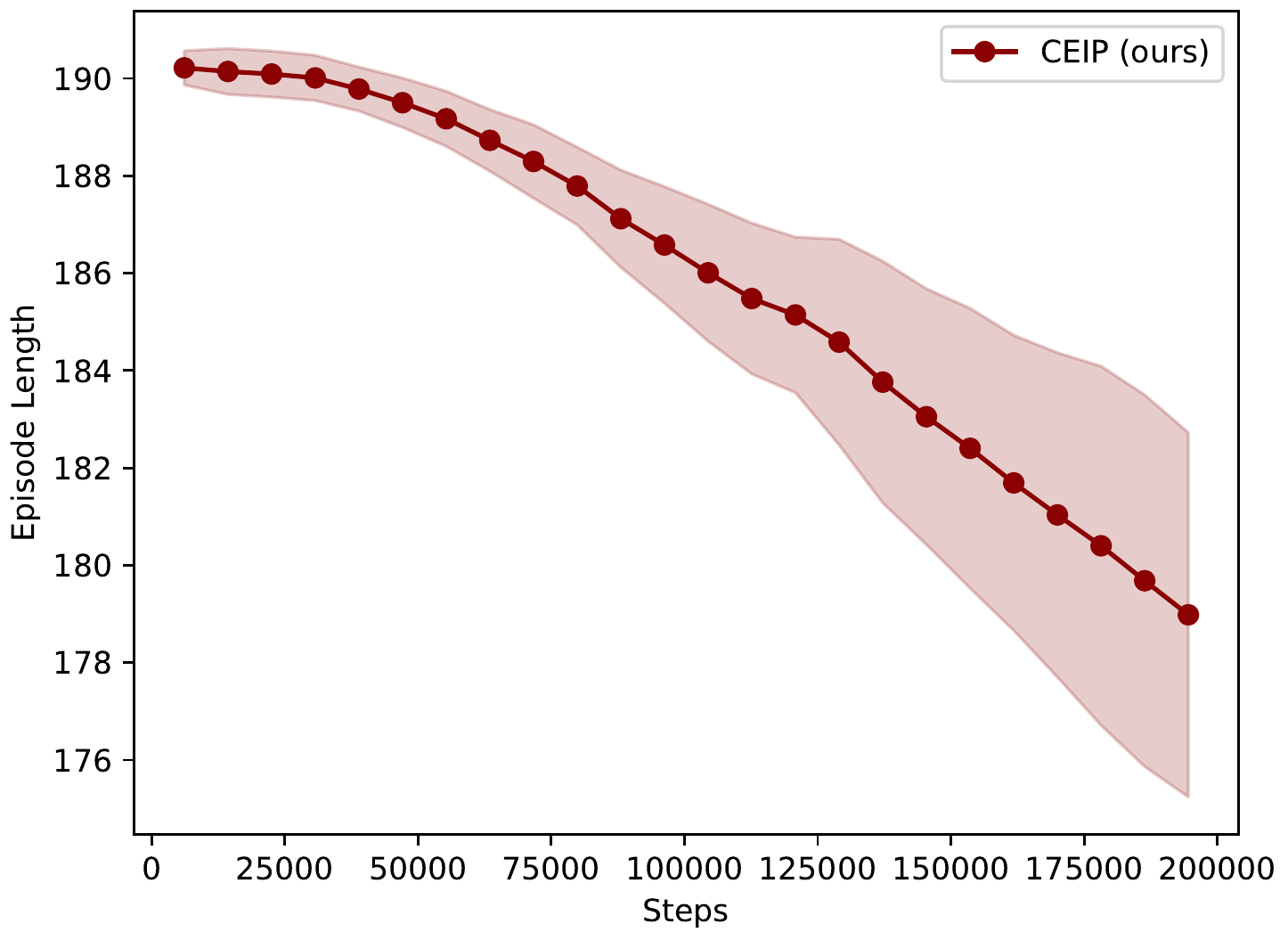}}
    \end{minipage}
    \caption{a) Illustration of the coefficient change of a trained CEIP model during an episode of Kitchen-SKiLD-A. This CEIP model is trained with the task-specific single flow and without the explicit prior. `TA-$8$' is the $8$-th single flow for the task-agnostic dataset, and `TS' is the single flow for the task-specific dataset. The grey dotted lines are the partition of different subtasks. b) Average episode length of our method on the Kitchen-SKiLD-A task. The episode ends immediately when all the tasks are completed; thus, shortening length means that RL helps to find policy with more efficient completion of tasks.}
    \label{fig:coeff_and_pathlen}
\end{figure}

\textbf{Does CEIP overly rely on the task-specific flow if it is used?} One concern for our method could be: does the task-specific single flow dominate the model? Theoretically, when all flows are perfect, a trivial combination of flows that minimizes the training objective is to set $\lambda_{n+1}=1$, $\mu_{n+1}=1$ for the task-specific single flow, and $\lambda_i\approx 0, \mu_i\approx 0$ for $i\neq n+1$. To study this concern, we plot the change of the coefficient $\mu$ during an episode in Fig.~\ref{fig:coeff_and_pathlen}. We observe that the single flow on the task-specific dataset is not dominating the combination of the flow, despite being trained on the task-specific dataset. The blue curve with legend `TA-8' in Fig.~\ref{fig:coeff_and_pathlen} shows the coefficient for the $8$th flow trained in the task-agnostic dataset. It exhibits an increase of $\mu$ at the end of an episode, as the last subtask in the target task is more relevant to the prior encoded in the $8$th flow. Intuitively, over-reliance in our design (Fig.~\ref{fig:arch_illu} in the Appendix) is discouraged, because of the softplus function and the positive offset applied on $\mu$. For over-reliance, all task-agnostic flows $f_i$ with $i\in\{1,2,\dots,n\}$ should have a coefficient of $\mu_i=0$, which is hard to approach due to the offset of $\mu$ and softplus. In fact, a degenerated CEIP is essentially PARROT+TS(+EX+forward), which is worse than our method but still a powerful baseline.

\textbf{Is reinforcement learning useful in cases with a perfect initial reward?} Fig.~\ref{fig:coeff_and_pathlen} shows the episode length of our method on Kitchen-SKiLD-A. Even if the reward is already perfect, the reinforcement learning process is still able to maximize discounted reward, which optimizes the path.

\subsection{Office Environment}
\label{sec:office}

\textbf{Environment and Dataset Setups.} We follow SKiLD, where a robot with $8$-dimensional action space and $97$-dimensional state space needs to put three randomly selected items on a table into three containers in the correct, randomly generated order. The agent will receive a $+1$ reward when it completes a subtask (e.g., picking up an item, or dropping the item at the right place), and $0$ otherwise.  This environment is even harder than the kitchen environment, as the agent must manipulate freely movable objects and the number of possible subtasks in the task-agnostic dataset is much larger than that in the kitchen environment. We use the same task-agnostic dataset as SKiLD, which contains $2400$ trajectories with randomized subtasks sampled from a script policy. For the task-specific dataset, we use $5$ trajectories for a particular combination of tasks.

\textbf{Experimental Setup.} Similar to the kitchen environment, we use $k$-means over the last state of each trajectory and partition the task-agnostic dataset into $24$ clusters. The architecture and training paradigm of the flow model are identical to those used in the kitchen environment. For RL training, we use PPO for 2M environment steps, with update interval being $4096$ environment steps, $60$ epochs per update, and a batch size of $64$. All other hyperparameters follow the kitchen environment setting. We run each method with $3$ different seeds.

\begin{wrapfigure}{r}{0.73\textwidth}
{
\begingroup
\centering
\begin{minipage}[c]{0.32\linewidth}
\subfigure[Main result]{\includegraphics[width=\linewidth]{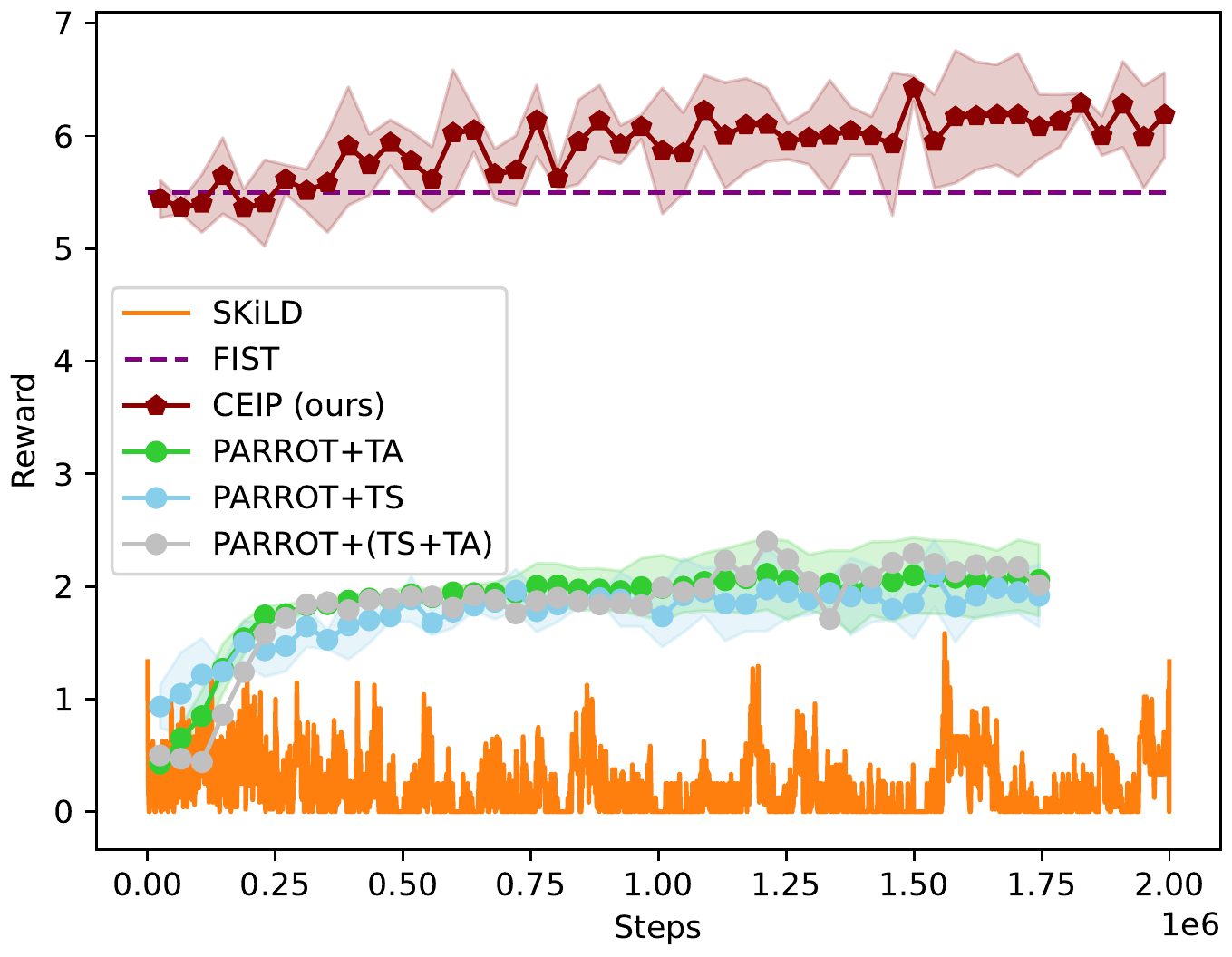}}
\end{minipage}
\begin{minipage}[c]{0.32\linewidth}
\subfigure[Ours ablation]{\includegraphics[width=\linewidth]{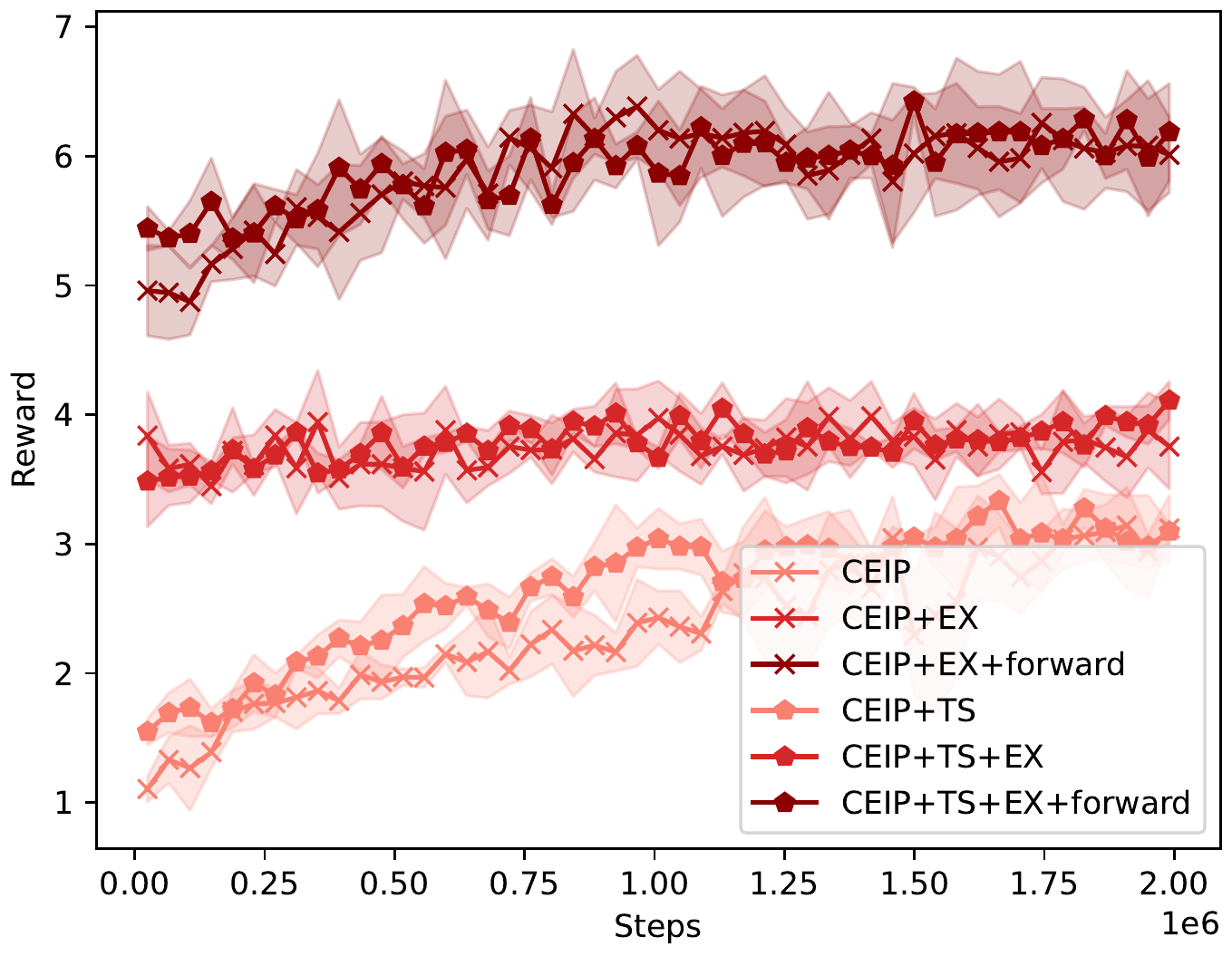}}
\end{minipage}
\begin{minipage}[c]{0.32\linewidth}
\subfigure[PARROT ablation]{\includegraphics[width=\linewidth]{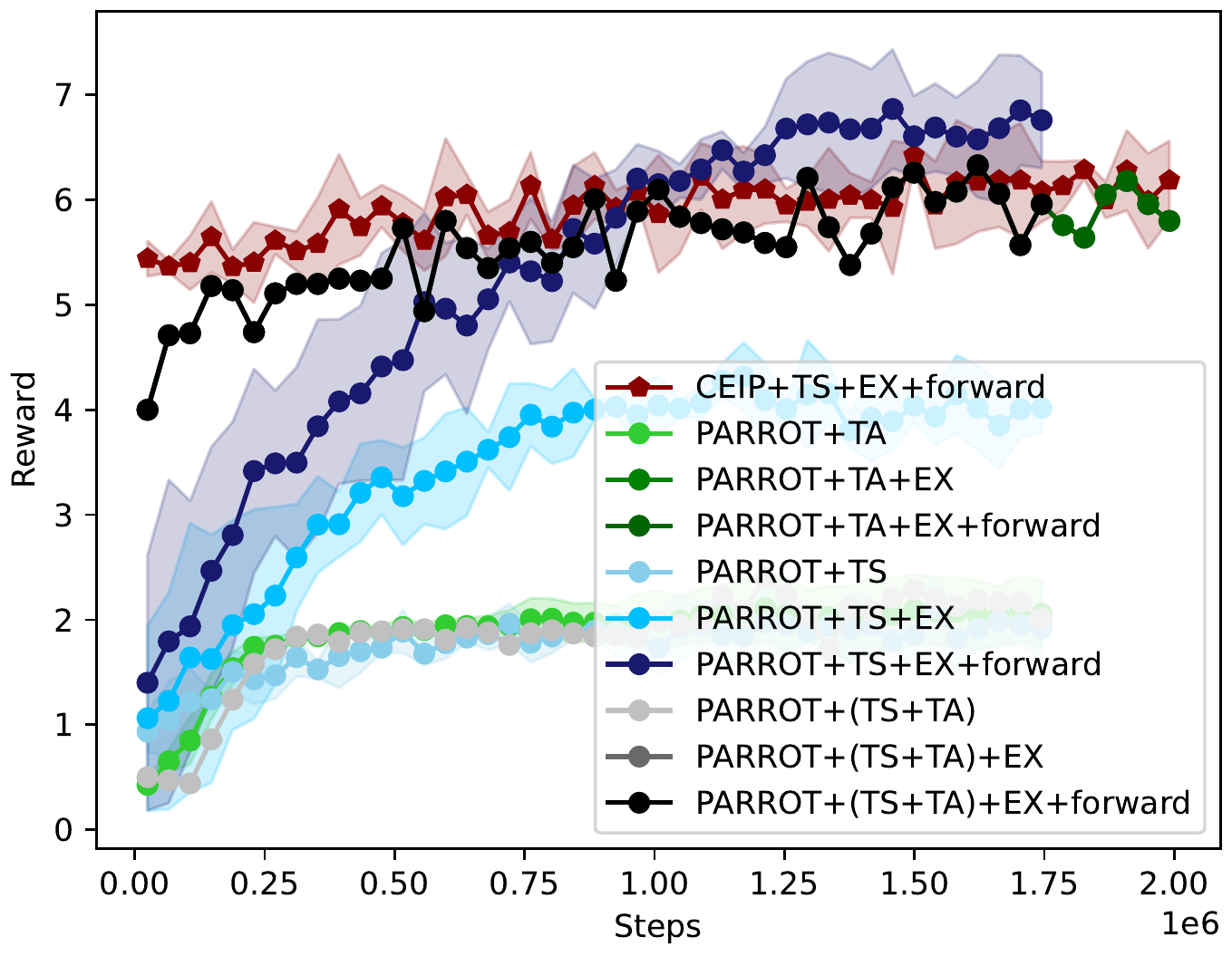}}
\end{minipage}
\caption{Main result and ablation of our method and PARROT on the office environment.}
\label{fig:office}
\endgroup
}
\end{wrapfigure}
\textbf{Main Results and Ablation.}  Fig.~\ref{fig:office}a shows the main  result across different methods. Our method with explicit prior, push-forward technique, and task-specific flow outperforms all baselines. FIST works well in this environment, probably because of two reasons: 1) there are a sufficient number of task-specific trajectories for the VAE-architecture, and 2) the office environment is less noisy than the kitchen environment.
However, as FIST does not contain a reinforcement learning stage, it has no chance to improve on a decent policy which could have been a good start for an RL agent. Fig.~\ref{fig:office}b shows the ablation of our method. While the task-specific single flow $f_{n+1}$ does not help in this environment, the explicit prior greatly improves results. Also, as illustrated, the reward curve of the variants with the explicit prior but without the push-forward skill does not grow, which is due to the agent getting stuck as described at the end of Sec.~\ref{sec:explicitprior}. 
Fig.~\ref{fig:office}c shows the ablation result of PARROT, which also emphasizes that the explicit prior and push-forward skill greatly improve results.

\section{Related Work}
\label{gen_inst}

\textbf{Reinforcement Learning with Demonstrations.} Using demonstrations to improve the sample efficiency of RL is an established direction~\cite{Schaal1996LfD, Wang2021LearningTW, Giusti2016Forest, Lynch2019LearningLP}. Recently, the use of task-agnostic demonstrations has gained popularity, as task-specific data need to be sampled from a particular expert and can be expensive to acquire~\cite{pertsch2020spirl, pertsch2021skild, Singh2021ParrotDB}. To utilize the prior, skill-based methods such as SPiRL~\cite{pertsch2020spirl}, SKiLD~\cite{pertsch2021skild}, and SIMPL~\cite{nam2022simpl} extract action sequences from the dataset with a VAE-based model, while TRAIL~\cite{Yang2021TRAIL} recovers transitions from a task-agnostic dataset with uniformly randomly sampled action. Our method considers the situation where both task-agnostic \emph{and} task-specific data exist and significantly improves results over prior work with similar settings, e.g., SKiLD~\cite{pertsch2021skild}.


\textbf{Action Priors.} An action prior is a common way to utilize demonstrations for reinforcement learning~\cite{pertsch2021skild} and imitation learning~\cite{Hakhamaneshi2022FIST}. Most work uses an implicit prior, where a probability distribution of actions conditioned on a state is learned by a deep net and then used to rule out unlikely attempts~\cite{Biza2021ActionPrior}, to form a hierarchical structure~\cite{pertsch2021skild, Singh2021ParrotDB}, or to serve as a regularizer for RL training~\cite{pertsch2021skild, Rengarajan2022LOGO}, preventing the agent to stray too far from expert demonstrations. Explicit priors are less explored. They come in the form of nearest neighbors~\cite{Arunachalam2022dexterbot} (as in our work) or in the form of locally weighted regression~\cite{pari2021surprising}. They are utilized in robotics~\cite{Arunachalam2022dexterbot,Chaplot2020SLAM, Schaal1994Juggling, pari2021surprising} and early work of RL with demonstrations~\cite{Brys2015RLfD}. 
Another way to explicitly use demonstrations includes 
filling the buffer of offline RL algorithms with transitions sampled from an expert dataset to help exploration~\cite{Vecerk2017LeveragingDF,Nair2020AWAC, Hester2018DQLfD}. Different from all such work, we propose a novel way of using both implicit and explicit  priors. 


\textbf{Normalizing Flow.} Normalizing flows are a generative model that can be used for variational inference~\cite{vdberg2018sylvester, Kingma2016VIflow} and density estimation~\cite{Papamakarios2017MAFDE, Huang2018NAF} 
and come in different forms: RealNVP~\cite{Dinh17RealNVP}, Glow~\cite{Kingma2018GlowGF}, or autoregressive flow~\cite{Papamakarios2017MAFDE}. Many methods use normalizing flows in reinforcement learning~\cite{Ma2020NFMAS, Tang2018BoostingTR, TouatiSRPV19, Ward2019ImprovingEI, mazoure2019leveraging, Khader2021Stable} and imitation learning~\cite{Chang2021ILFLOW}. However, most prior work uses normalizing flows as a strong density estimator to exploit a richer class of policies. Most closely related to our work is PARROT~\cite{Singh2021ParrotDB}, which trains a single normalizing flow as an implicit prior. Different from our work, PARROT does not differentiate tasks among the task-agnostic dataset and does not use an explicit prior. More importantly, different from prior work, we develop a simple yet effective way to combine flows using learned coefficients. While there are some approaches that combine  flows via variational mixtures~\cite{Ciobaru2021Mixtures, Pires2020VariationalMO}, they have not been shown to succeed on challenging RL tasks.

\textbf{Few-shot Generalization.} 
Few-shot generalization~\cite{TriantafillouZD20} is broadly related, as 
a model is first trained across different datasets, and then adapted to a new dataset with small sample size. For example, similar to our work, FLUTE~\cite{Triantafillou2021FLUTE},  SUR~\cite{Dvornik2020SUR}, and URT~\cite{Lu2021URT} use models  for multiple datasets, which are then combined via weights for few-shot adaptation. Other methods have shared parameters across different tasks and only used some components within the model for adaptation~\cite{Puigcerver2021STL, Triantafillou2021FLUTE, Zintgraf2019FastCA, Flennerhag2020MetaLearningWW,RenYehNEURIPS2020}. While most work focuses on classification tasks, we address more complex RL tasks. Also, different from existing work, we found training  of independent 1-layer flows without shared layers to be more flexible, and free from negative transfer as also reported by~\cite{Javaloy2021RotoGradGH}. 

\section{Discussion and Conclusion}
\label{sec:conclusion}
We developed \textbf{CEIP}, a method for reinforcement learning which combines explicit and implicit priors obtained from task-agnostic and task-specific demonstrations. For implicit priors we use normalizing flows. For explicit priors we use a database lookup with a push-forward retrieval. In three challenging environments, we show that \textbf{CEIP} improves upon baselines.

\textbf{Limitations.} Limitations of CEIP are as follows: 1) \textit{Training time}. The use of demonstrations requires training a decent number of flows which can be time-consuming, albeit mitigated to some extent by parallel training. 2) \textit{Reliance on optimality of expert demonstrations.} Similar to prior work like SKiLD~\cite{pertsch2021skild} and FIST~\cite{Hakhamaneshi2022FIST}, our method assumes availability of optimal state-action trajectories for the target task. Accuracy of those demonstrations impacts results. Future work will focus on improving robustness and generality. 3) \textit{Balance between the degree of freedom and generalization in fitting the flow mixture.} Fig.~\ref{fig:fetchreach_plot_ablation_ours_dataset} reveals that more degrees of freedom in the flow mixture improve results of CEIP. Our current design uses a linear combination which offers $O(n)$ degrees of freedom ($\mu$ and $\lambda$), where $n$ is the number of flows. In contrast, too many degrees of freedom will result in overfitting. It is interesting future work to study this tradeoff.

\textbf{Societal impact.} Our work helps to train  RL agents more efficiently from  demonstrations for the same and closely related tasks, particularly when the environment only provides sparse rewards. If successful, this expands the applicability of automation. However, increased automation may also cause job loss which negatively impacts society.

\textbf{Acknowledgements.} This work was supported in part by NSF under Grants 1718221, 2008387, 2045586, 2106825, MRI 1725729, NIFA award 2020-67021-32799, the Jump ARCHES endowment through the Health Care Engineering Systems Center, the National Center for Supercomputing Applications (NCSA) at the University of Illinois at Urbana-Champaign through the NCSA Fellows program, and the IBM-Illinois Discovery Accelerator Institute. We thank NVIDIA for a GPU.

\newpage
\bibliographystyle{abbrvnat}
\bibliography{neurips_2022}

\newpage
\appendix

\section*{Appendix: CEIP: Combining Explicit and Implicit Priors for Reinforcement Learning with Demonstrations}
\label{sec:app}

This Appendix is organized as follows. First, we reiterate and highlight our key observations. In Sec.~\ref{sec:alg}, we then provide the pseudocode for training the implicit prior and the downstream reinforcement learning. Afterwards, we provide additional implementation details of the proposed method and major baselines in Sec.~\ref{sec:implementation}, and additional details of experimental settings in Sec.~\ref{sec:app1}. In Sec.~\ref{sec:extraexp}, we provide additional experimental results and ablation studies. In Sec.~\ref{sec:comp_resource}, we describe the computational resources consumed by and the training time of each method. Finally, in Sec.~\ref{sec:license}, we describe the licenses of assets which we used to develop our code.

The key findings of our work include the following:

\begin{itemize}
    \item \textbf{Is a task-specific flow necessary?} In environments where the episode length is relatively short and the dynamics are relatively simple, CEIP works better without the task-specific flow, explicit prior, and push-forward technique as the training complexity is unnecessarily increased. This is shown in Sec.~\ref{sec:expFR}.
    \item \textbf{When is a task-specific flow helpful?} In environments where some tasks of the task-specific dataset are not part of the task-agnostic dataset, a  flow trained on the task-specific dataset improves performance. This is shown in Sec.~\ref{sec:expKT}.
    \item \textbf{How related should the tasks in the task-agnostic dataset be to the task at hand?} For both PARROT and CEIP, more related data in the task-agnostic dataset are beneficial. However, CEIP can automatically discover and compose related flows; in contrast, PARROT works better only when the dataset fed into the normalizing flow is manually picked to be more relevant to the target task. This is shown in Sec.~\ref{sec:expFR}.
    \item \textbf{Will ground-truth labels help the performance of CEIP?} Ground-truth labels will sometimes improve the performance of CEIP; however, this is not always the case. This is shown in Sec.~\ref{sec:expKT}.
    \item \textbf{How will simple baselines, e.g., behavior cloning and replaying demonstrations do?} We find those simple baselines to not work very well, which indicates the non-trivial nature of our testbed. However, introducing an explicit prior will significantly improve the performance of behavior cloning. This is shown in Sec.~\ref{sec:expKT}.
    \item \textbf{How robust is CEIP with respect to the precision of task-specific demonstrations?} Similar to prior work such as FIST, imprecise task-specific demonstrations will affect performance. Nevertheless, we find CEIP to be more robust than prior work. This is shown in Sec.~\ref{sec:expKT}.
    \item \textbf{What is the impact of using an explicit prior in PARROT?} PARROT results improve when an explicit prior is used, which further supports the design of CEIP. See ablation studies in Sec.~\ref{sec:expFR} and Sec.~\ref{sec:expKT}.
\end{itemize}

To easily compare CEIP to baselines, we summarize all results achieved at the end of the training process for the proposed method and baselines on all testbeds in Table~\ref{tab:summary}. To better understand the behavior of each method, please also see the code and videos of trajectories which are part of this Appendix.

\begin{table}[t]
\setlength{\tabcolsep}{1pt}
    \centering
    \begin{tabular}{cccccc}\toprule
        Environment & CEIP (ours) & PARROT+TA & PARROT+TS & FIST & SKiLD  \\ \midrule
        Fetchreach-4.5  &$\mathbf{-10.03}^\dagger${\scriptsize $\pm 0.64$} & $-19.33${\scriptsize$\pm 9.59$} & $-20.30${\scriptsize $\pm 10.62$} & $-34.80${\scriptsize$\pm 8.33$} & $-39.91${\scriptsize$\pm 0.14$}       \\
        Fetchreach-5.5 & $\mathbf{-9.76}^\dagger${\scriptsize$\pm 0.47$} & $-20.49${\scriptsize $\pm 11.51$} & $-14.32${\scriptsize$\pm 7.53$} & $-39.86${\scriptsize$\pm 0.50$} & $-38.38${\scriptsize$\pm 2.81$}\\
        Fetchreach-6.5 & $\mathbf{-9.08}^\dagger${\scriptsize$\pm 0.36$} & $-14.52${\scriptsize$\pm 9.44$} & $-18.52${\scriptsize$\pm 2.34$} & $-38.30${\scriptsize$\pm 5.28$} & $-40.00${\scriptsize $\pm 0.00$}\\
        Fetchreach-7.5 & $\mathbf{-10.29}^\dagger${\scriptsize$\pm 0.67$} & $\mathbf{-10.34}${\scriptsize$\pm 0.79$} & $\mathbf{-10.24}${\scriptsize$\pm 0.69$} & $-39.87${\scriptsize $\pm 0.72$} & $-38.45${\scriptsize $\pm 2.67$}\\ \midrule
        Kitchen-SKiLD-A & $\mathbf{4.00}${\scriptsize $\pm 0.00$} & $2.52${\scriptsize $\pm 0.96$} & $0.51${\scriptsize $\pm 0.46$} & $2.70${\scriptsize $\pm 1.23$} & $0.06${\scriptsize $\pm 0.10$}\\
        Kitchen-SKiLD-B & $\mathbf{3.93}${\scriptsize $\pm 0.08$} & $1.13${\scriptsize $\pm 0.35$} & $1.25${\scriptsize$\pm 0.60$} & $1.17${\scriptsize$\pm 0.93$} & $0.48${\scriptsize$\pm 0.48$}\\ \midrule
        Kitchen-FIST-A & $\mathbf{3.95}${\scriptsize $\pm 0.05$} & $1.94${\scriptsize $\pm 0.07$} & $2.40${\scriptsize $\pm 0.31$} & $0.33${\scriptsize $\pm 0.70$} & $0.67${\scriptsize $\pm 1.15$}\\
        Kitchen-FIST-B & $\mathbf{3.89}${\scriptsize $\pm 0.07$} & $0.00${\scriptsize $\pm 0.00$} & $1.85${\scriptsize $\pm 0.05$} & $1.20${\scriptsize$\pm 0.54$} &  $0.00${\scriptsize $\pm 0.00$}\\
        Kitchen-FIST-C & $\mathbf{3.92}${\scriptsize $\pm 0.06$} & $0.96${\scriptsize $\pm 0.06$} & $2.07${\scriptsize $\pm 0.23$} & $0.00${\scriptsize$\pm 0.00$}&  $0.33${\scriptsize $\pm 0.57$}\\
        Kitchen-FIST-D & $\mathbf{3.94}${\scriptsize $\pm 0.07$} & $1.92${\scriptsize$\pm 0.06$} & $2.27${\scriptsize$\pm 0.24$} & $0.53${\scriptsize $\pm 0.50$}& $1.67${\scriptsize $\pm 0.58$}\\
        Office & $\mathbf{6.33}${\scriptsize$\pm 0.30$} & $2.05${\scriptsize $\pm 0.31$} & $1.97${\scriptsize $\pm 0.22$} & $5.50${\scriptsize $\pm 1.12$}& $0.50${\scriptsize $\pm 0.50$} \\ \bottomrule
    \end{tabular}
    \caption{Summary of the results of each method on all environments at the end of training (higher is better). For CEIP (our method), we are not using the explicit prior, task-specific single flow, and push-forward technique for fetchreach (which is denoted by `$\dagger$'). We use all of them for the other experiments. For PARROT, we are not using the explicit prior, task-specific single flow, and push-forward technique, as all of them are our contributions. However, as shown in ablation study in Sec.~\ref{sec:extraexp}, these components are general and can be used to improve the performance of PARROT.}
    \label{tab:summary}
\end{table}

\section{Algorithm Details} 
\label{sec:alg}

Alg.~\ref{alg:NF} provides the pseudocode for training the implicit prior. Alg.~\ref{alg:RL} illustrates how we use the policy $\pi(z|s)$ and the flows to compute the real-world action $a$, when an explicit prior is available (i.e., condition $u=[s, s_{\text{next}}]$) and when using the push-forward technique.

\begin{algorithm}[t]
\caption{Training of Implicit Prior}
\label{alg:NF}
\SetAlgoVlined
\SetKwInOut{Input}{Input}
\SetKwInOut{Output}{Output}
\SetKw{KwBy}{by}
\Input{dataset $D_1, D_2, ..., D_n ,D_{\text{TS}}$}
\Input{training epoch for single flow $M$, for combination $M_2$}
\Input{learning rate $a$}
\Output{normalizing flow $f_{TS}$, parameterized by $\mu(u)$, $\lambda(u)$, $c_i(u)$, and $d_i(u)$ where $i\in\{1,2,\dots,n+1\}$}
\newlength{\commentWidth}
\setlength{\commentWidth}{6cm}
\newcommand{\atcp}[1]{\tcp*[f]{\makebox[\commentWidth]{#1\hfill}}}
\Begin{
\sf{
  \tcp{Training single flows}
  \nl \For(\atcp{recall that we denote $D_{\text{TS}}=D_{n+1}$}){$i\in\{1,2,\dots,n+1\}$}{
      \nl \For(\atcp{for loop over epochs}){$j\in\{1,2,\dots,M\}$}{ 
          \nl \ForEach(\atcp{for each data point}){$(u,a)\sim D_i$}{
              \nl {$z_0\gets\frac{a-d_i(u)}{\exp\{c_i(u)\}}$}\atcp{elementwise division}
              \nl {$L=\log p_z(z_0)-c_i(u)^T\mathbf{1}$}\atcp{$z\sim N(0,I)$}
              \nl {$c_i\gets c_i+ a\times\frac{\partial L}{\partial c_i}$}\\
              \nl {$d_i\gets d_i+a\times\frac{\partial L}{\partial d_i}$}
          }
     }
 }
  \tcp{Training the combination of flows}
  \nl \For(\atcp{for loop over epochs}){$j\in\{1,2,\dots,M_2\}$}{
      \nl \ForEach(\atcp{for each data point}){$(u,a)\sim D_{\text{TS}}$}{
          \nl {$\mu_0\gets \mu(u)$}\\
          \nl {$\lambda_0\gets \lambda(u)$}\\
          \nl {$c = \sum_{i=1}^{n+1}\mu_{0,i}c_i(u)$}\\
          \nl {$d = \sum_{i=1}^{n+1}\lambda_{0,i}d_i(u)$}\\
          \nl {$z_0\gets\frac{a-d}{c}$}\atcp{elementwise division}\\
          \nl {$L\gets\log p_z(z_0)-c^T\mathbf{1}$}\atcp{$z\sim N(0, I)$}\\
          \nl {$\mu\gets\mu+a\times\frac{\partial L}{\partial\mu}$}\\
          \nl {$\lambda\gets\lambda+a\times\frac{\partial L}{\partial \lambda}$}
      }
  }
}
}
\end{algorithm}

\begin{algorithm}[t]
\caption{Step Function of Reinforcement Learning}
\label{alg:RL}
\SetAlgoVlined
\SetKwInOut{Input}{Input}
\SetKwInOut{Output}{Output}
\SetKw{KwBy}{by}
\Input{current state $s$, RL policy $\pi(z|s)$}
\Output{action in actual action space $a$}
\newlength{\commentWidthB}
\setlength{\commentWidthB}{9cm}
\newcommand{\atcp}[1]{\tcp*[f]{\makebox[\commentWidthB]{#1\hfill}}}
\Begin{
\sf{
    \tcp{$r$ is the last step referred to in the trajectory}
    \If{A new episode begins}{\ForEach(\\\atcp{reset last reference in each trajectory}){$\tau\in D_{\text{TS}}$}{$r(\tau)\gets -1$}}
    \nl \ForEach{$\tau\in D_{\text{TS}}$}{
         \nl \ForEach(\\\atcp{Assume this is the $i$-th step}){$(s_{\text{key}},a,s_{\text{next}})\in\tau$}{
             \nl{\If(\\\atcp{The second term is an indicator function}){$(s_0, j_0, \tau_0)$ undefined or $(s_{\text{key}}-s)^2+[i\leq r(\tau)]<(s_0-s)^2+[j_0\leq r(\tau_0)]$}{\nl{$s_0\gets s_{\text{key}}$}\\\nl{$j_0\gets i$}\\\nl{$\tau_0\gets \tau$}}}
         }
    }
    \nl {$r(\tau_0)\gets j_0$}\atcp{update last reference for the chosen trajectory}\\
    \nl {$\mu_0\gets \mu(u)$}\\
    \nl {$\lambda_0\gets \lambda(u)$}\\
    \nl {$c \gets \sum_{i=1}^{n+1}\mu_{0,i}c_i(u)$}\\
    \nl {$d \gets \sum_{i=1}^{n+1}\lambda_{0,i}d_i(u)$\atcp{get transformation from latent to action space}}\\
    \nl {$\text{\fontfamily{cmtt}\selectfont Sample } z_0 \text{\fontfamily{cmtt}\selectfont\ from RL policy } \pi(z|s)$}\\
    \nl {$a\gets c\odot z_0+d$}
}
}
\end{algorithm}

\section{Additional Implementation Details}
\label{sec:implementation}

We provide our code in the github repository \url{https://github.com/289371298/CEIP} for reference.  

\subsection{CEIP}
\subsubsection{Architecture.} We use slightly different architectures for fetchreach and kitchen/office, because the number of dimensions of the states and actions in fetchreach is much smaller than that in the other two experiments. Moreover, the size of fetchreach is much smaller too. Hence, a smaller network is used for fetchreach to prevent overfitting.

\textbf{Fetchreach.} For each single flow, we use a pair of simple Multi-Layer Perceptron (MLP), one for $c_i(u)$ and the other one for $d_i(u)$. Each network has two hidden layers of width $32$ for each single flow. The number of dimensions for the feature is $20$ (with explicit prior) or $10$ (without explicit prior). For the combination of  flows, we use one fully-connected neural net with two hidden layers of width $32$, which outputs both $\mu$ and $\lambda$. $\mu$ has an additional softplus activation and a $10^{-4}$ offset. If not otherwise specified, all activation functions in this section are ReLU.

\begin{figure}[t]
    \centering
    \includegraphics[width=\linewidth]{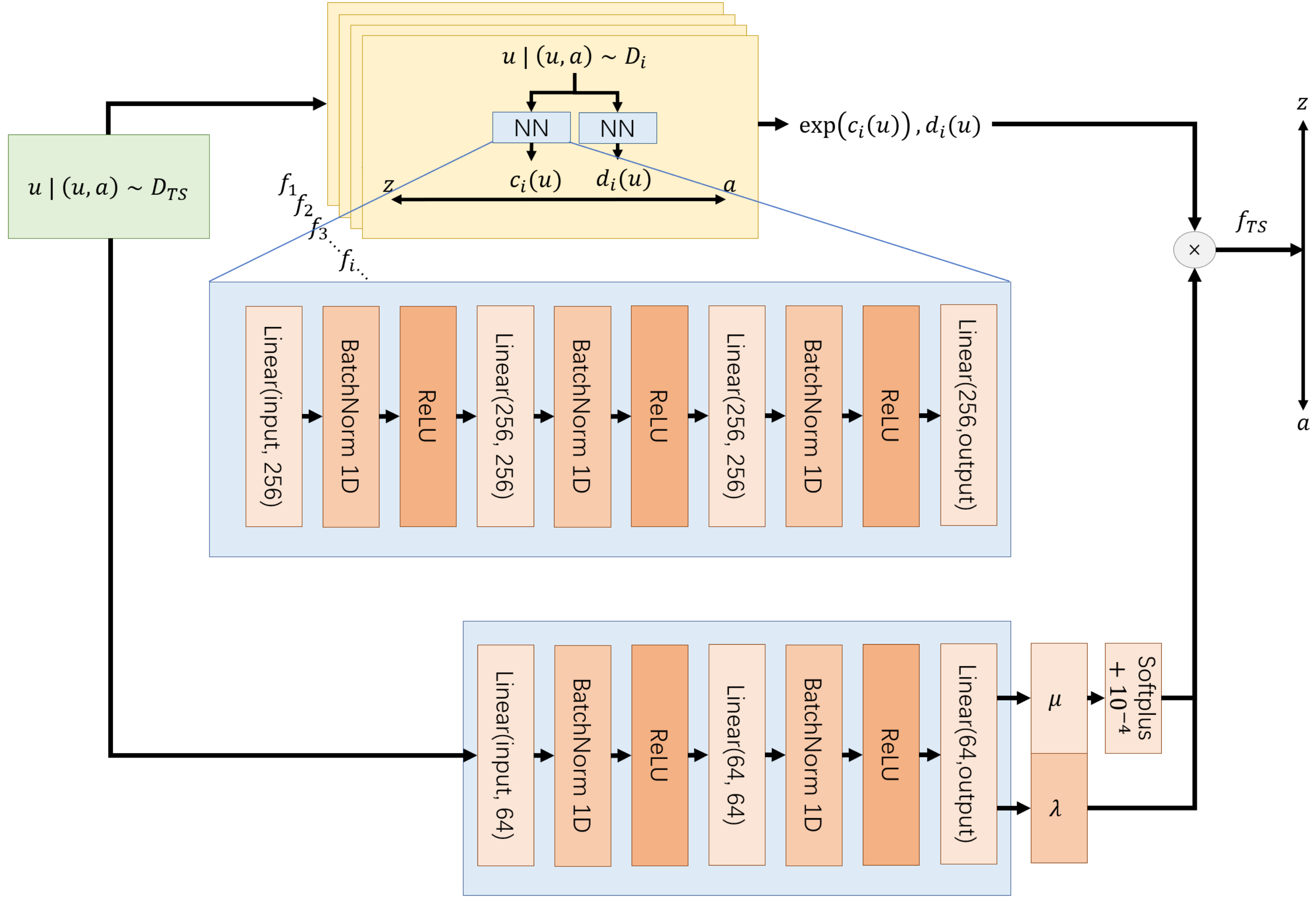} 
    \caption{Illustration of our architecture used for kitchen and office environments.}
    \label{fig:arch_illu}
\end{figure}

\textbf{Kitchen and Office.} The architecture for the kitchen and office environments is roughly the same as that for the fetchreach environment. The difference is that we use three hidden layers of width $256$ for $c_i$ and $d_i$ of each single flow, and that we use two hidden layers of width $64$ for $\mu$ and $\lambda$. Also, we use a batchnorm function before each ReLU activation. See Fig.~\ref{fig:arch_illu} for an illustration.

\subsubsection{Flow Training.} We use the standard flow training method~\cite{Kobyzev2021NFreview} for training the task-agnostic and task-specific single flows $f_1, \dots, f_{n+1}$, which is to maximize the (empirical) log-likelihood
\begin{equation}
\label{eq:flowloss_app}
\begin{aligned}
\max_{f_i}&\ \mathbb{E}_{(u,a)\sim D_i}\log p_a(a|u),\\
\text{where }\log p_a(a|u)=\log p_z(f_i^{-1}(a; u))&+\log \left|\frac{\partial f_i^{-1}(a; u)}{\partial a}\right|
=\log p_z(f_i^{-1}(a; u))-c_i(u)^T\mathbf{1}, \text{ and }\\
f^{-1}_i(a;u)&=z=\frac{a-d_i(u)}{\exp\{c_i(u)\}}.
\end{aligned}
\end{equation}
Here, $c_i(u)\in\mathbb{R}^q$, $d_i(u)\in\mathbb{R}^q$ are trainable deep nets. The $\exp$ function and division are applied elementwise. We use a standard normal distribution over the latent space, i.e., $p_z =  N(0,I)$. Moreover, we use maximization w.r.t.\ $f_i$ to denote maximization w.r.t.\ the parameters of the deep nets $c_i$, $d_i$.
To train the combined flow, we use a similar loss function to Eq.~\eqref{eq:flowloss_app}, i.e., 
\begin{equation}
\max_{f_{\text{TS}}}\ E_{(u,a)\in D_{\text{TS}}}\log p_a(a|u),\text{ where }\log p_a(a|u)=\log p_z(f_{\text{TS}}^{-1}(a; u))+\log \left|\frac{\partial f_{\text{TS}}^{-1}(a; u)}{\partial a}\right|.
\end{equation}
Again, $p_z$ is a standard normal distribution. Here, maximization w.r.t.\ $f_\text{TS}$ denotes maximization w.r.t.\ the parameters of the deep nets $\mu$ and $\lambda$ as shown in Fig.~\ref{fig:arch_illu}.

\textbf{Training Hyperparameters.} To train each single flow, we use $1000$ epochs on each cluster of the task-agnostic dataset $D_1, D_2, \dots, D_{n}$ and task-specific $D_{n+1}$ with a batchsize of $256$. We use the Adam~\cite{KingmaB2014Adam} optimizer with a learning rate of $0.001$ and a gradient clipping at norm $10^{-4}$. For each dataset, we randomly draw $80\%$ of the state-action pairs / transitions (regardless of which trajectory they are in) as the training set and use the rest for validation. We use an early stopping that triggers when the current number of batches fed into the network is greater than $1000$ (fetchreach) or $4000$ (kitchen/office) and the validation loss does not improving during the last $20\%$ of the batches. The model with the lowest loss on the validation set is stored and utilized. Each flow is trained separately and parameters are not shared. Note, we did not optimize the implementation for efficiency, but this can be  accelerated via parallelization.

\subsubsection{Reinforcement Learning.} We use a well-established reliable implementation of RL algorithms, stable-baselines3\footnote{https://stable-baselines3.readthedocs.io/en/master/}, to carry out reinforcement learning. As stable-baselines3 needs a bounded action space, we set the latent (action) space $\mathcal{Z}$ of the RL agent $\pi(z|s)$ to be $[-3,3]$ on each dimension.
\subsection{PARROT}
PARROT can be seen as a  special case of CEIP, where the number of single flows is $1$ and $\mu=1, \lambda=1$. This single flow is trained on all task-agnostic data. The original PARROT does not use an explicit prior or a push-forward technique, which are our contribution in this work. But these components can be added to PARROT in the same way as they are used in our method. For a fair comparison, PARROT uses exactly the same architecture and training paradigm of a single flow as CEIP. 

\subsection{SKiLD}
%
As CEIP, SKiLD also uses an implicit prior. However, different from CEIP which is flow-based, SKiLD uses a VAE-based architecture where the latent space is for an action sequence called ``skill,'' and the decoder of the VAE maps actions from latent space to actual action sequences. In addition, SKiLD uses two implicit priors that take the current state as input and mimic the state-action sequence encoder, one for the entire task-agnostic dataset and the other for the task-specific dataset. To utilize both priors, a discriminator that takes the current state as input is trained. This discriminator approximates the confidence of  the task-specific prior. A reward shaping in the downstream RL stage is then used to drive the agent back to states similar to those in the task-specific dataset, where the discriminator reports higher confidence for the task-specific prior. The reward shaping also encourages the RL agent to form a policy similar to the task-agnostic prior when the confidence is low, and a policy similar to the task-specific prior when the confidence is high. SKiLD does not use an explicit prior or the push-forward technique. However, in a similar spirit, the reward-shaping mechanism encourages the agent to visit states similar to those in the task-specific dataset. We follow the settings described by  SKiLD~\cite{pertsch2021skild}, except for some minor modifications to better adapt SKiLD to the environments. These modifications are discussed next.
{
\interfootnotelinepenalty=10000

We change the configuration mostly for the fetchreach environment, because skills with $10$ steps are too long for the fetchreach environment with $40$ steps in an episode, and because the number of dimensions of the data and the number of datapoints are much smaller than they are in other environments. Therefore, we shorten a skill from $10$ to $3$ steps, and reduce the size of the skill prior and posterior, which are now $3$-layer MLPs with width $32$ instead of the original $5$-layer MLP with width $256$. Also, as the dataset size decreases, we change the number of epochs. For the skill prior, we use a batchsize of $20$, and train for $7500$ cycles over the task-agnostic dataset (for each cycle, one sub-trajectory of length $3$ is sampled for each trajectory).\footnote{See ``RepeatedDataLoader'' in SKiLD's official repository \url{https://github.com/clvrai/spirl/blob/5cd34db7c5e48137550801bf5ac3f8c452590e2c/spirl/utils/pytorch\_utils.py} and \url{https://github.com/clvrai/spirl/blob/5cd34db7c5e48137550801bf5ac3f8c452590e2c/spirl/train.py} for the meaning of ``cycles.''} For the posterior, we use $30$K training cycles over the task-specific dataset. The discriminator is trained for $300$ epochs, sampling both task-agnostic and task-specific datasets. For RL, we use the settings employed for the kitchen environment in the original paper of SKiLD, where the hyperparameter $\alpha=5$ is fixed.
}
For the kitchen and office environments, we follow the original paper and use the same architecture: a $5$-layer MLP with width $128$ for the skill prior and posterior, a linear layer and long-short term memory (LSTM) with width $128$ for the encoder, and a $3$-layer MLP with width $32$ for the discriminator. The training paradigm is almost the same as the one in the original paper, except that the cycles over the task-specific dataset are increased due to a decreased dataset size. We also use exactly the same RL settings as the original paper.



\subsection{FIST}

Conceptually, FIST can be viewed as SKiLD combined with an explicit prior. However, FIST uses pure imitation learning, while SKiLD includes a reinforcement learning phase. Also, different from SKiLD, FIST only uses one prior, which is first trained on the task-agnostic dataset and then fine-tuned on the task-specific dataset. To decide which key is the ``closest'' to the query in dataset retrieval, FIST conducts contrastive learning for the distance metric between states using the InfoNCE loss~\cite{INFONCE}, where the positive sample is the future state (exactly $H$ steps later, where $H$ is the length of a skill) of a state in a dataset, and the negative samples are the future states of other states in the same dataset. This metric is trained on the combined task-agnostic and task-specific data. However, in our experiment we found that using Euclidean distance as the metric suffices to achieve good result.

For FIST, we mostly follow the settings described in the original paper~\cite{Hakhamaneshi2022FIST}, with the exception of some minor modifications. Similar to SKiLD, on fetchreach we use $3$ steps for a skill, and a lighter architecture for the skill prior and posterior network with $2$ hidden layers of width $32$ instead of $5$ hidden layers of width $128$ for the other experiments. We use the settings for the kitchen environment in the original paper for all other experiments. Moreover, the original FIST is occasionally  unstable at the beginning of skill prior training in the office environment, due to an initial loss being too large. To remove this instability, we add  gradient clipping at norm $10^{-3}$ during the first $100$ steps.



\section{Additional Details of Experimental Settings}
\label{sec:app1}
In this section, we  introduce additional details related to the environment settings and dataset settings for each environment.

\begin{figure}[t]
\vspace{0.2cm}
{
\begingroup

\centering
\begin{minipage}[c]{0.32\linewidth}
\subfigure[Fetchreach]{\includegraphics[height=0.8\linewidth]{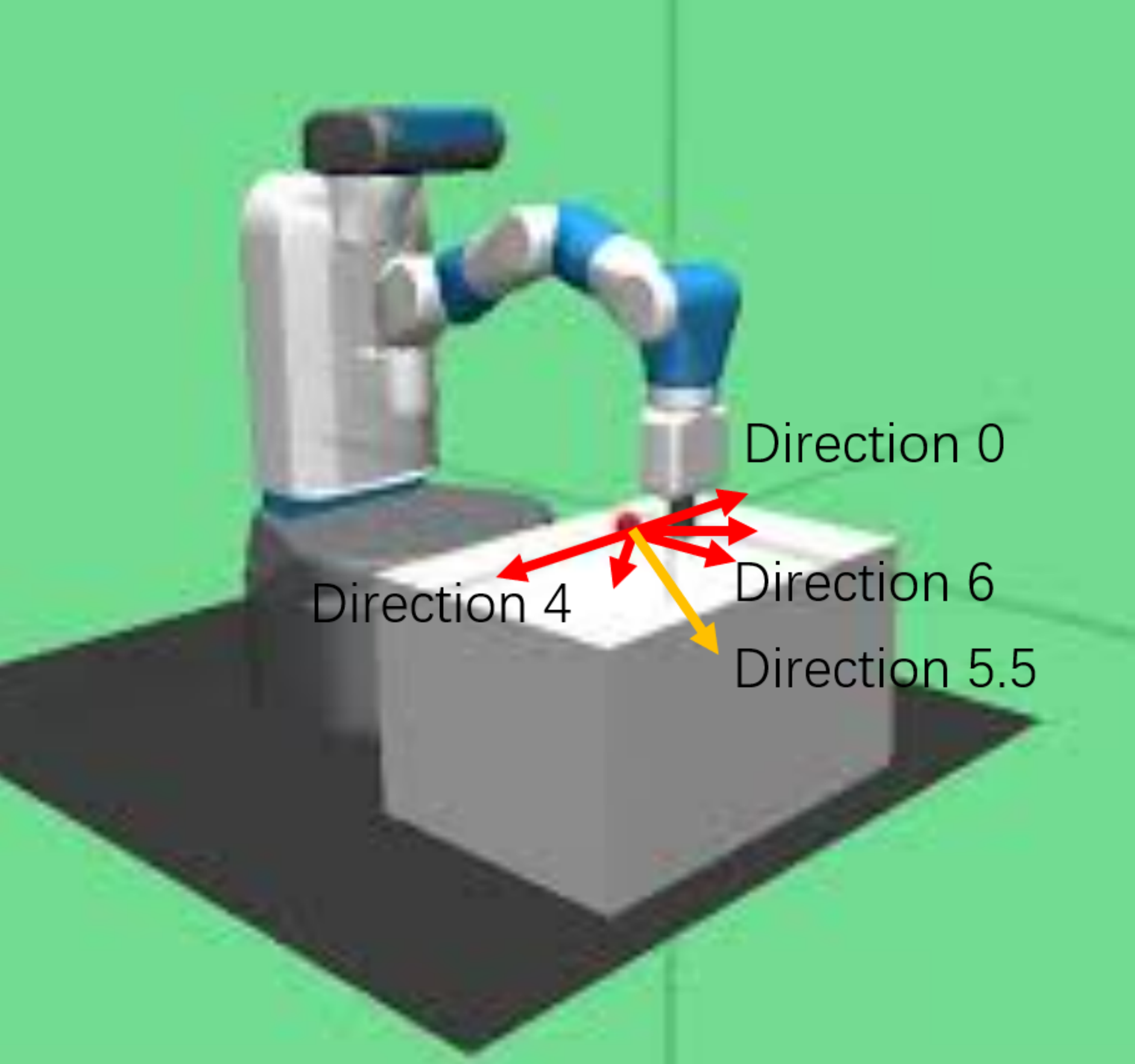}\label{fig:fr01}}

\end{minipage}
\begin{minipage}[c]{0.32\linewidth}
\subfigure[Kitchen]{\includegraphics[height=0.8\linewidth]{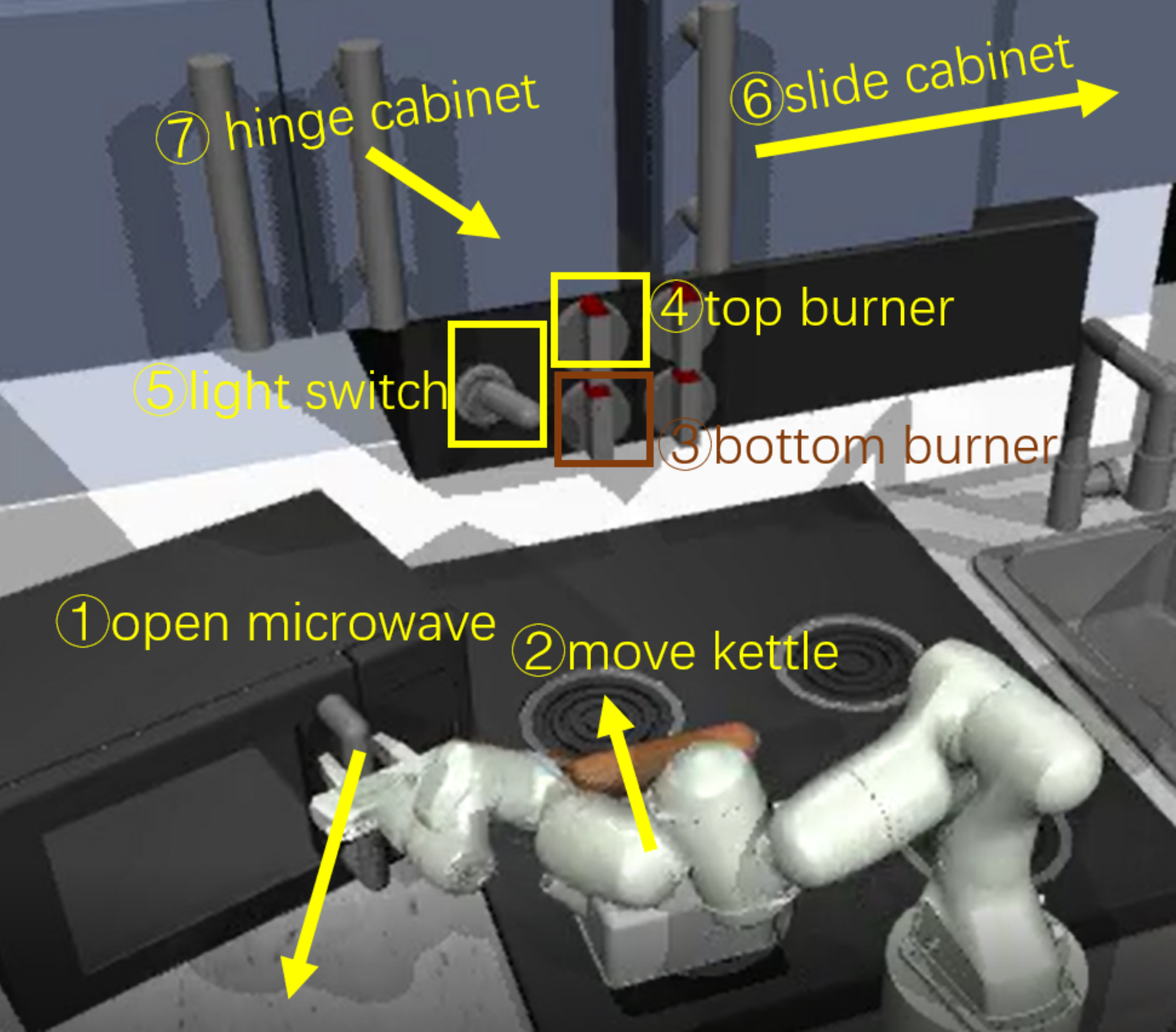}\label{fig:kc00}}

\end{minipage}
\begin{minipage}[c]{0.32\linewidth}
\subfigure[Office]{\includegraphics[height=0.8\linewidth]{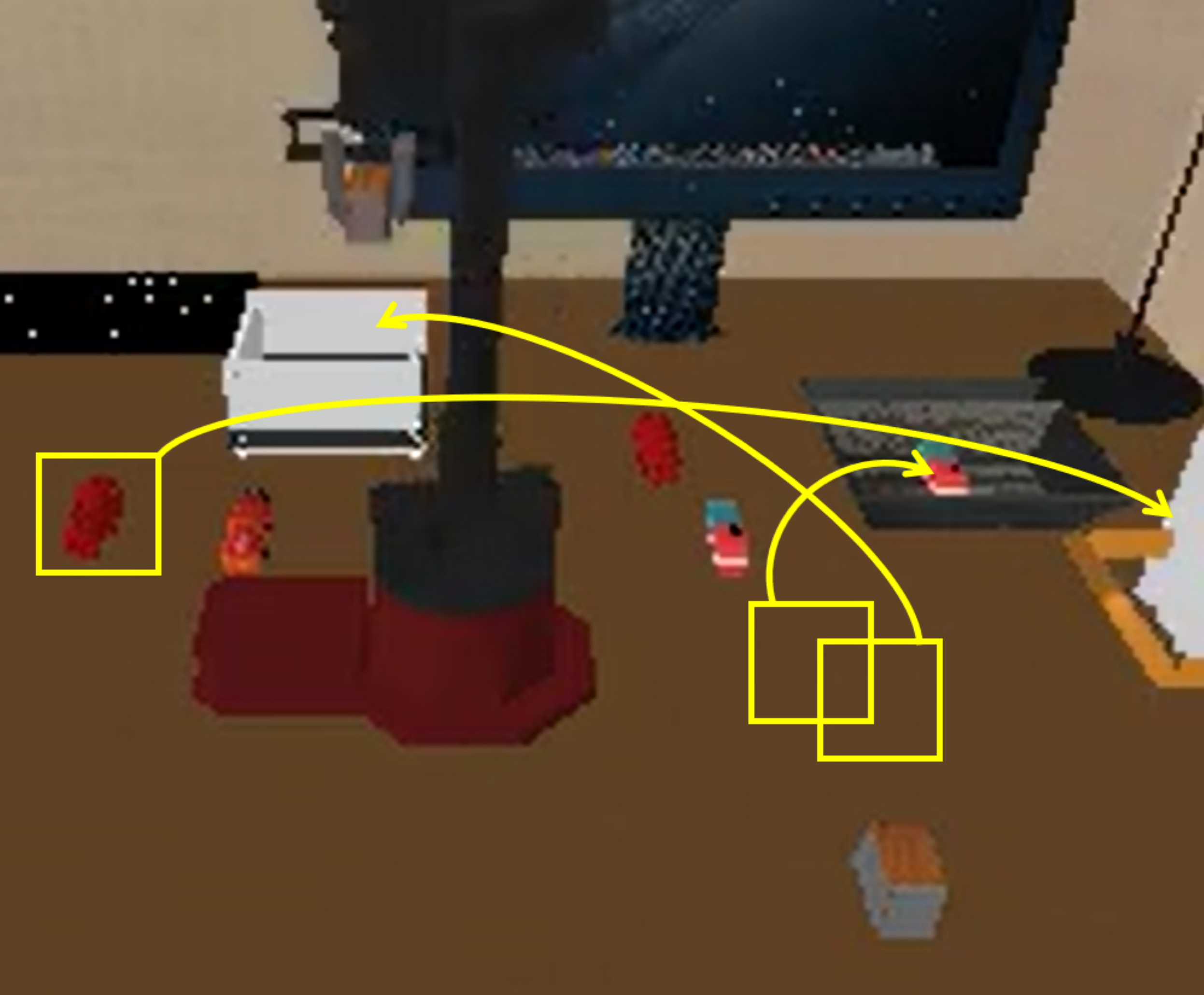}\label{fig:office00}}
\end{minipage}
\vspace{-0.2cm}
\caption{Illustration of each environment. For fetchreach, the task-agnostic dataset consists of demonstrations which move the gripper in the directions of the red arrows, and the task-specific dataset contains demonstrations which move the gripper in the directions of the yellow arrows. For the kitchen environment, the agent needs to complete four out of seven tasks mapped on the picture in the correct order. For the office environment, the agent needs to put items in the container as illustrated in the figure, using the correct order.
}
\label{fig:env}

\endgroup
}
\vspace{-0.3cm}
\end{figure}

\subsection{Fetchreach}
\begin{figure}[t]
\vspace{0.2cm}
{
\begingroup

\centering
\begin{minipage}[c]{0.32\linewidth}
\subfigure[No randomization]{\includegraphics[height=0.8\linewidth]{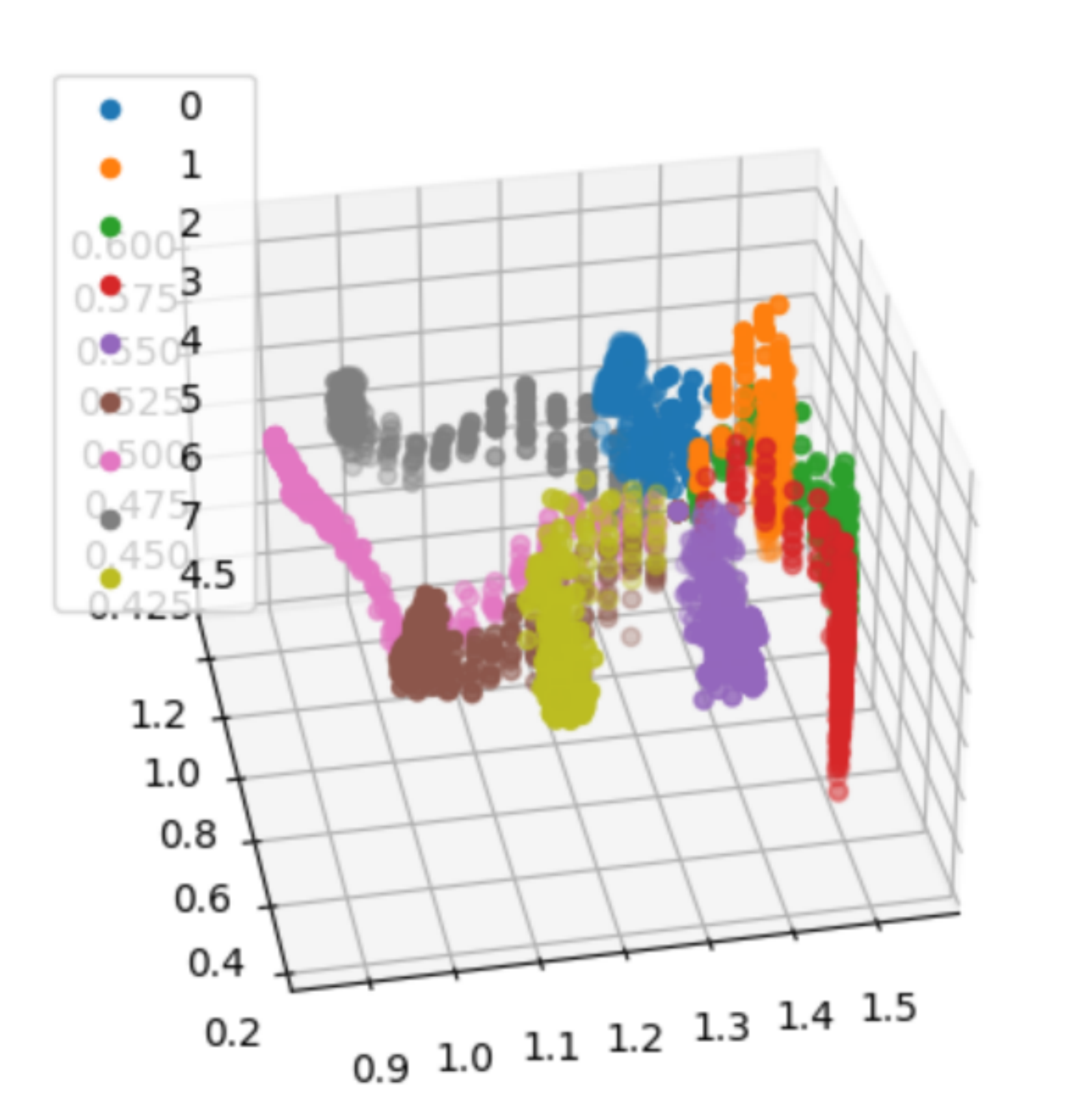}}
\end{minipage}
\begin{minipage}[c]{0.32\linewidth}
\subfigure[Random action at each step]{\includegraphics[height=0.8\linewidth]{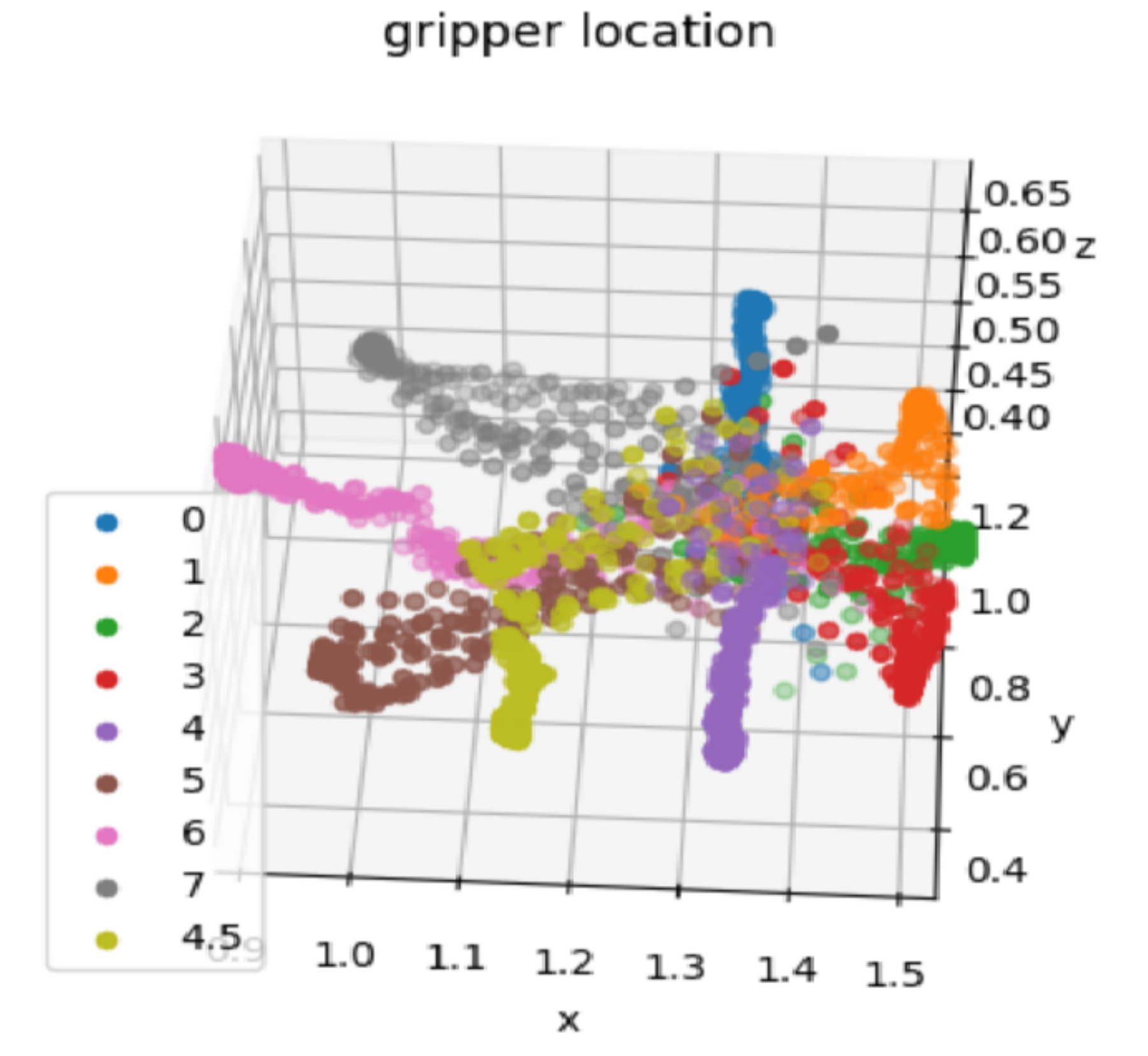}}
\end{minipage}
\begin{minipage}[c]{0.32\linewidth}
\subfigure[Our randomization]{\includegraphics[height=0.8\linewidth]{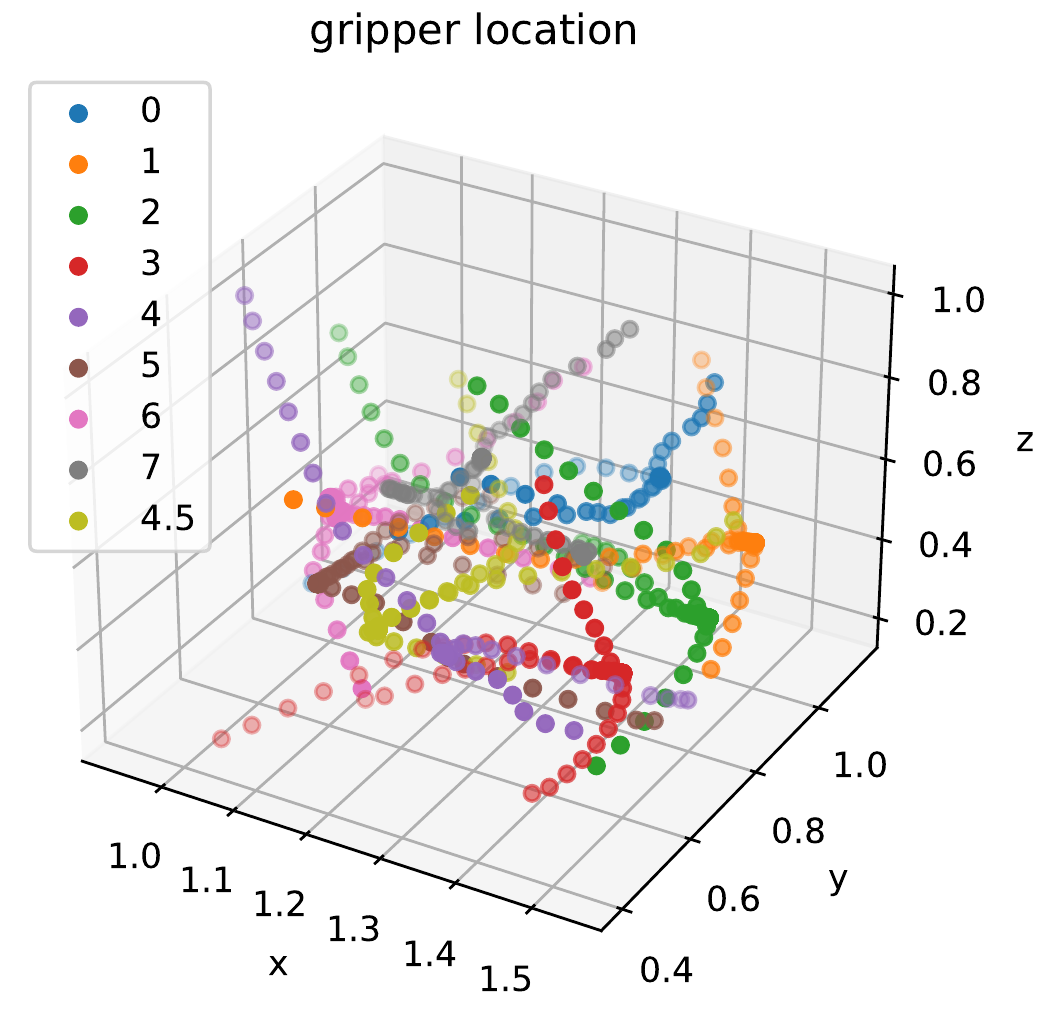}}
\end{minipage}
\vspace{-0.2cm}
\caption{Illustration of expert trajectories with no start randomization (left), sampling a randomized action for $10$ steps (middle), and our way of randomization (right). Note that our randomization greatly increases the variation of the trajectories.}
\label{fig:fr00}
\endgroup
}
\vspace{-0.3cm}
\end{figure}

\textbf{Environment Settings.} In our version of fetchreach (illustrated in Fig.~\ref{fig:fr01}), we need to train a robot arm to move its gripper to a given but unknown location as quickly as possible, and stay there once the goal is reached. The state is $10$-dimensional, with the first three dimensions describing the current location of the gripper. The other dimensions are the openness of the gripper and the current velocity. For each of the $40$ steps, the agent needs to output a $4$-dimensional action $a\in [-1, 1]^{4}$, where the first three dimensions are the direction which the gripper is moving to and the fourth is the openness of the gripper (unused in this experiment). The agent receives a reward of $0$ if the Euclidean distance between the gripper and the target is smaller than $0.05$, and $-1$ otherwise. A perfect agent should achieve a reward of around $-10$. The goal denoted as direction $d$ (e.g., direction $4.5$) is generated by first assigning a direction $d\in[0, 8)$, then selecting the goal with the Euclidean distance being $0.3$ away and the azimuth being $\frac{d\pi}{4}$, and finally applying a uniform noise of $U[-0.015, 0.015]$ on each of the three dimensions. In order to test the robustness of the algorithms and increase difficulty, before each episode begins, we first sample a random action from a normal distribution, and then let the agent execute the action for $x$ steps, where $x\sim U[5, 20]$. This greatly increases the variety of the trajectories, as shown in Fig.~\ref{fig:fr00}.

\textbf{Dataset Settings.} The dataset is acquired by first training an RL agent with soft actor critic (SAC) which receives the negative current Euclidean distance as a reward until convergence, and then sampling trajectories on the trained RL agent. For each direction in $\{0,1,2,\dots,7\}$, $40$ trajectories ($1600$ steps) are sampled. For each direction in $\{4.5,5.5,6.5,7.5\}$, $4$ trajectories ($160$ steps) are sampled.


\subsection{Kitchen-SKiLD}
\textbf{Environment Settings}. We adopt the same setting as SKiLD~\cite{pertsch2021skild} and FIST~\cite{Hakhamaneshi2022FIST}, where an agent needs to finish four out of seven tasks in the correct order. The tasks are: open the microwave, move the kettle, turn on the light switch, turn on the bottom burner, turn on the top burner, slide the right cabinet, and hinge the left cabinet. The agent needs to complete all four tasks within $280$ timesteps, and a $+1$ reward is given when one task is completed. The state is $60$-dimensional, where the first $9$ dimensions describe the current location of the robot, the next $21$ dimensions describe the current object location (the unrelated objects will be zeroed out), and the rest are constant and describe the initial location of each object. The action $a$ is $9$-dimensional where $a\in [-1, 1]^9$, which controls the joints of the arm. A uniform noise of $[-0.1, 0.1]$ is applied to the observation of the robot in every step.\footnote{See SKiLD's official repository \url{https://github.com/kpertsch/d4rl/blob/master/d4rl/kitchen/adept\_envs/franka/robot/franka\_robot.py} for details.}


\textbf{Dataset Settings.} We use exactly the same task-agnostic dataset as SKiLD, which includes $33$ different task sequences with a total of $136950$ state-action pairs generated by `relay policy learning'~\cite{Gupta2019RelayPL}. We choose the first trajectory from the task-specific dataset of SKiLD as our task-specific dataset, which includes $214$ state-action pairs for task A and $262$ state-action pairs for task B. Task A is ``move the kettle, turn on the bottom burner, turn on the top burner, and slide the right cabinet;'' task B is ``open the microwave, turn on the light switch, slide the right cabinet, and hinge the left cabinet.''

\subsection{Kitchen-FIST}

\textbf{Dataset Settings.} We use exactly the same task-agnostic dataset as FIST~\cite{Hakhamaneshi2022FIST}. There are four pairs of task-agnostic and task-specific datasets, which are illustrated in  Table~\ref{tab:task}.

\begin{table}[t]
\setlength{\tabcolsep}{4pt}
    \centering
    \begin{tabular}{cccc}\toprule
        Task & Subtask Missing & Target Task & Dataset Size \\\midrule
        A & Top Burner & Microwave, Kettle, Top Burner, Light Switch & 66823 / 210 \\
        B & Microwave & Microwave, Bottom Burner, Light Switch, Slide Cabinet & 52898 / 200 \\
        C & Kettle & Microwave, Kettle, Slide Cabinet, Hinge Cabinet & 53576 / 246 \\
        D & Slide Cabinet & Microwave, Kettle, Slide Cabinet, Hinge Cabinet & 45267 / 246 \\ \bottomrule 
    \end{tabular}
    \caption{List of four different settings in Kitchen-FIST. The dataset size format is ``task-agnostic / task-specific.'' The dataset size is counted in state-action pairs.}
    \label{tab:task}
\end{table}

\subsection{Office}

\textbf{Environment Settings}. We adopt the settings from SKiLD, where we need to train a robot arm to put three out of seven items into three containers (illustrated in Fig.~\ref{fig:office00}), which are the two trays and one drawer. The robot arm receives a $97$-dimensional state and has $8$ dimensions of actions which control the position and angle of the gripper, as well as the continuous gripper actuation. There are $8$ subtasks in the experiments: pick up the first/second item, drop the first/second item in the right place, open the drawer, pick up the third item, drop the third item correctly, and close the drawer. The agent will receive a $+1$ reward upon completion of each subtask. The episode length is at most $350$ steps, and will finish as soon as all the tasks are finished.


\textbf{Dataset Settings.} We use the same task-agnostic dataset and a subset of the task-specific dataset. We use a subset of the task-specific dataset because SKiLD,  CEIP, PARROT+TS, and FIST can all solve the problem very well with the whole task-specific dataset, making it hard to compare. The task-agnostic dataset contains $2417$ trajectories from randomly-generated tasks, which has $7\times 6\times 5=210$ possibilities and contains $456033$ state-action pairs. The task-specific dataset contains $5$ trajectories with $1155$ state-action pairs.

\section{Additional Experimental Results}
\label{sec:extraexp}

This section includes additional experimental results, which are ablation studies and auxiliary metrics that help to better understand the properties of different methods. See the beginning of the Appendix for a summary of the key findings. Please also see our supplementary material for sample videos of each trained method on the kitchen and office environments.

\subsection{Abbreviations for Ablation Tests}
In our experiments, we test multiple variants of CEIP and PARROT for a more thorough ablation. To more easily differentiate the variants of both methods with different components, we will use abbreviations  as listed in Table~\ref{tab:abbrCEIP} and Table~\ref{tab:abbrPARROT}. 

\begin{table}
   \setlength{\tabcolsep}{1pt}
    \centering
    
    \begin{tabular}{cccc}\toprule
        Method & \ \shortstack{Task-specific\\ flow}\ &\ \shortstack{Explicit\\ prior}\ & \ \shortstack{Push-\\forward}\ \\ \midrule
        CEIP & & & \\
        CEIP+EX & & \checkmark &  \\
        CEIP+EX+forward & & \checkmark & \checkmark  \\
        CEIP+TS & \checkmark & & \\
        CEIP+TS+EX & \checkmark & \checkmark & \\
        CEIP+TS+EX+forward & \checkmark & \checkmark & \checkmark \\
        
        \bottomrule
    \end{tabular}
    \caption{Abbreviations for variants of CEIP. See Fig.~\ref{fig:fetchreach_plot_ablation_ours_dataset} for difference between CEIP, CEIP+2way, and CEIP+4way; the latter two only appear in the fetchreach ablation.}
    \label{tab:abbrCEIP}
\end{table}

\begin{table}
   \setlength{\tabcolsep}{1pt}
    \centering
    
    \begin{tabular}{ccccc}\toprule
        Method & \shortstack{Use \\ task-agnostic data} & \shortstack{Use \\ task-specific data} & \ \shortstack{Explicit \\ prior}\  & \ \shortstack{Push-\\forward}\  \\ \midrule
        PARROT+TA & \checkmark & & & \\
        PARROT+TS & & \checkmark & & \\
        PARROT+(TS+TA) & \checkmark & \checkmark & & \\
        PARROT+TA+EX & \checkmark & & \checkmark & \\
        PARROT+TS+EX & & \checkmark & \checkmark & \\
        PARROT+(TS+TA)+EX & \checkmark & \checkmark & \checkmark & \\
        PARROT+TA+EX+forward & \checkmark & & \checkmark & \checkmark \\
        PARROT+TS+EX+forward & &\checkmark &\checkmark &\checkmark \\
        PARROT+(TS+TA)+EX+forward & \checkmark &  \checkmark& \checkmark & \checkmark \\
        PARROT+2way+TS & two most related directions & \checkmark & & \\
        PARROT+4way+TS & four most related directions & \checkmark & & \\
        PARROT+2way & two most related directions & & & \\
        \bottomrule
    \end{tabular}
    \caption{Abbreviations for variants of PARROT. All variants of PARROT only use a single flow for all data, which is the key difference between CEIP and PARROT with explicit prior. ``2way'' and ``4way'' only appear in the fetchreach environment where there are $8$ directions in the task-agnostic dataset.}
    \label{tab:abbrPARROT}
\end{table}

\subsection{Fetchreach}
\label{sec:expFR}
\begin{figure}[t]
    \centering
    \subfigure[Direction 4.5]{ 
    \begin{minipage}[b]{0.23\linewidth}
    \includegraphics[width=\linewidth]{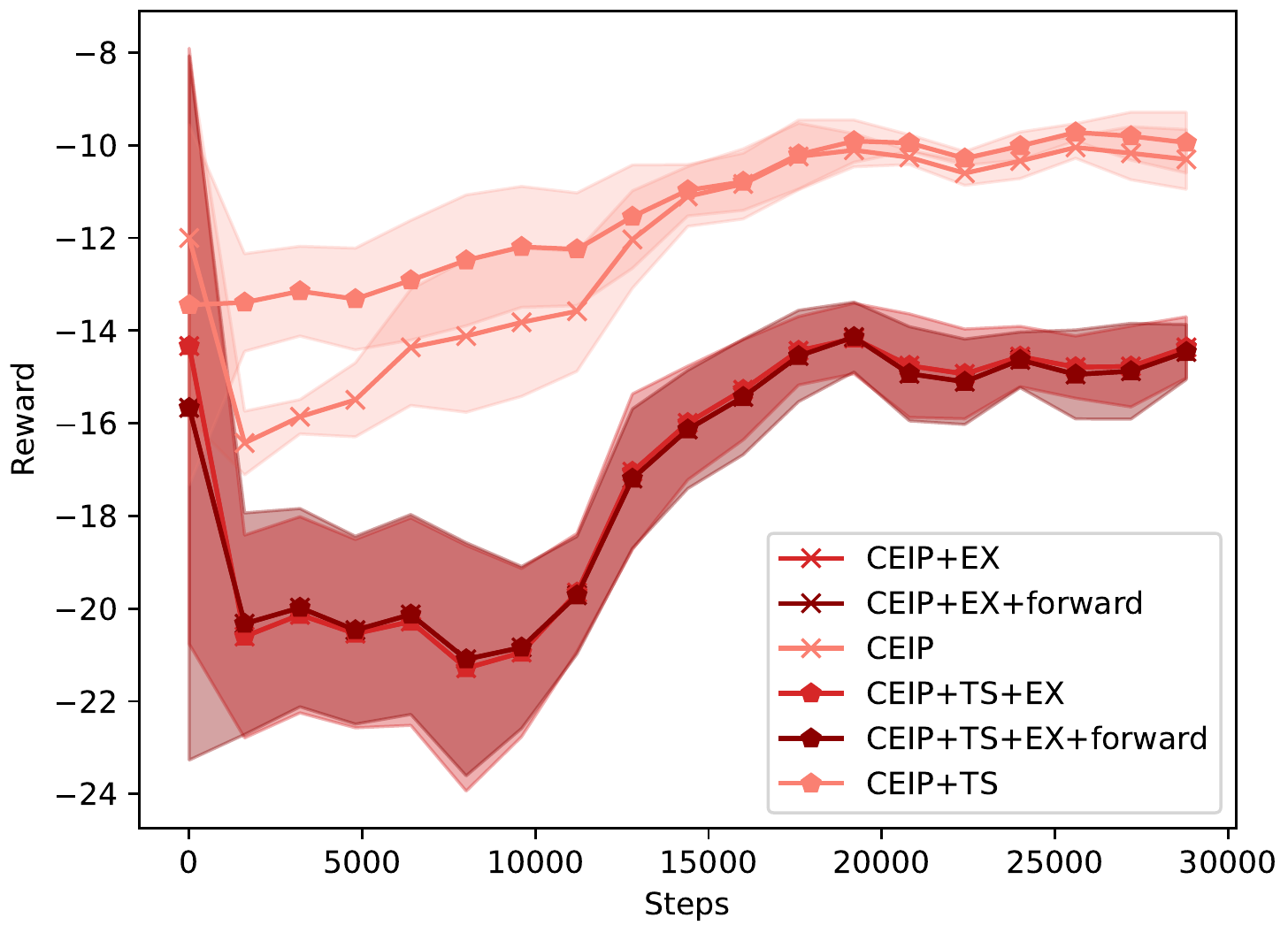}
    \end{minipage}
    }
    \subfigure[Direction 5.5]{
    \begin{minipage}[b]{0.23\linewidth}
    \includegraphics[width=\linewidth]{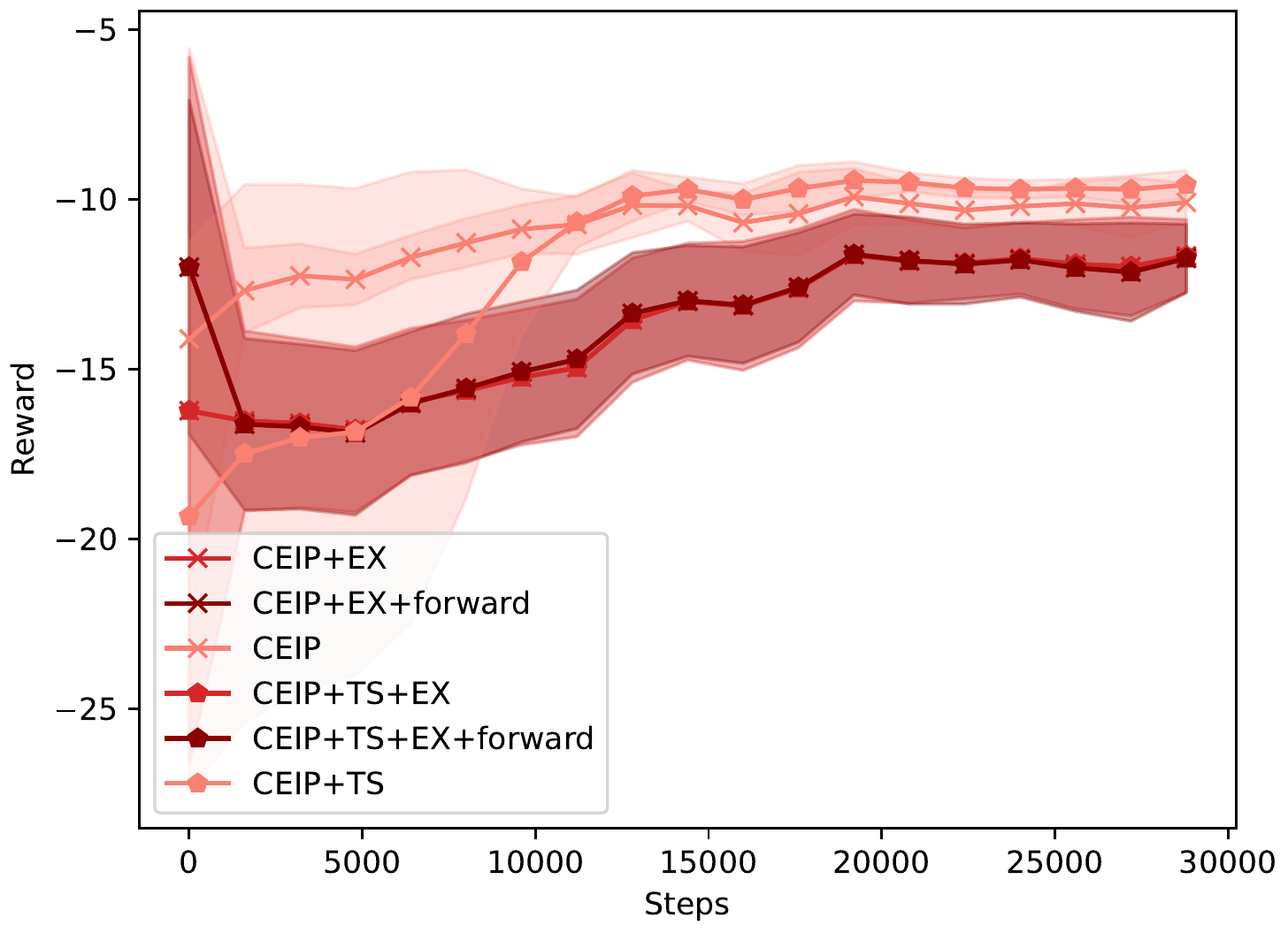}
    \end{minipage}
    }
    \subfigure[Direction 6.5]{ 
    \begin{minipage}[b]{0.23\linewidth}
    \includegraphics[width=\linewidth]{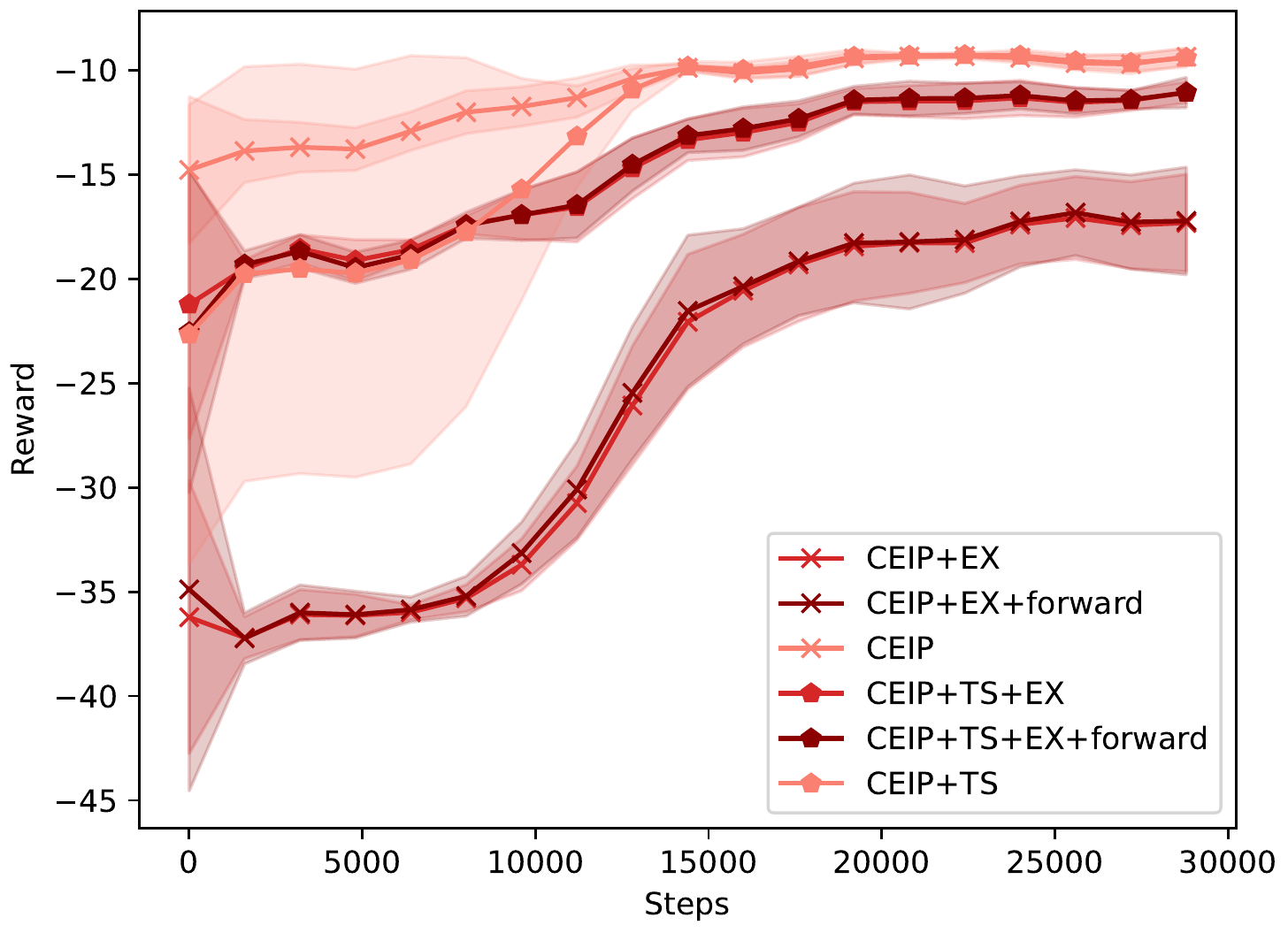}
    \end{minipage}
    }
    \subfigure[Direction 7.5]{
    \begin{minipage}[b]{0.23\linewidth}
    \includegraphics[width=\linewidth]{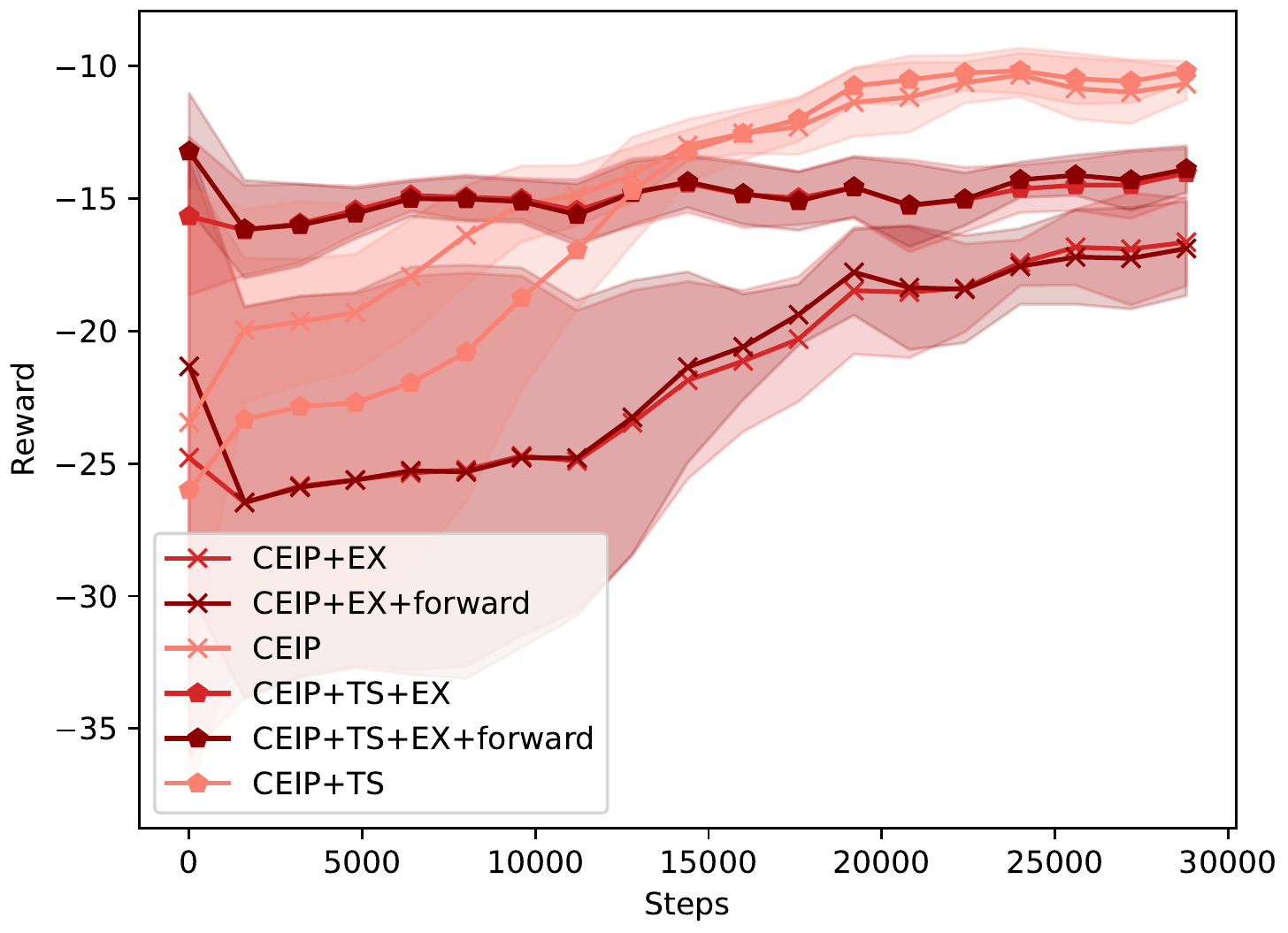}
    \end{minipage}
    }
    \caption{Ablation on architecture and components of our method. We observe that the reward actually grows slower when using the task-specific single flow, explicit prior, and push-forward technique, likely because the training complexity is unnecessarily increased.}
    \label{fig:fetchreach_plot_ablation_ours_arch}
\end{figure}

\textbf{Ablation on components of our method.} Fig.~\ref{fig:fetchreach_plot_ablation_ours_arch} shows the ablation study on different components of our method. Results for our method show that using an explicit prior and the push-forward technique slows down the reward growth during RL training, if applied on a relatively easy and short-horizon environment. This is likely because we unnecessarily add training complexity to an environment with a relatively easy task, short horizon, and well-clustered task-agnostic datasets. However, our method with those components still works better than many baselines and  performs well given more steps.

\begin{figure}[t]
    \centering
    \subfigure[Direction 4.5]{ 
    \begin{minipage}[b]{0.23\linewidth}
    \includegraphics[width=\linewidth]{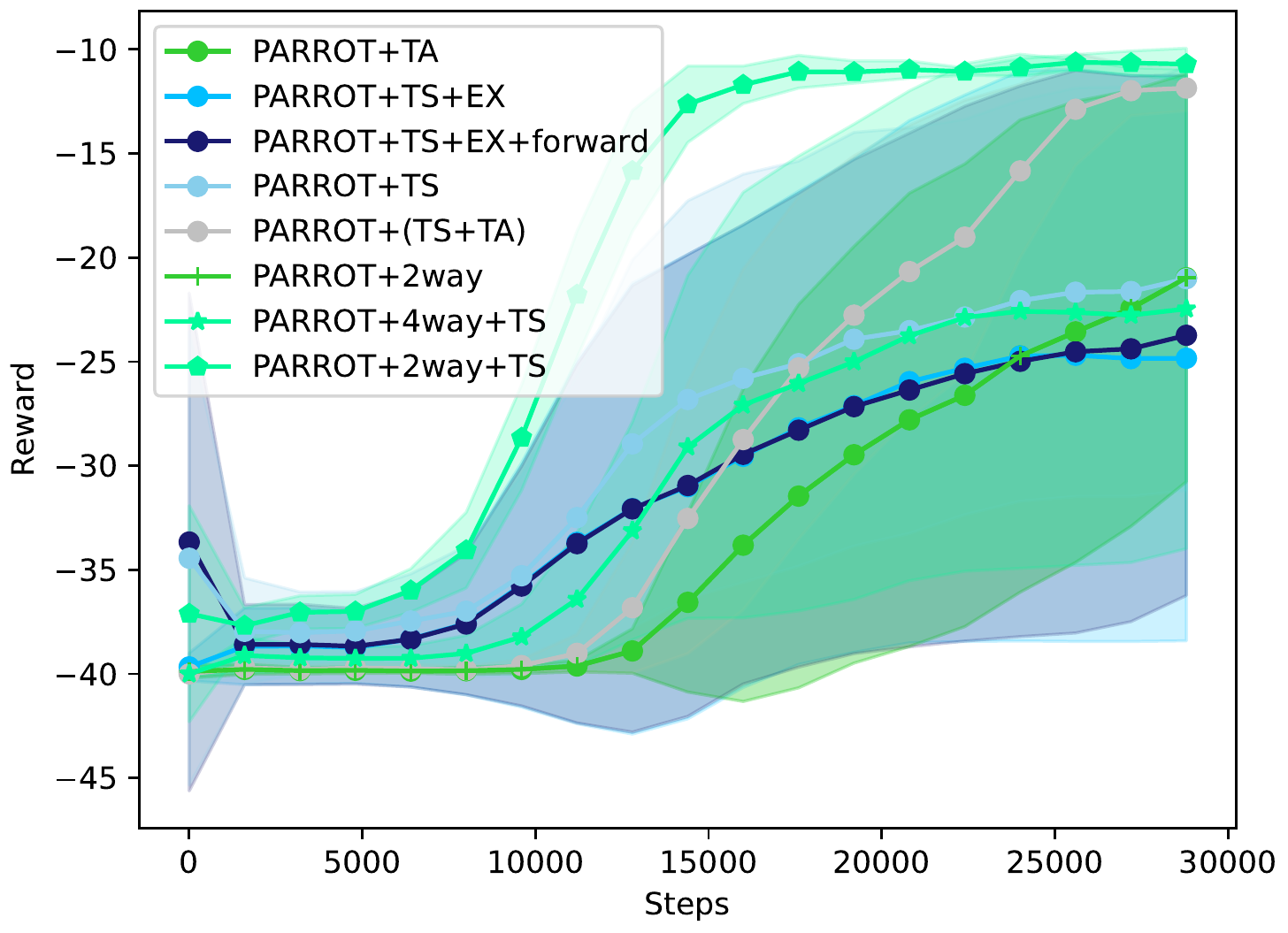}
    \end{minipage}
    }
    \subfigure[Direction 5.5]{
    \begin{minipage}[b]{0.23\linewidth}
    \includegraphics[width=\linewidth]{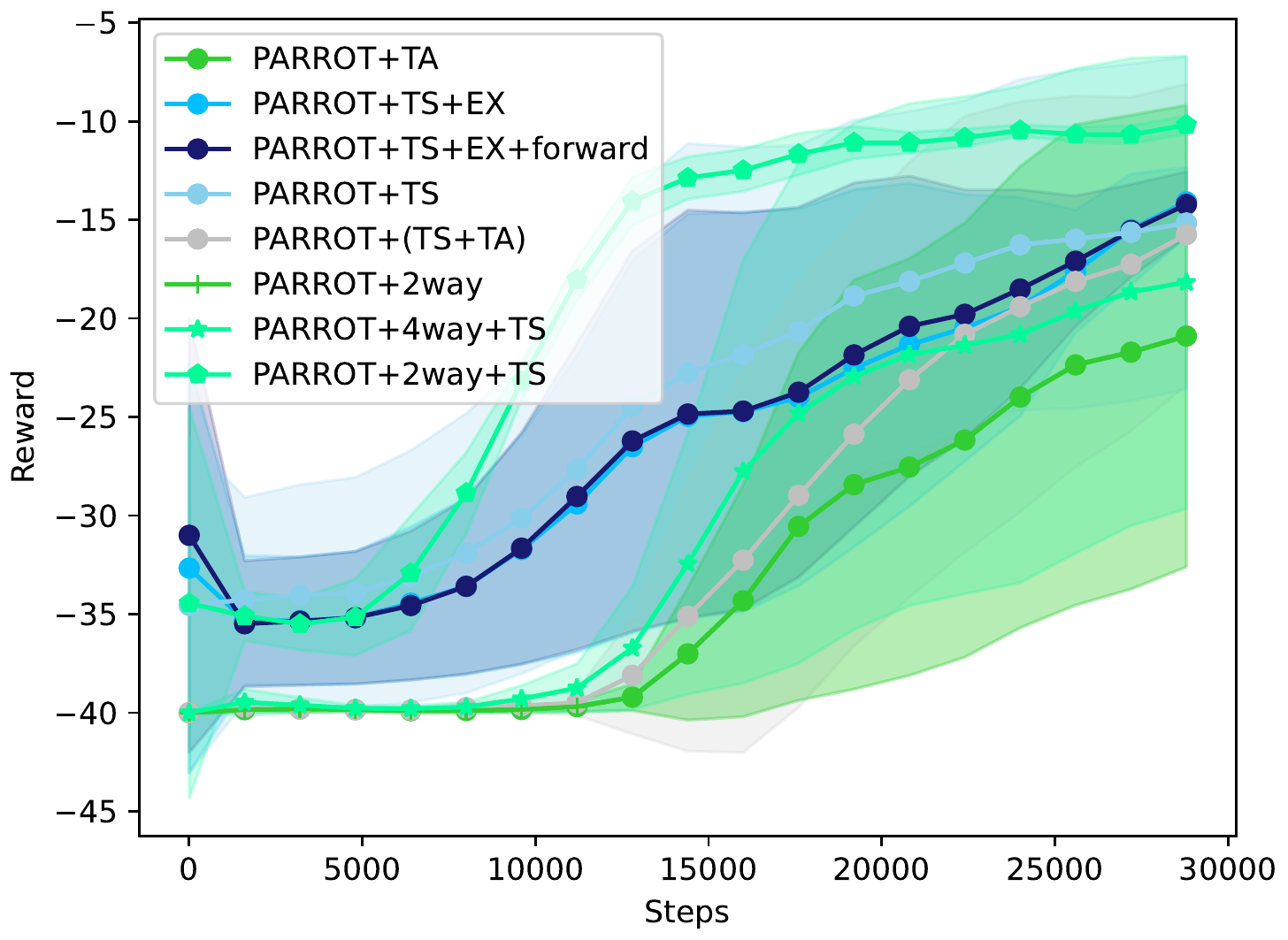}
    \end{minipage}
    }
    \subfigure[Direction 6.5]{
    \begin{minipage}[b]{0.23\linewidth}
    \includegraphics[width=\linewidth]{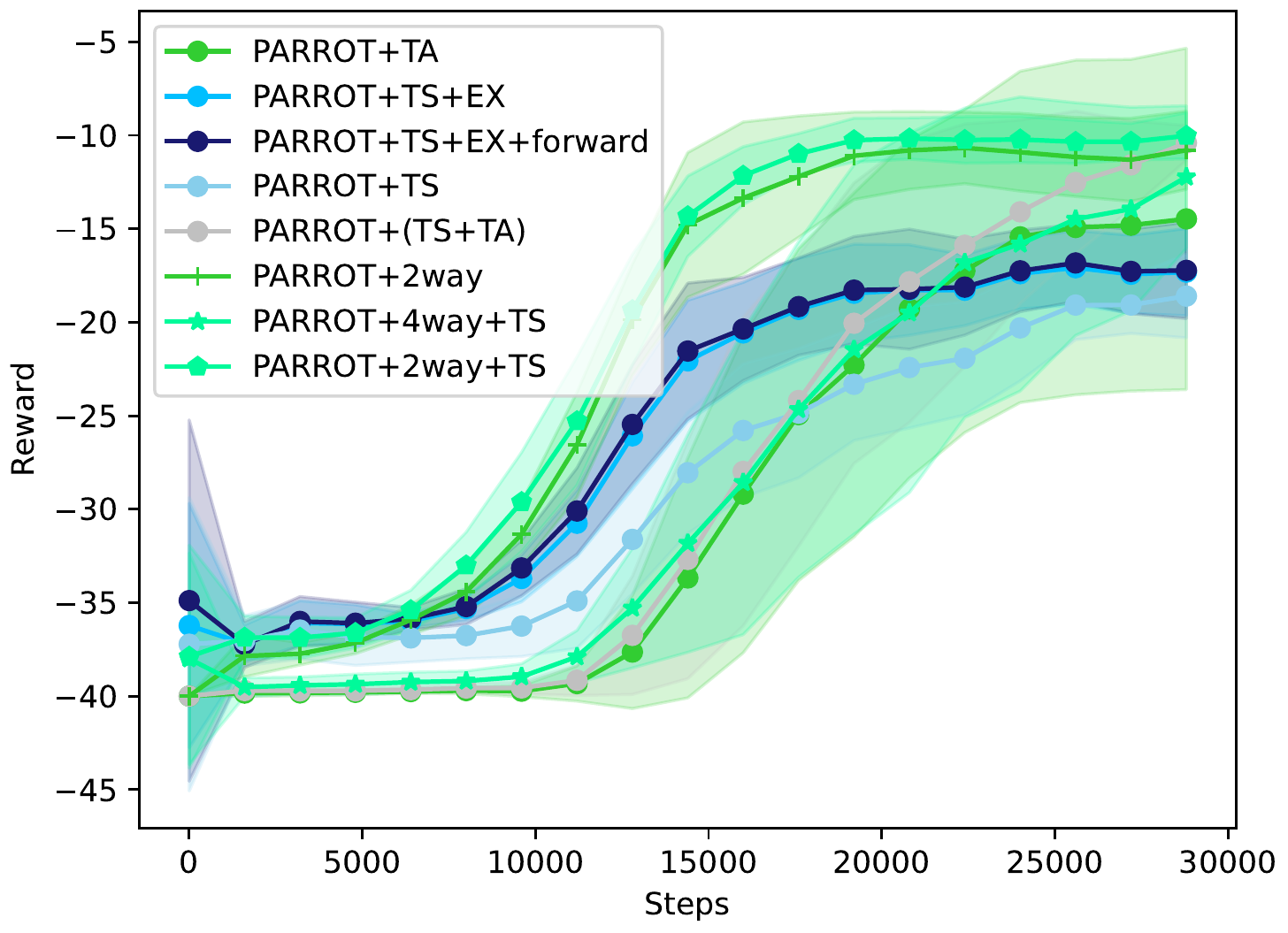}
    \end{minipage}
    }
    \subfigure[Direction 7.5]{
    \begin{minipage}[b]{0.23\linewidth}
    \includegraphics[width=\linewidth]{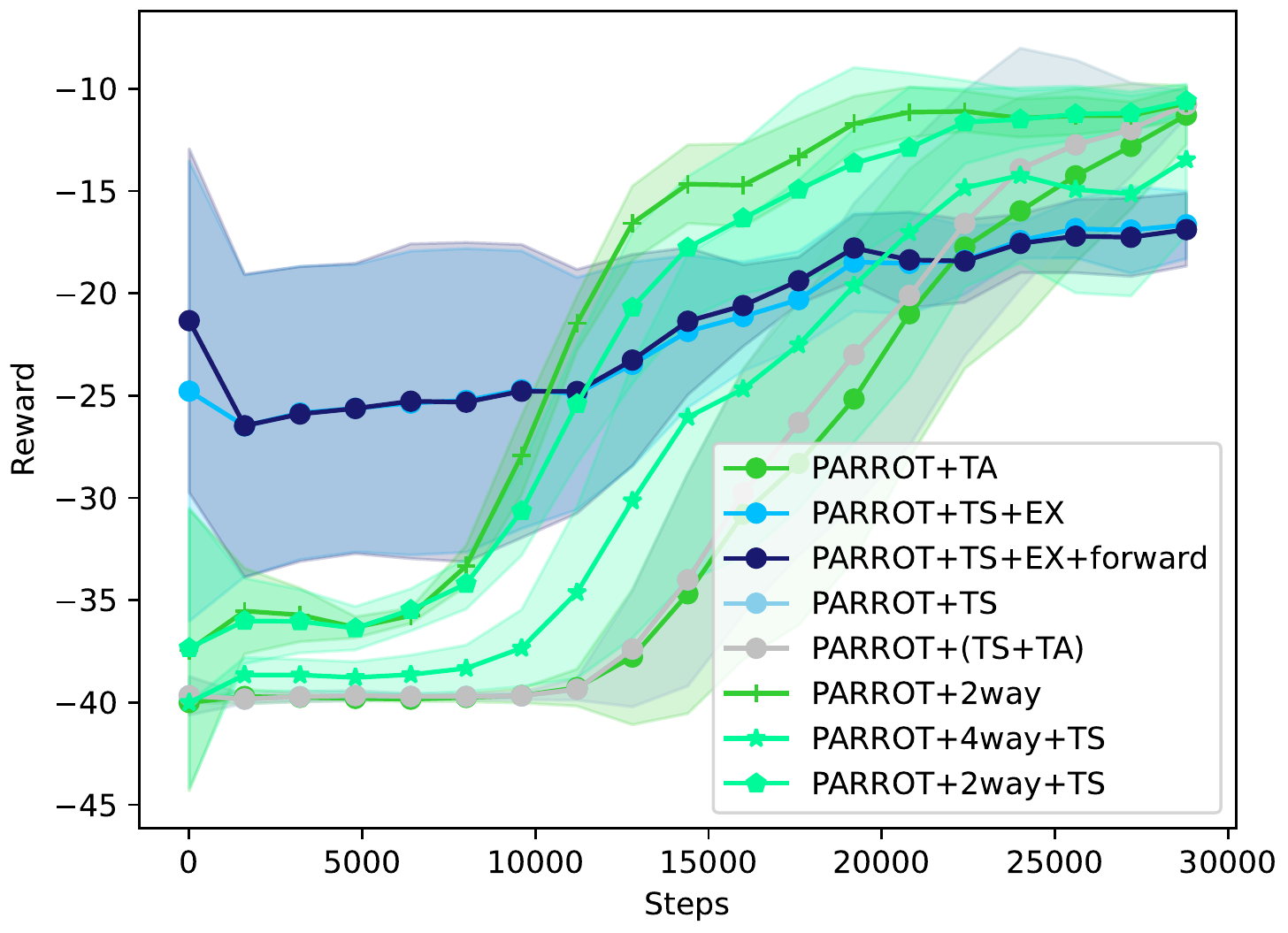}
    \end{minipage}
    }
    \caption{Ablation on the data used for PARROT. ``2way'' and ``4way'' mean that we feed the two/four directions in the task-agnostic dataset that are closest to the target direction (e.g., if the target direction is $4.5$, we refer to data from directions $4$ and $5$ as ``2way,'' and data from directions $3,4,5,6$ as ``4way''). Note, PARROT with the task-specific data and ``2way'' data is significantly better than other variants of PARROT. However, PARROT is only improved when such data are selected manually, while our method can automatically combine the flows and select useful priors. Also, PARROT with ``2way'' data from the task-agnostic dataset but without task-specific dataset works well, but it is unstable, which emphasizes the importance of the task-specific dataset even if it is small.}
    \label{fig:fetchreach_plot_ablation_parrot_dataset}
\end{figure}
\textbf{Ablation on components and data relevance in PARROT.} To better understand the properties of PARROT, we ablate the data used when training PARROT (see Fig.~\ref{fig:fetchreach_plot_ablation_parrot_dataset}). We select a subset of the task-agnostic data that is more relevant to the task-specific dataset, and study how data in the task-agnostic dataset with different levels of relevance to the downstream task affect results. We also test the effect of the explicit prior and the push-forward technique using task-specific data only. The results can be summarized as follows: 1) using the explicit prior and the push-forward technique  slows down the reward growth during RL training, if applied on a relatively easy and short-horizon environment; 2) selecting more relevant data for PARROT is an effective way to improve PARROT, which supports our motivation to combine the flows to select the most useful prior.

\textbf{Illustration of trajectories.} To validate the effect of the implicit prior, Fig.~\ref{fig:traj} shows the trajectories generated by our method and PARROT without any RL training. We clearly observe that the trajectories generated by PARROT become more accurate when the data are more related (from left to right), which is achieved by manual selection but can be done automatically in CEIP. Our method improves when more flows are used (from right to left), as more flows increase expressivity. 
 
{
\begingroup
\begin{figure}[t]
\begin{minipage}[c]{0.32\linewidth}
\subfigure[PARROT+(TS+TA)]{\includegraphics[width=\linewidth]{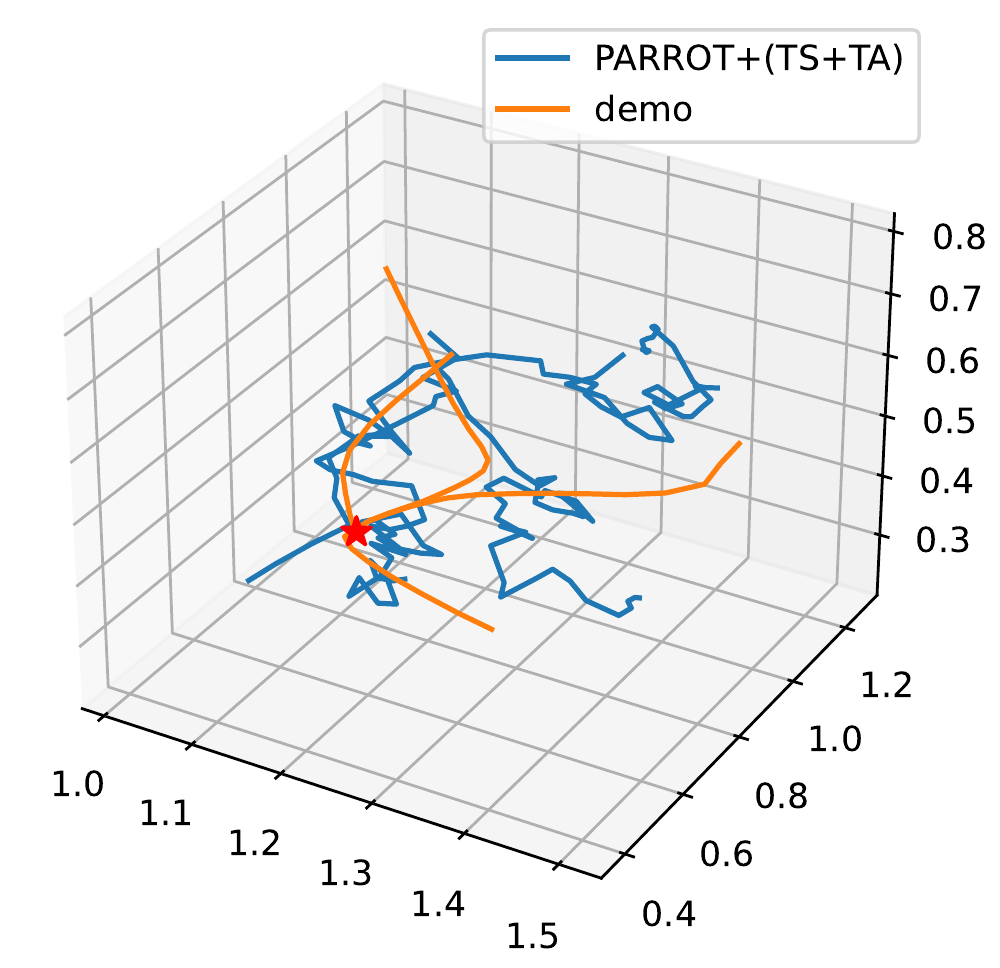}}
\subfigure[ours]{\includegraphics[width=\linewidth]{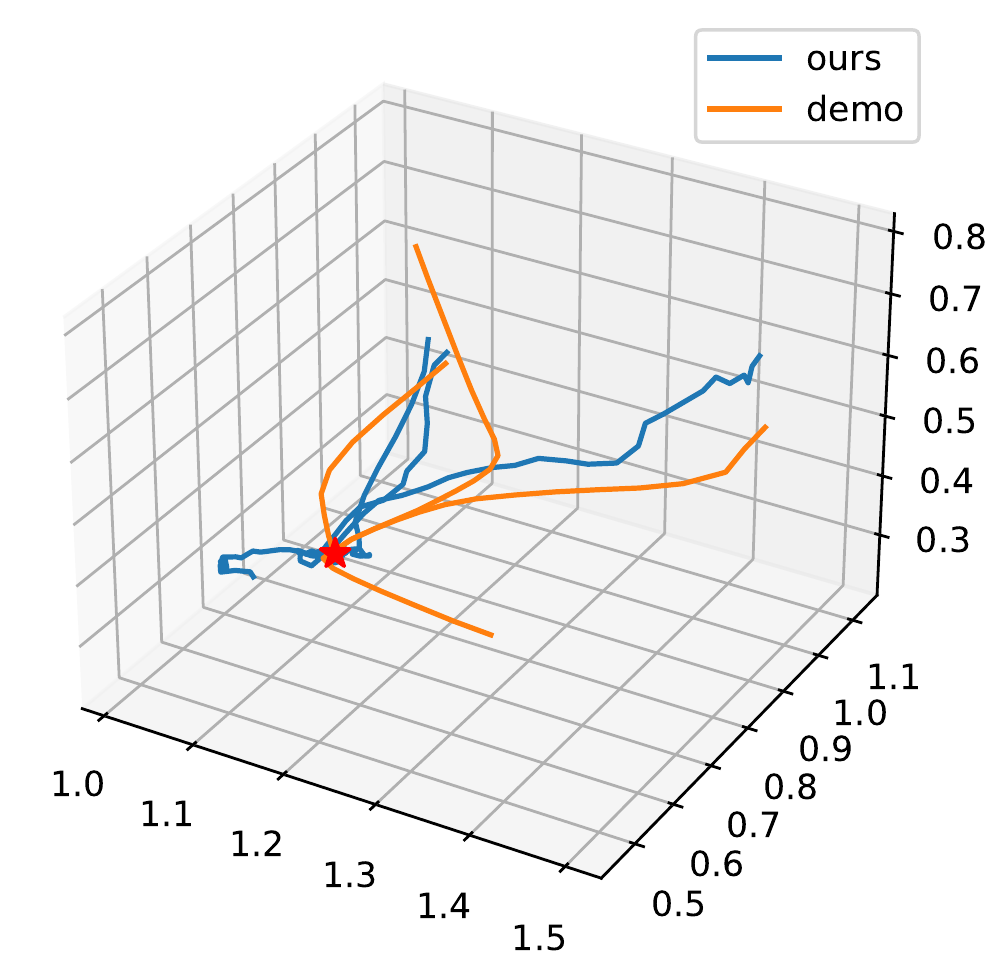}}
\end{minipage}
\begin{minipage}[c]{0.32\linewidth}
\subfigure[PARROT+4way+TS]{\includegraphics[width=\linewidth]{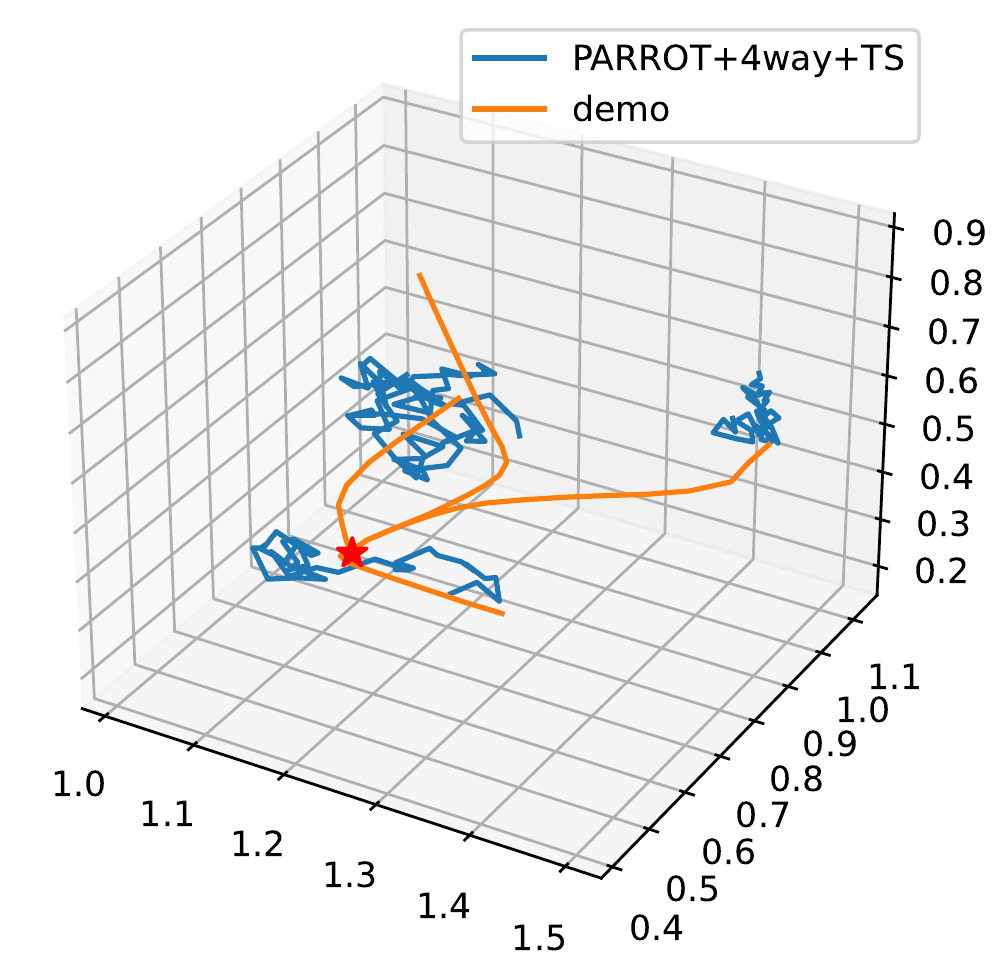}}
\subfigure[ours with 4 flows]{\includegraphics[width=\linewidth]{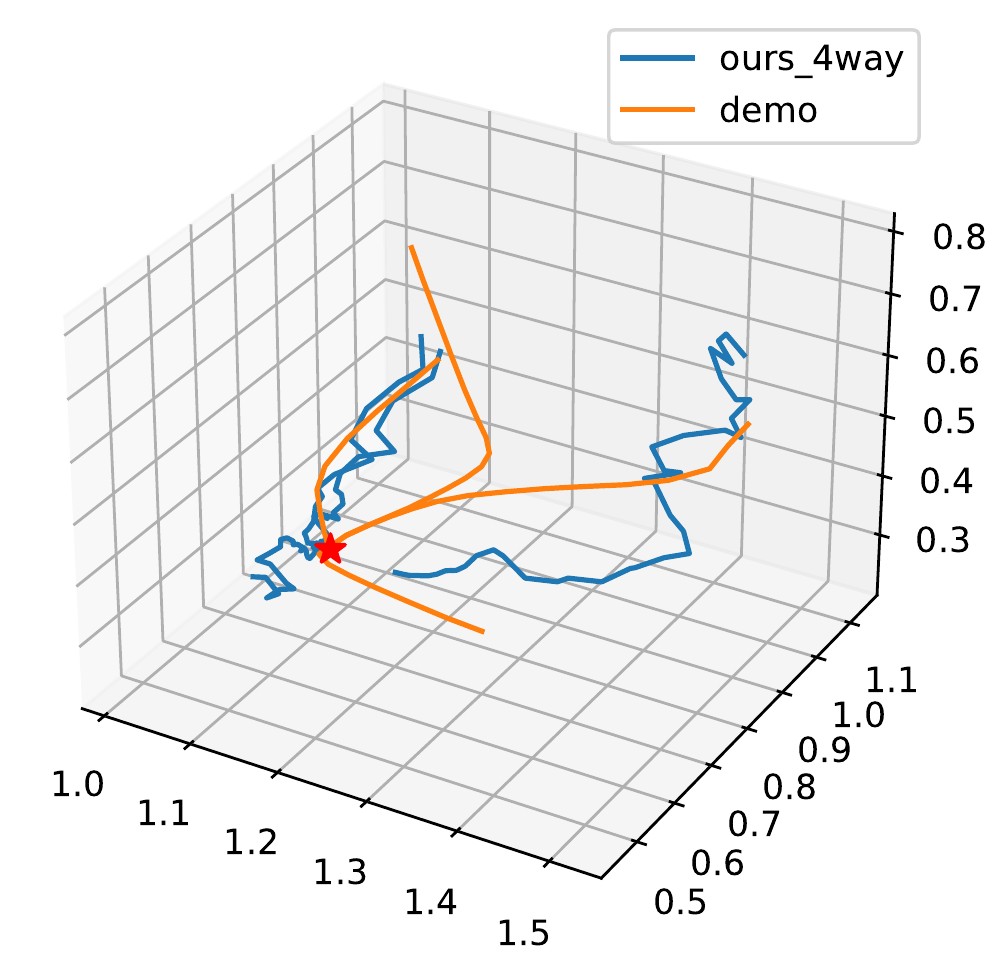}}

\end{minipage}
\begin{minipage}[c]{0.32\linewidth}

\subfigure[PARROT+2way+TS]{\includegraphics[width=\linewidth]{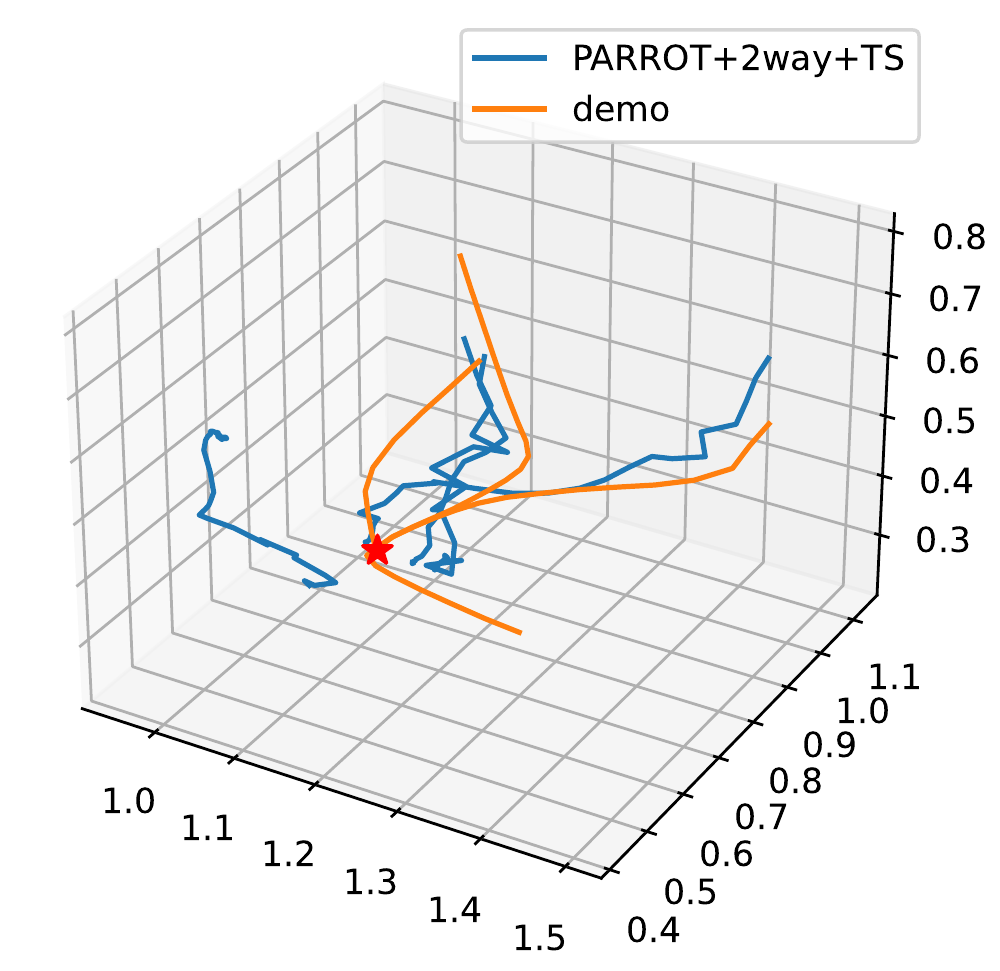}}
\subfigure[ours with 2 flows]{\includegraphics[width=\linewidth]{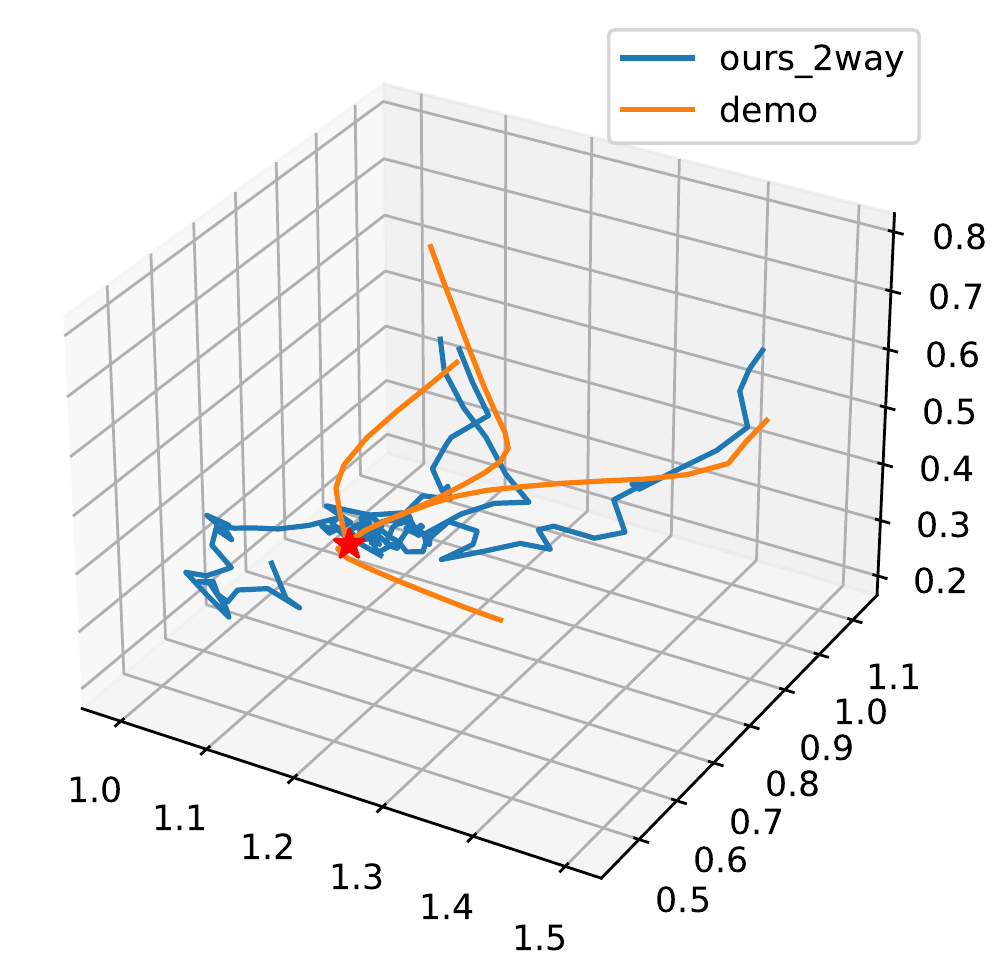}}
\end{minipage}
\begin{minipage}[c]{0.24\linewidth}
\end{minipage}
\caption{Illustration of trajectories generated by our method and PARROT under the direction 4.5 setting in the fetchreach environment  \textbf{without any RL training}. Both methods do not use the explicit prior or the push-forward technique. Our method does not use the task-specific single flow. For PARROT, 2/4way means the two/four most related directions in the task-agnostic dataset (i.e., directions 4, 5 / 3, 4, 5, 6). For our method, 2/4 flows are trained on the two/four most related directions in the task-agnostic dataset. The orange line is the task-specific dataset for reference. All orange lines converge at the red star, which is the goal.}
\label{fig:traj}
\end{figure}
\endgroup
}

Fig.~\ref{fig:traj_others} illustrates the trajectories generated by each method after RL training. As shown in the figure, our method exhibits smoother trajectories after RL training, enabling the agent to reach its goal faster. FIST, SKiLD, and na\"ive RL fail to generate trajectories that steadily converge to the goal. Although PARROT+(TS+TA) (PARROT with both task-agnostic and task-specific datasets) struggles at the beginning of RL, the prior enables the agent to reach the goal occasionally. Because of this, it  learns to rule out other infeasible directions. PARROT+TA fails to reach the goal when the starting location is too far, as it has no idea about how to reach the goal. 

{
\begingroup
\begin{figure}[t]
\begin{minipage}[c]{0.32\linewidth}
\subfigure[ours]{\includegraphics[width=\linewidth]{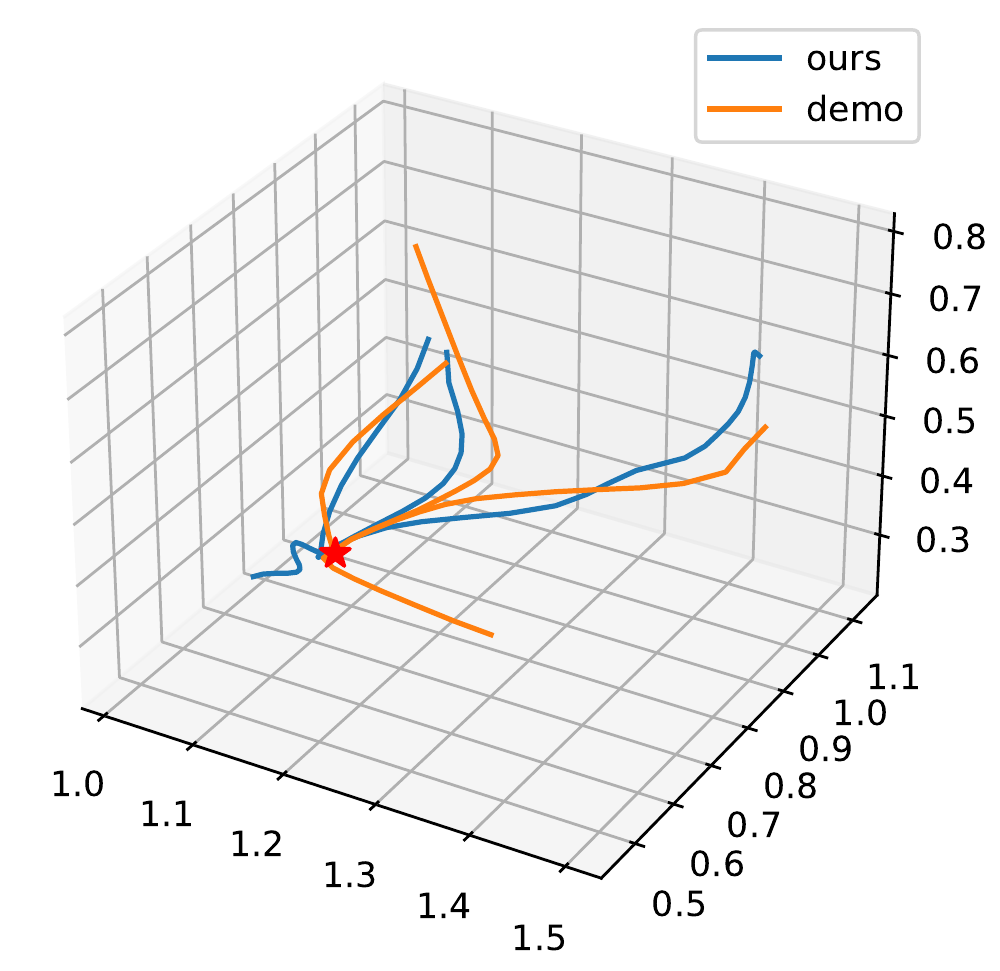}}
\subfigure[PARROT+(TS+TA)]{\includegraphics[width=\linewidth]{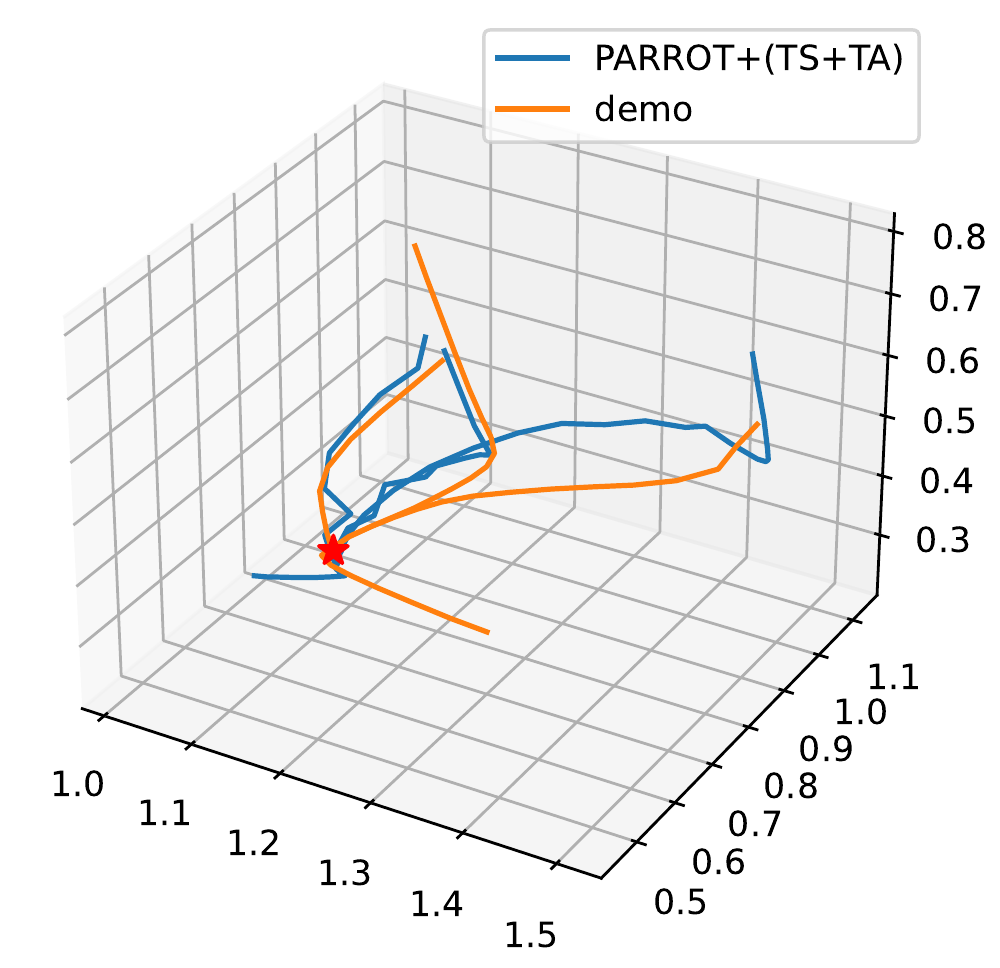}}
\vspace{146pt}
\end{minipage}
\begin{minipage}[c]{0.32\linewidth}
\subfigure[SKiLD]{\includegraphics[width=\linewidth]{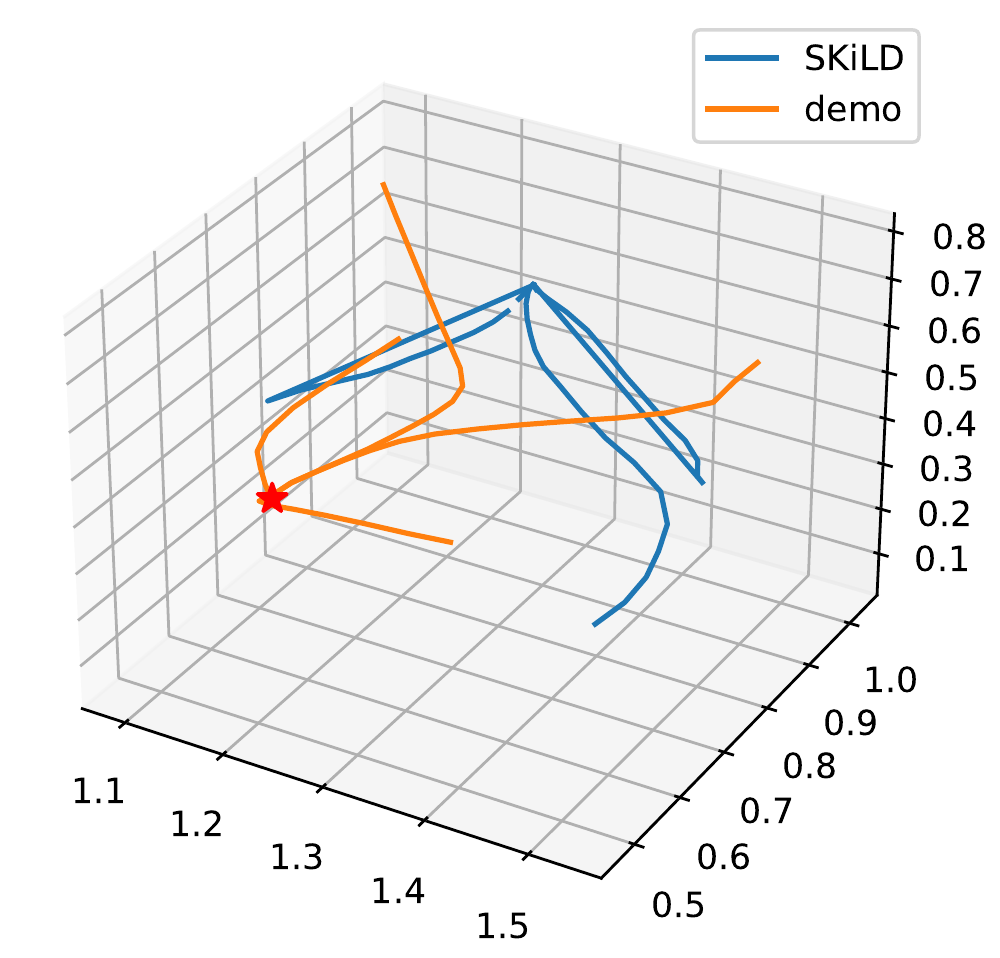}}
\subfigure[PARROT+TA]{\includegraphics[width=\linewidth]{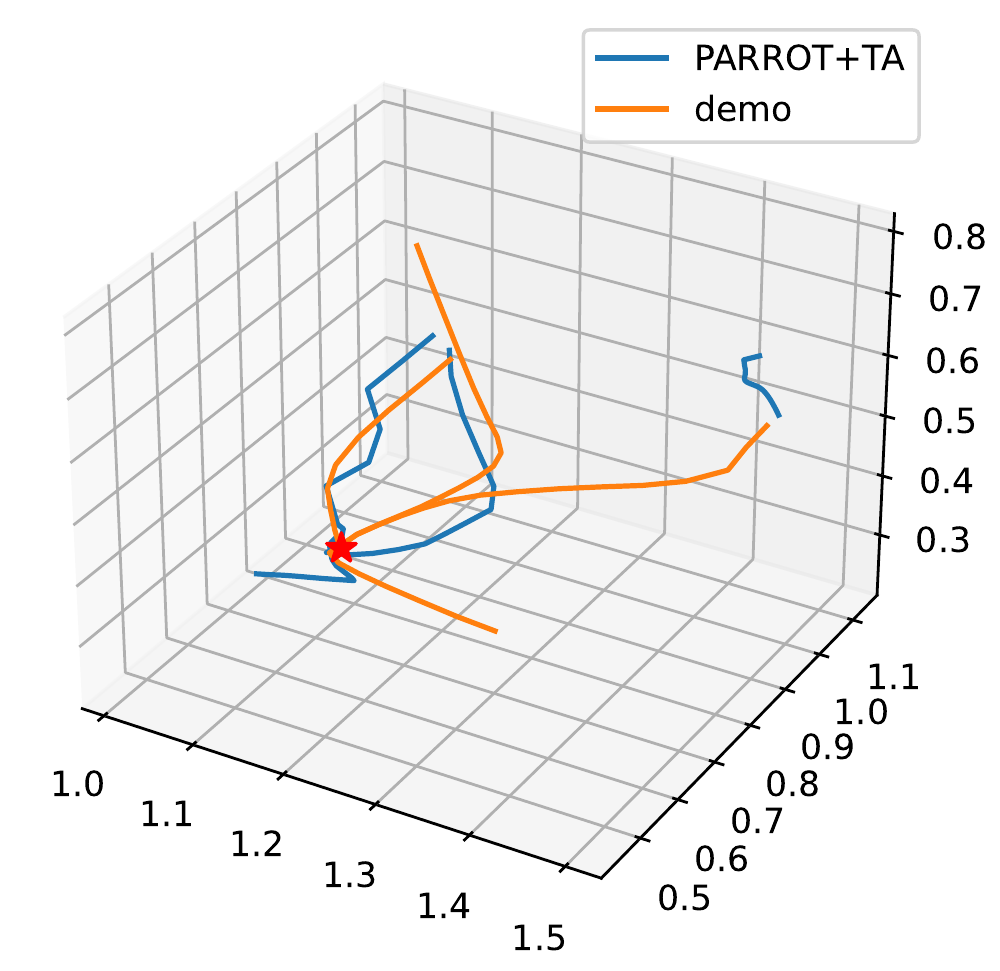}}
\subfigure[naive]{\includegraphics[width=\linewidth]{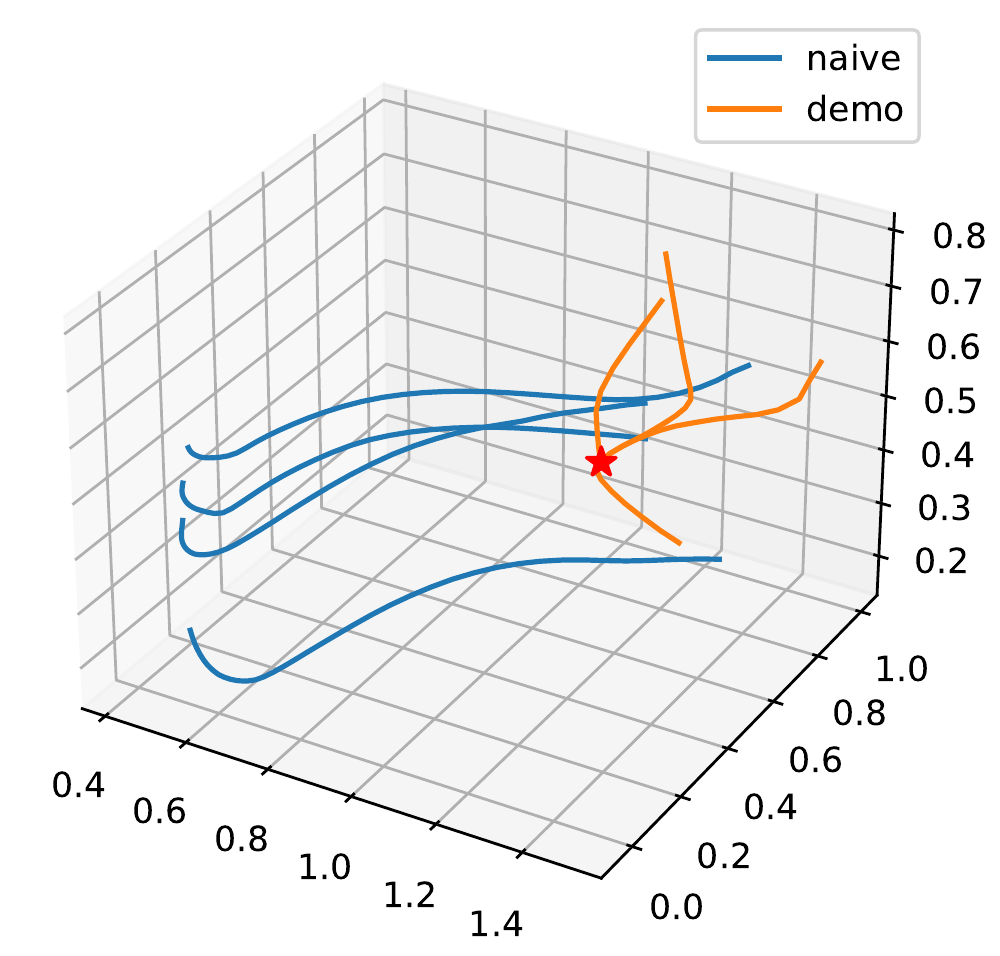}}
\end{minipage}
\begin{minipage}[c]{0.32\linewidth}
\subfigure[FIST]{\includegraphics[width=\linewidth]{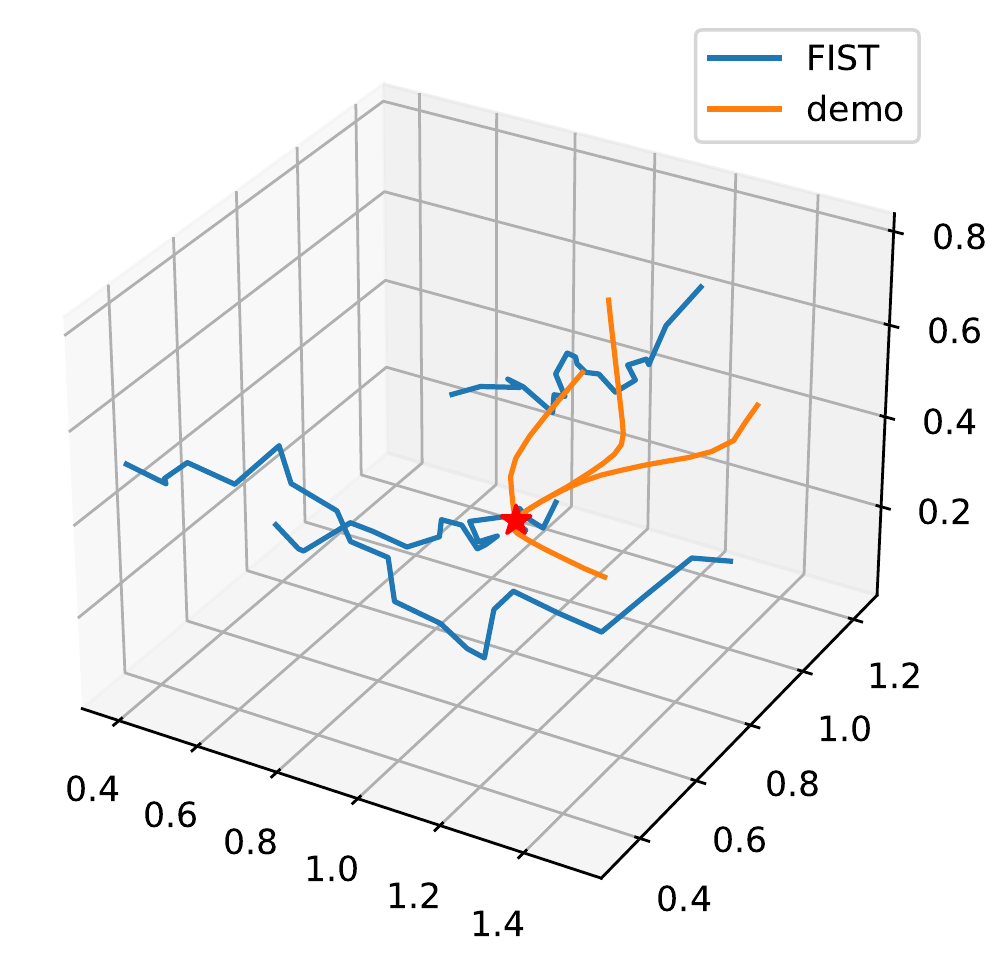}}
\subfigure[PARROT+TS]{\includegraphics[width=\linewidth]{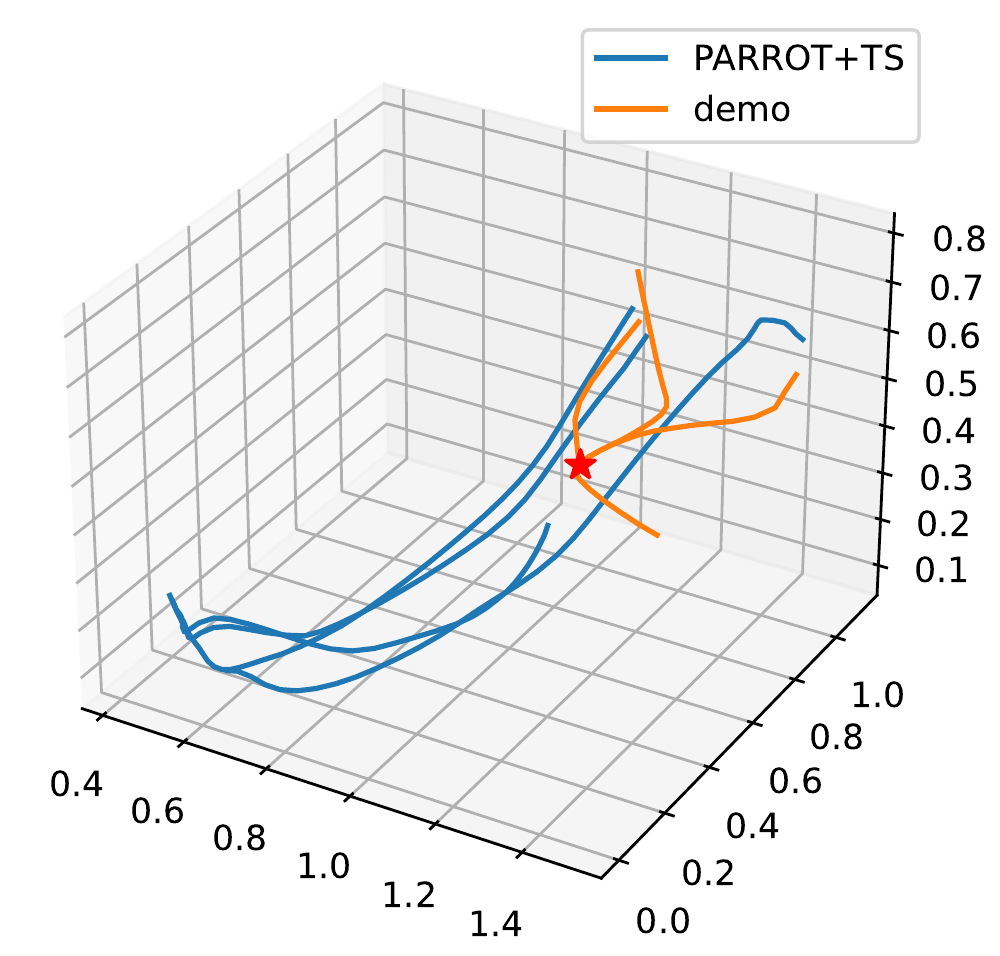}}
\vspace{146pt}
\end{minipage}
\begin{minipage}[c]{0.24\linewidth}
\end{minipage}
\caption{Illustration of trajectories generated by all methods \textbf{after RL training}; similar to Fig.~\ref{fig:traj}, the blue curves are the trajectories, the orange curves are the demonstrations, and the red star is the goal. Our method and PARROT do not use the explicit prior or the push-forward technique. Our method does not use the task-specific single flow.}
\label{fig:traj_others}
\end{figure}
\endgroup
}

\subsection{Kitchen and Office}
\label{sec:expKT}

{

\begingroup
\begin{figure}[t]
\begin{minipage}[c]{0.32\linewidth}
\subfigure[Kitchen-SKiLD-A]{\includegraphics[width=\linewidth]{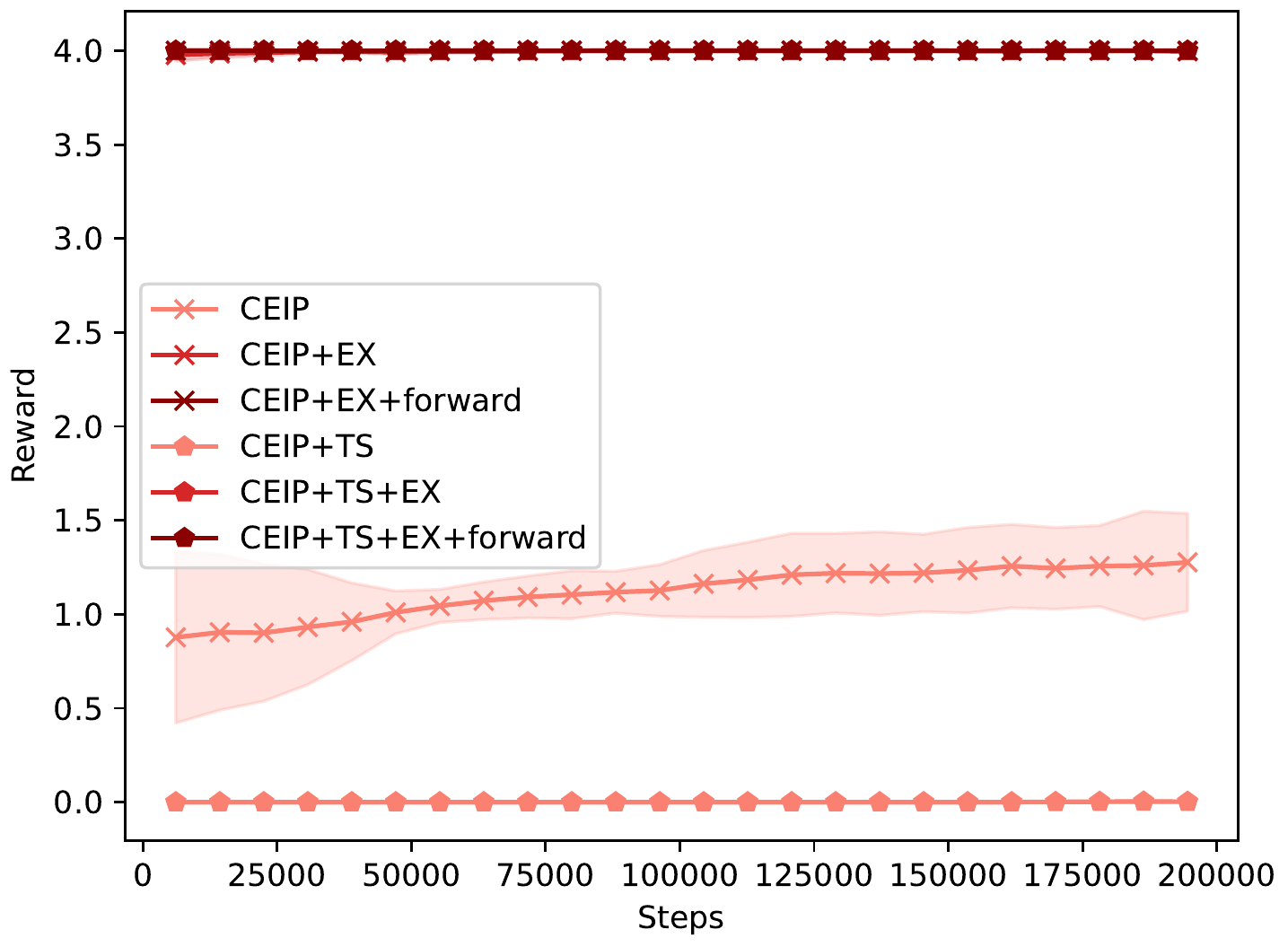}}
\subfigure[Kitchen-FIST-B]{\includegraphics[width=\linewidth]{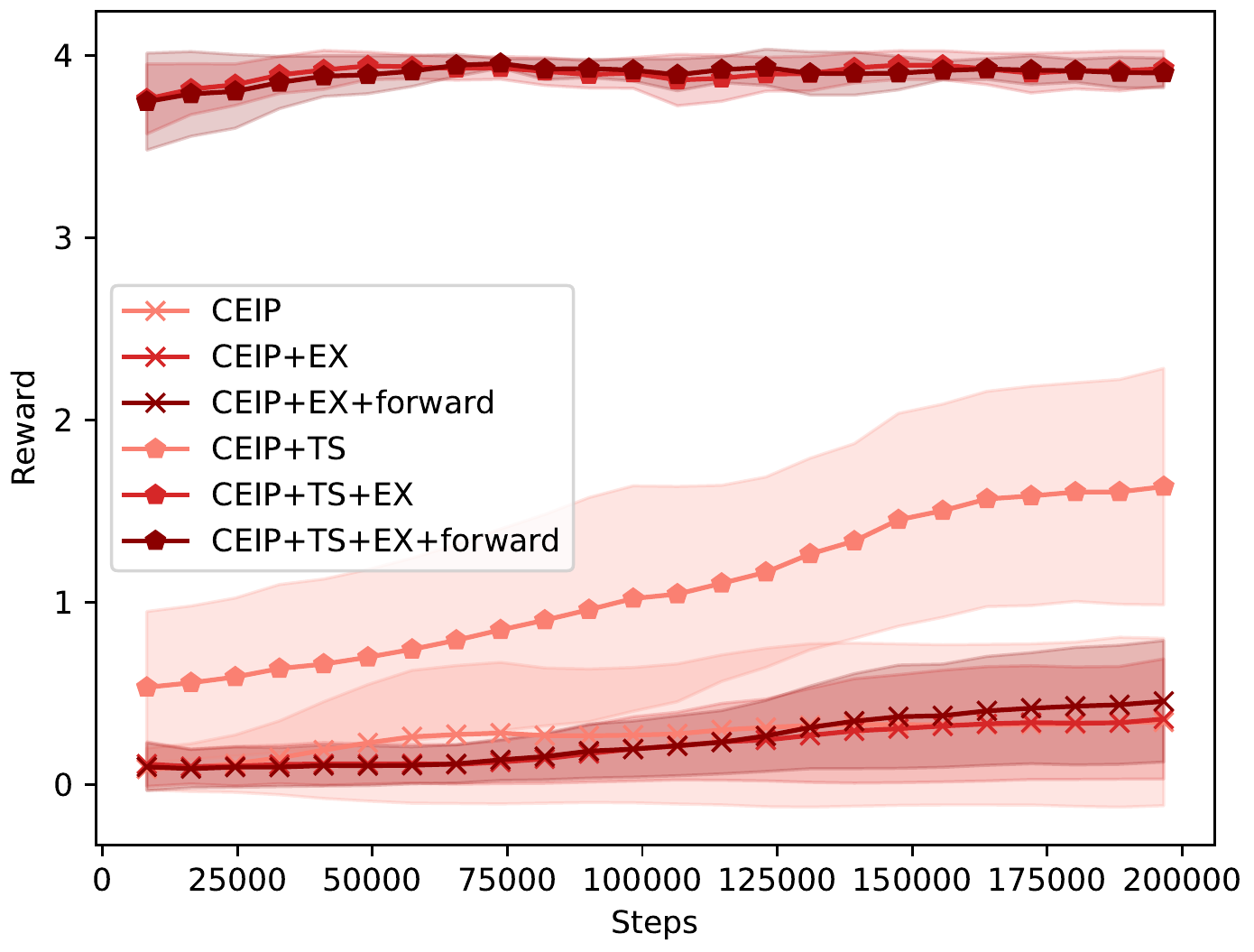}}
\end{minipage}
\begin{minipage}[c]{0.32\linewidth}
\subfigure[Kitchen-SKiLD-B]{\includegraphics[width=\linewidth]{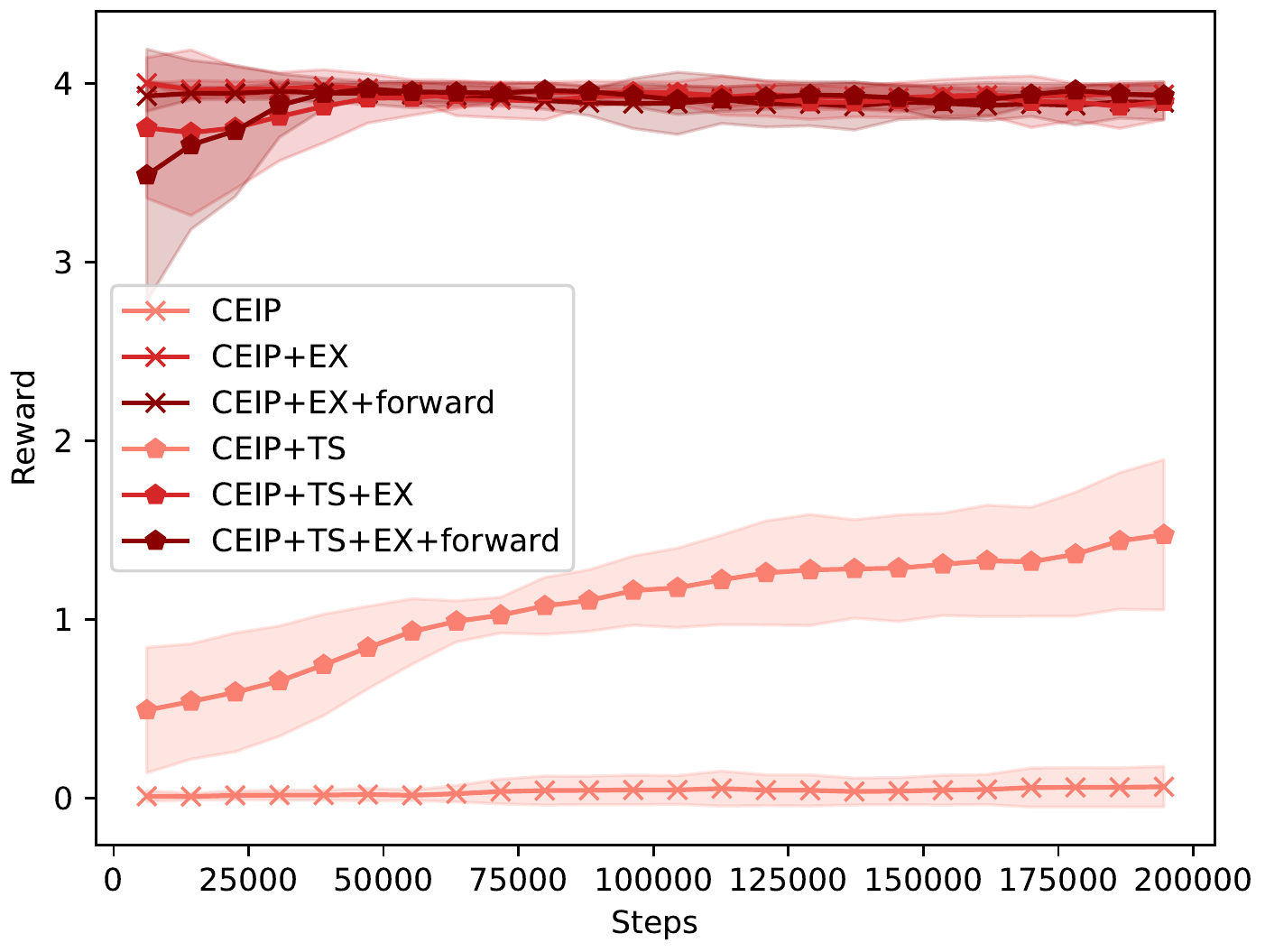}}
\subfigure[Kitchen-FIST-C]{\includegraphics[width=\linewidth]{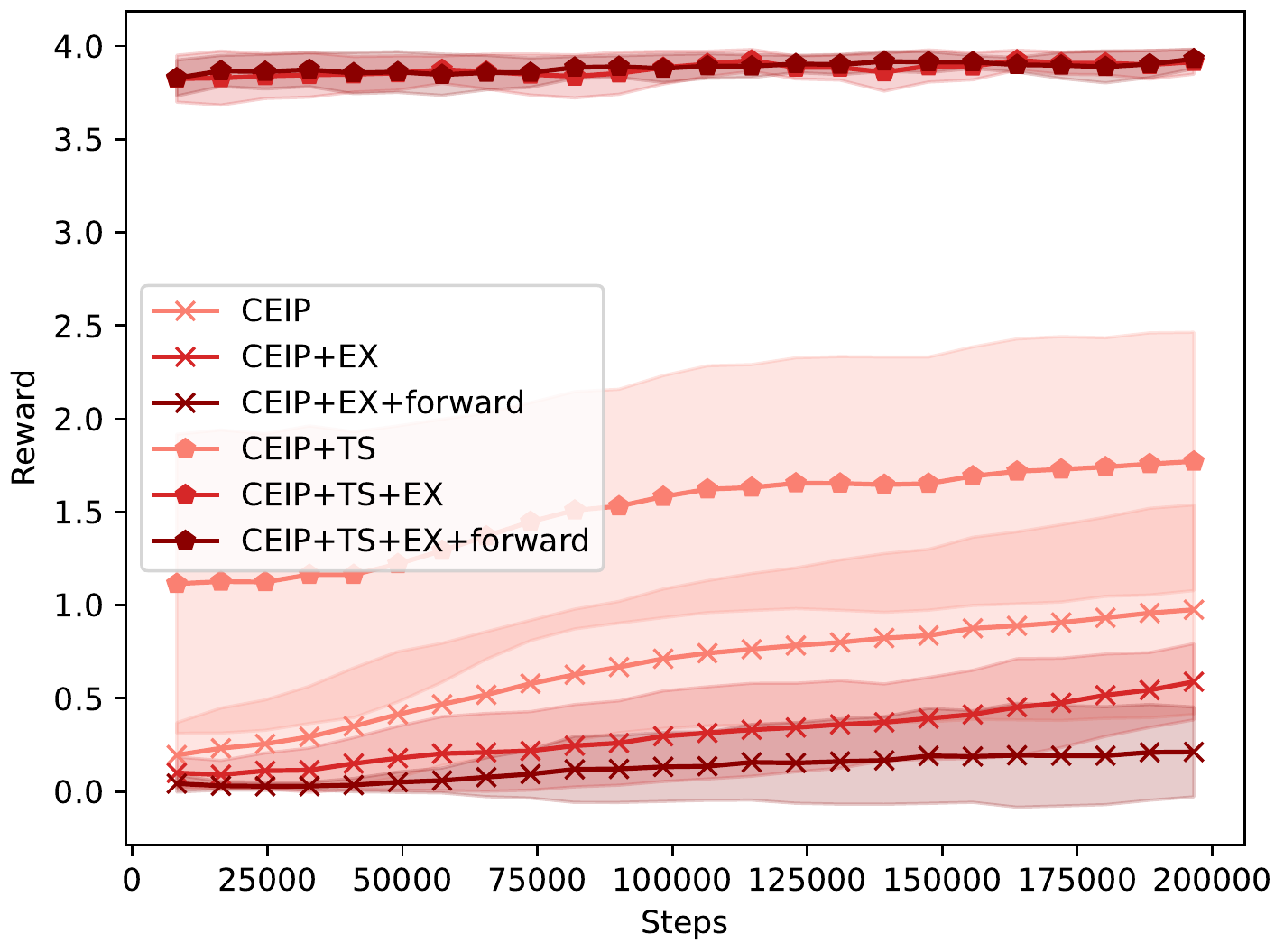}}
\end{minipage}
\begin{minipage}[c]{0.32\linewidth}
\subfigure[Kitchen-FIST-A]{\includegraphics[width=\linewidth]{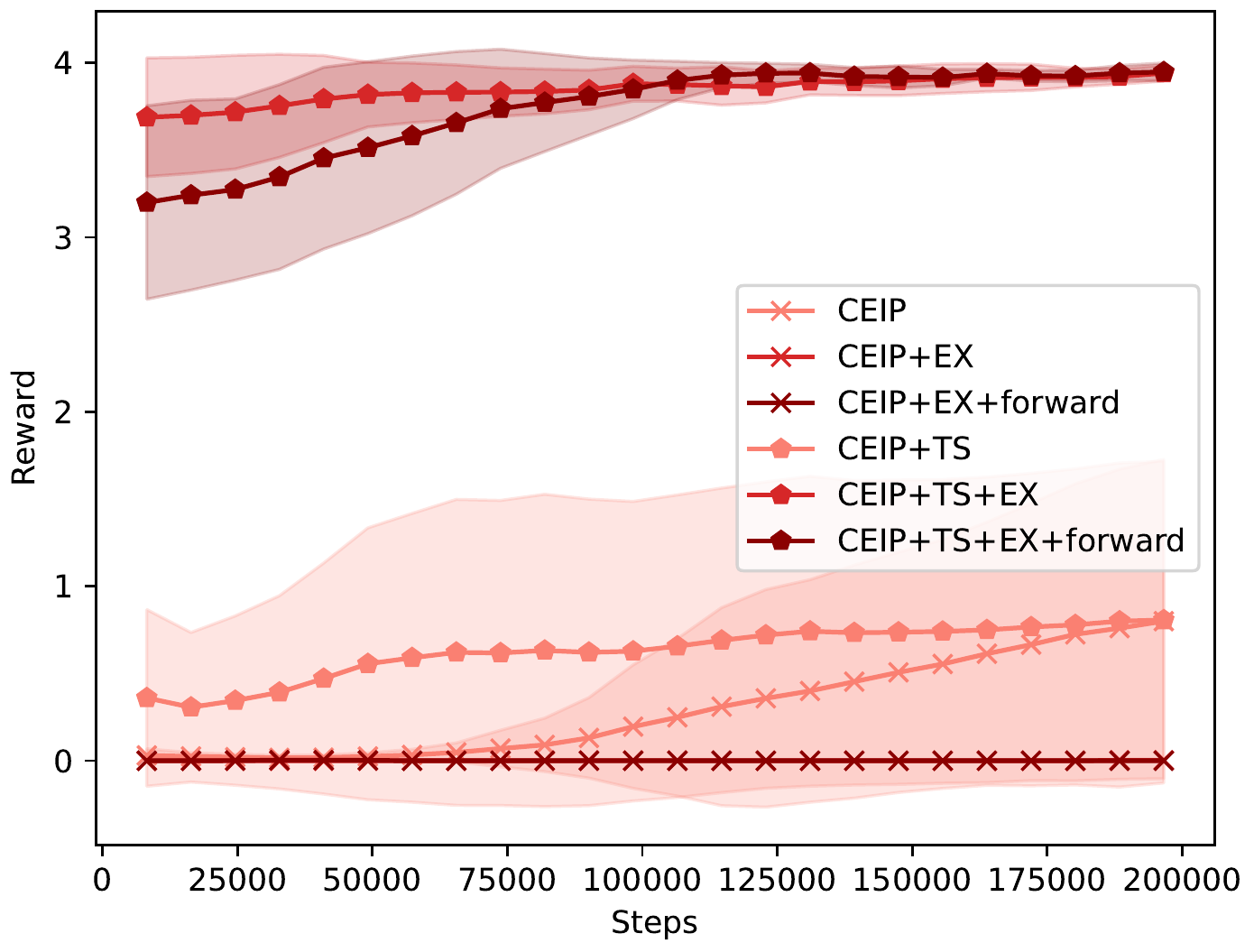}}
\centering
\subfigure[Kitchen-FIST-D]{\includegraphics[width=\linewidth]{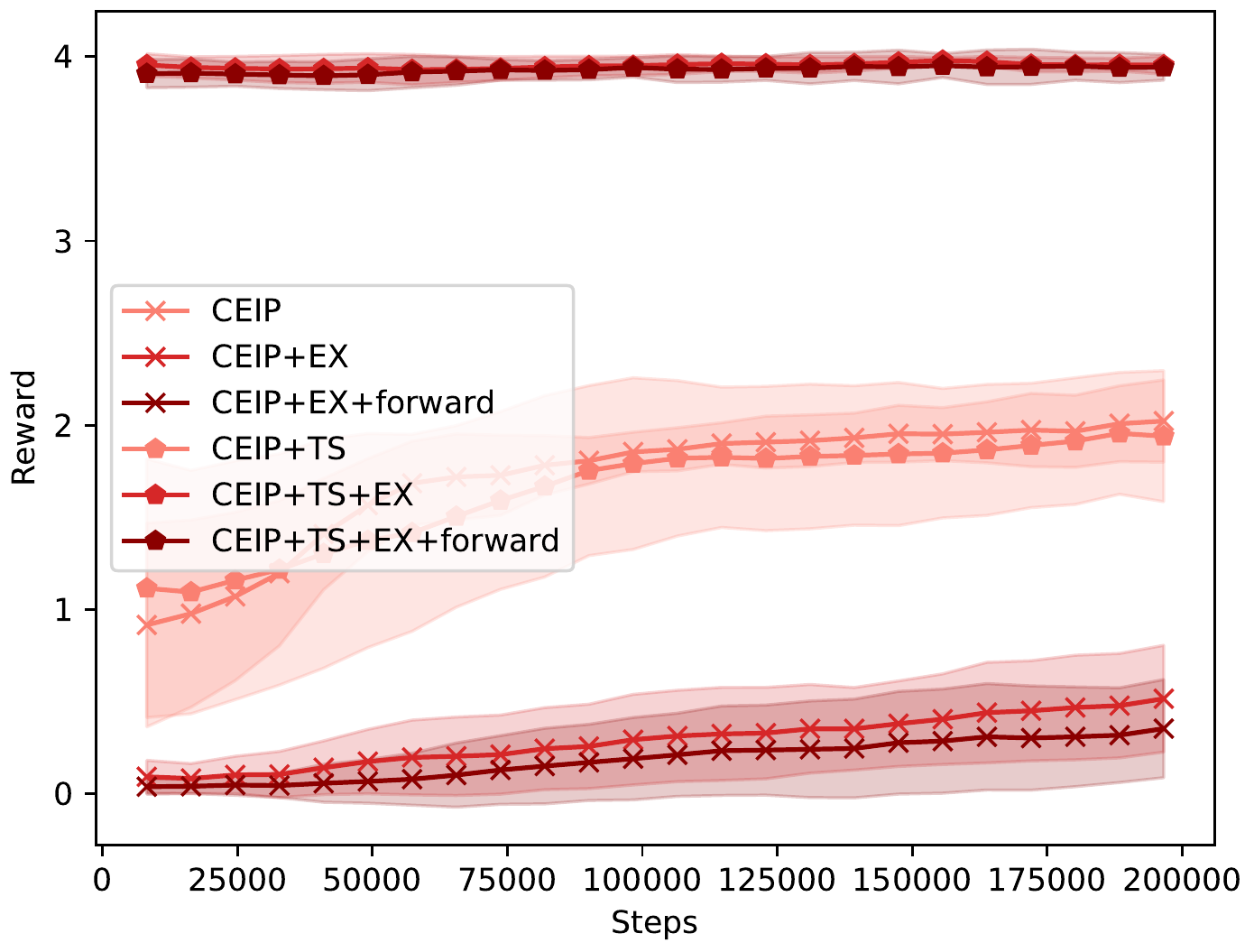}}
\end{minipage}
\begin{minipage}[c]{0.24\linewidth}
\end{minipage}
\caption{Ablation on the components of our method in the kitchen environment. For both environments, the presence of an explicit prior greatly enhances the results; for kitchen-FIST where part of the target task disappears from the task-agnostic dataset, a task-specific flow is also very important.}
\label{fig:kitchen_ablation_ours}
\end{figure}
\endgroup
}

\textbf{Ablation on components of CEIP.} Fig.~\ref{fig:kitchen_ablation_ours} shows the difference of performance using different architectures for our method. We observe that the explicit prior plays a crucial role in both Kitchen-SKiLD and Kitchen-FIST. Also, for Kitchen-FIST, where one of the target sub-tasks is only part of the task-specific data, the presence of the task-specific single flow $f_{n+1}$ is also crucial for  success. We do not find the push-forward technique to help much in this setting. 

{
\begingroup
\begin{figure}[t]
\begin{minipage}[c]{0.32\linewidth}
\subfigure[Kitchen-SKiLD-A]{\includegraphics[width=\linewidth]{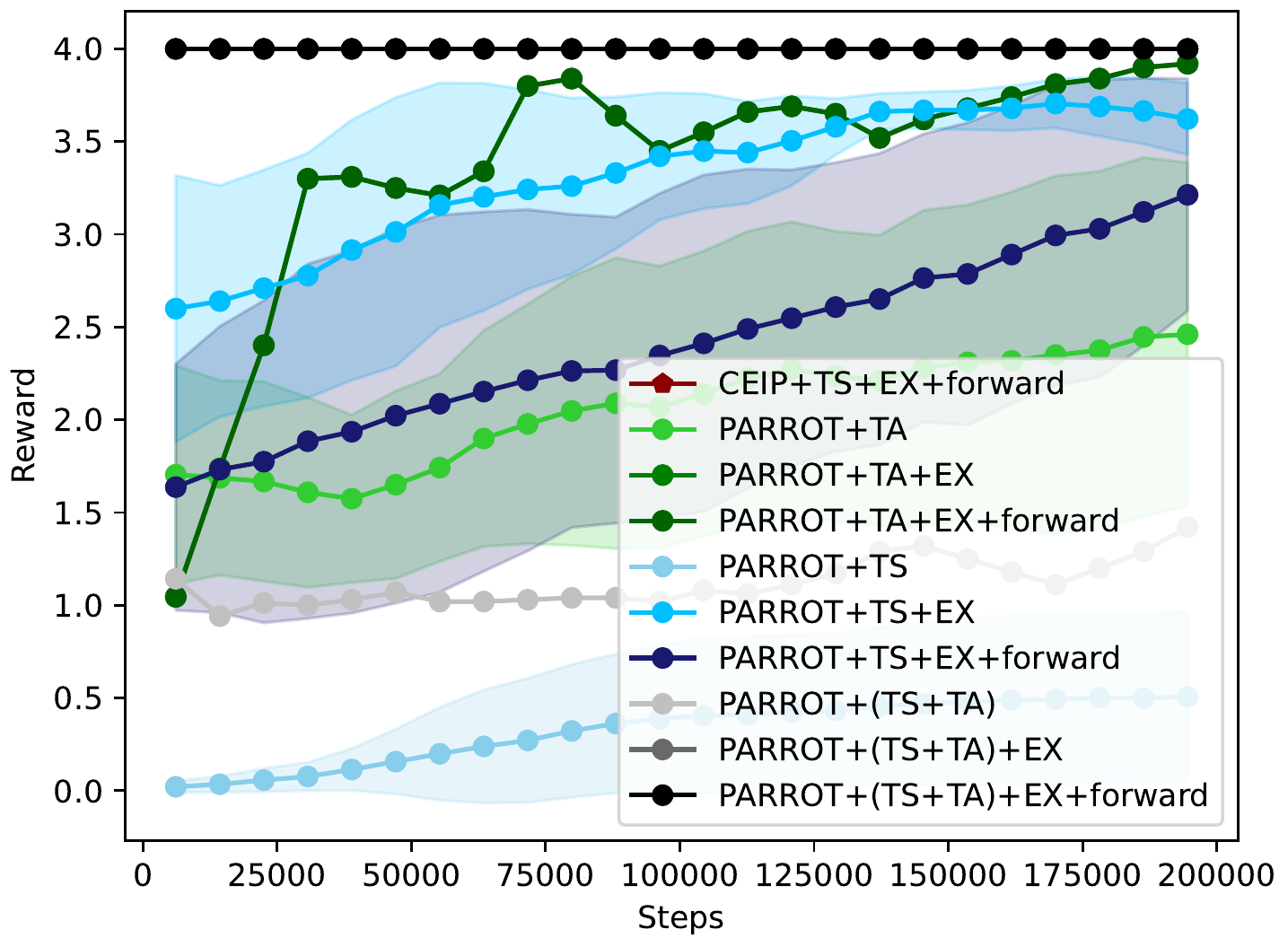}}

\subfigure[Kitchen-FIST-B]{\includegraphics[width=\linewidth]{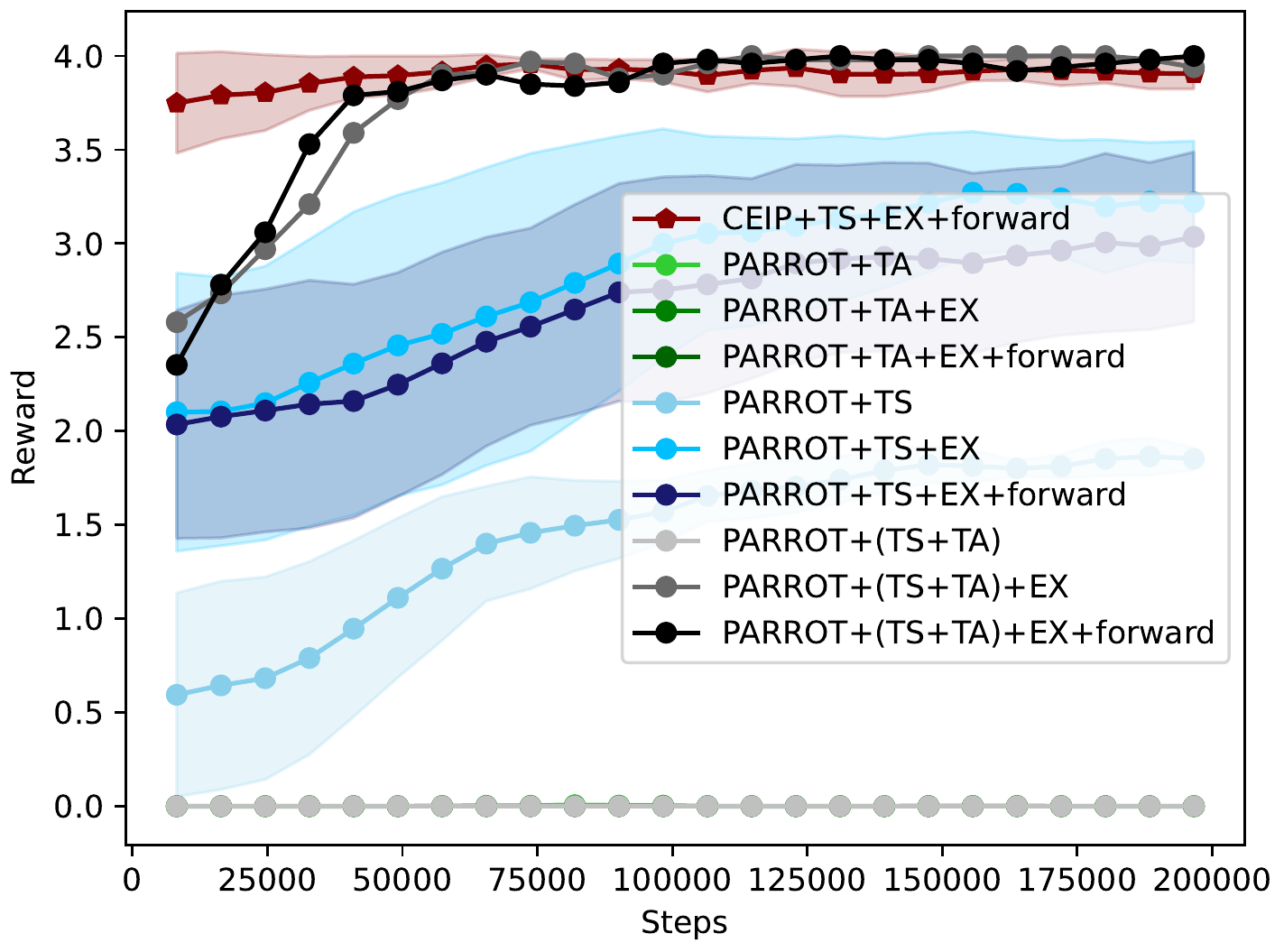}}

\end{minipage}
\begin{minipage}[c]{0.32\linewidth}
\subfigure[Kitchen-SKiLD-B]{\includegraphics[width=\linewidth]{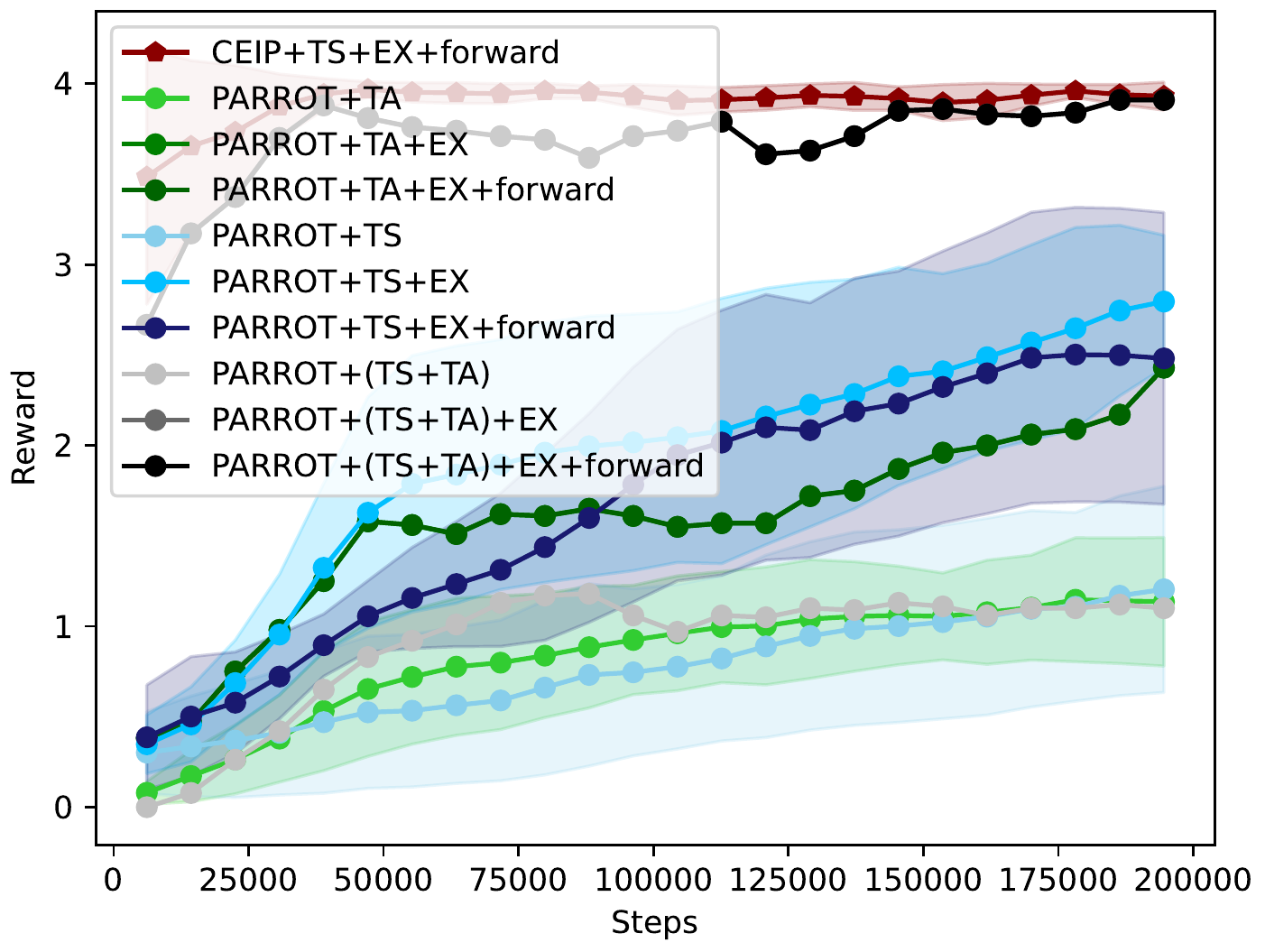}}

\subfigure[Kitchen-FIST-C]{\includegraphics[width=\linewidth]{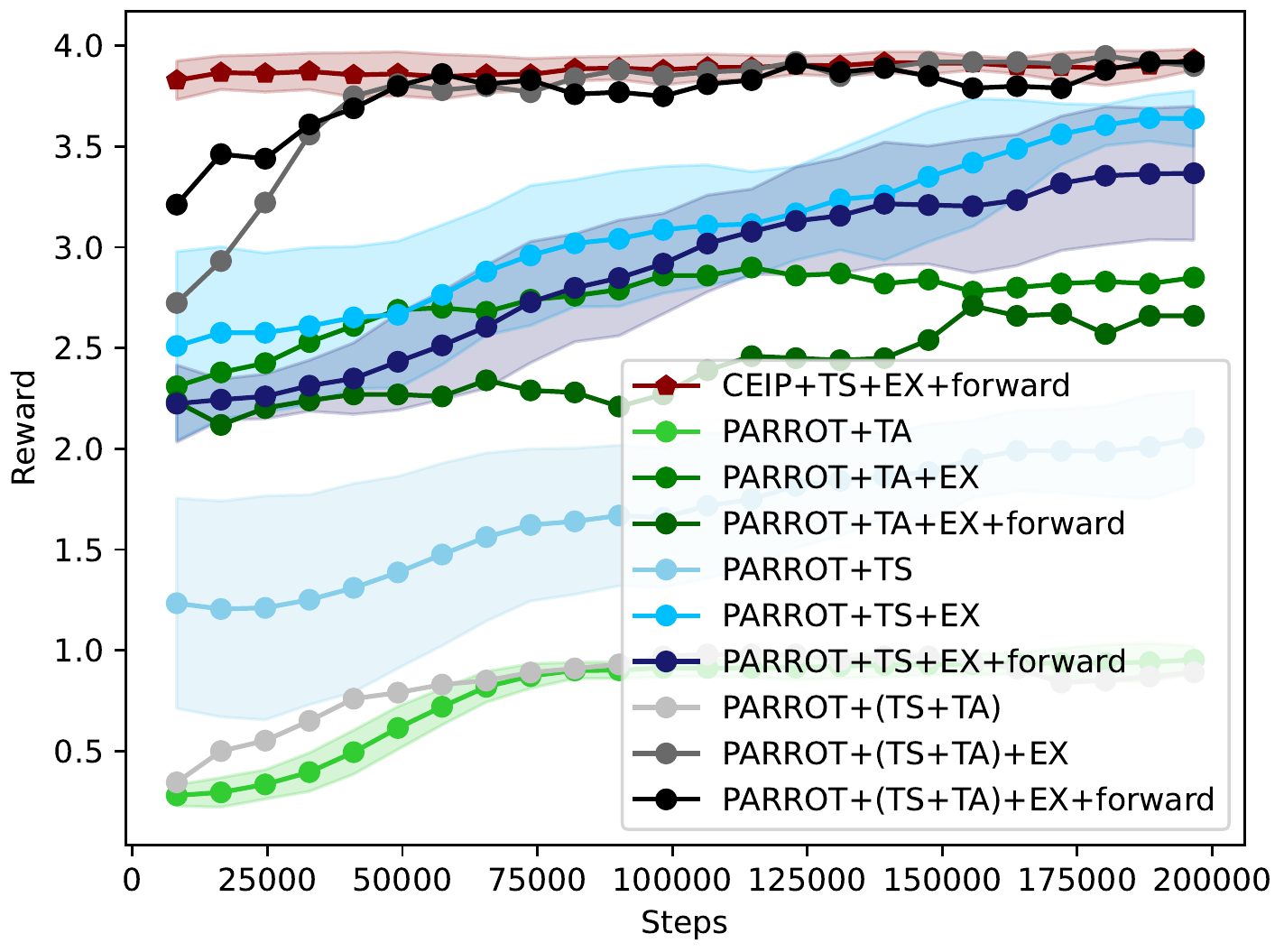}}

\end{minipage}
\begin{minipage}[c]{0.32\linewidth}
\subfigure[Kitchen-FIST-A]{\includegraphics[width=\linewidth]{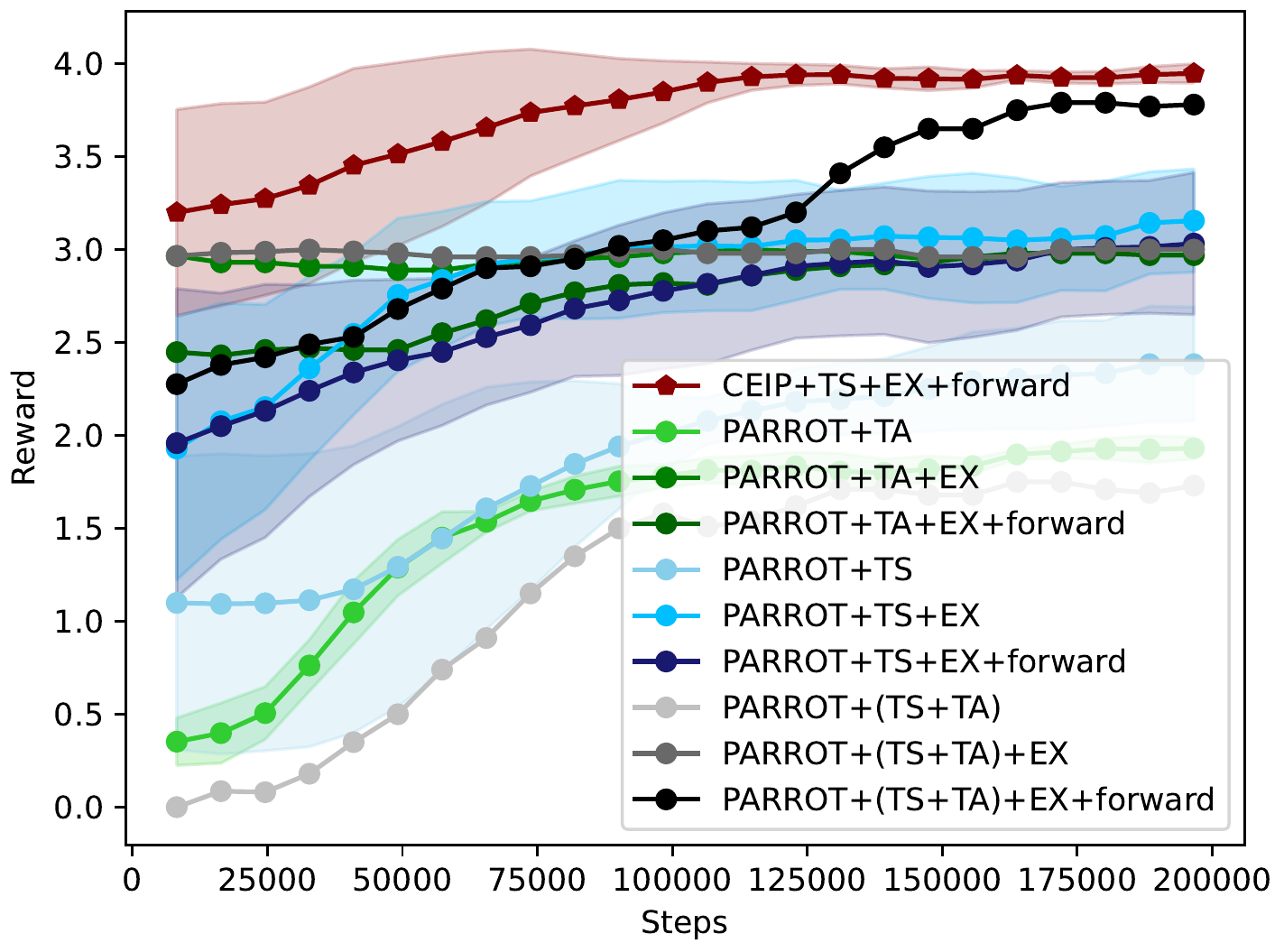}}
\centering
\subfigure[Kitchen-FIST-D]{\includegraphics[width=\linewidth]{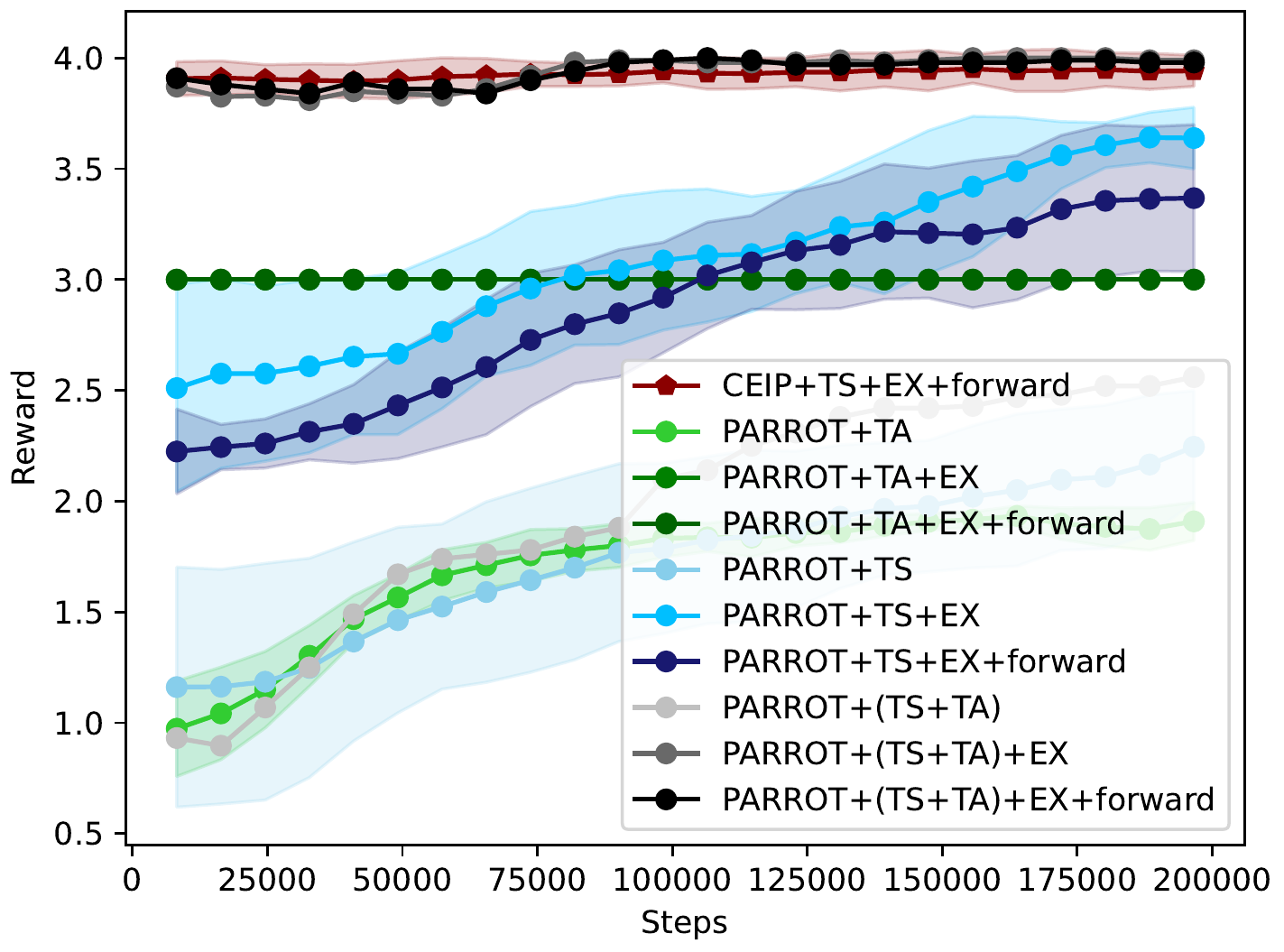}}

\end{minipage}
\caption{Ablation results for PARROT in the kitchen environment. For convenience, we also list CEIP+TS+EX+forward for reference. Note, CEIP+TS+EX+forward outperforms all variants of PARROT. Similar to CEIP, PARROT can be improved by using an explicit prior. Note, in Kitchen-FIST-B, PARROT+TA cannot learn anything, because the very first subtask in the target task sequence is missing in the task-agnostic dataset. It can only learn all subtasks before the missing subtask.}
\label{fig:kitchen_ablation_PARROT}
\end{figure}
\endgroup
}

\textbf{Ablation on components in PARROT.} Fig.~\ref{fig:kitchen_ablation_PARROT} shows the difference when different architectures are used. As one target sub-task is completely missing from the task-agnostic data, PARROT+TA fails as expected. Also note that the explicit prior  boosts the results of PARROT, making it comparable to our method if given enough training time.

\textbf{Ablation on the effect of using ground-truth labels.} Table~\ref{tab:GT1} and Table~\ref{tab:GT2} show the performance comparison between using ground-truth labels and labels acquired by $k$-means in the kitchen environment.\footnote{The office environment has 210 ground-truth labels, which is hard to train.} As we are using $24$ labels in the main paper, but not all the task-agnostic datasets have 24 ground-truth labels, we also show the result using ground-truth pruned to 24 labels for a fair comparison. For Kitchen-SKiLD where the number of ground-truth labels is 33, there are exactly 9 labels that have no more than 3 demonstrations. We merge each of them into the label that is next to them in the dictionary order of concatenated task names. For kitchen-FIST where the number of ground-truth labels is $x$, $x<24$, we select the $24-x$ labels with the most demonstrations, and divide them evenly into two halves; each half is a new label. Note, no task information is taken into account when merging.

For readability, we use some suffixes in Table~\ref{tab:GT1} and Table~\ref{tab:GT2} to differentiate variants of CEIP in the ``label'' column. The meaning of the suffixes are as follows:

\begin{itemize}
\item\textbf{GT24}: Ground-truth labels, but  merged or split to form 24 labels;
\item\textbf{GT}: Ground-truth labels; the number of subtasks differs;
\item\textbf{KM}: K-means labels.
\end{itemize}



\begin{table}[t]
\setlength{\tabcolsep}{3pt}
    \centering{
    \begin{tabular}{cccccccc}\toprule
         Label & Method & \shortstack{Kitchen-\\SKiLD-A} & \shortstack{Kitchen-\\SKiLD-B} & \shortstack{Kitchen-\\FIST-A} & \shortstack{Kitchen-\\FIST-B} & \shortstack{Kitchen-\\FIST-C} & \shortstack{Kitchen-\\FIST-D} \\ \midrule
         GT & CEIP+TS+EX & \textbf{4} & 3.96 & 3.59 & \textbf{4} & 3.94 & 3.6 \\
         GT & CEIP+TS+EX+forward & \textbf{4} & 3.95 & 3 & \textbf{4} & 3.81 & 3.4 \\
         GT24 & CEIP+TS+EX & \textbf{4} & \textbf{4} & 3.68 & 3.76 & 3.8 & 3.96 \\
         GT24 & CEIP+TS+EX+forward & \textbf{4} & \textbf{4} & 3.24 & 3.75 & 3.85 & 3.9 \\
         KM & CEIP+TS+EX & \textbf{4} & 3.81 & 3.44 & 3.8 & \textbf{4} & 3.75 \\
         KM & CEIP+TS+EX+forward & \textbf{4} & 3.32 & 3.41 & \textbf{4} & 3.94 & 3.76 \\
         \bottomrule
    \end{tabular}}
    \caption{Ground-truth label and $k$-means label impact for CEIP+TS+EX and CEIP+TS+EX+forward before RL.}
    \label{tab:GT1}
\end{table}

\begin{table}[t]
\setlength{\tabcolsep}{3pt}
    \centering{
    \begin{tabular}{cccccccc}\toprule
         Label & Method & \shortstack{Kitchen-\\SKiLD-A} & \shortstack{Kitchen-\\SKiLD-B} & \shortstack{Kitchen-\\FIST-A} & \shortstack{Kitchen-\\FIST-B} & \shortstack{Kitchen-\\FIST-C} & \shortstack{Kitchen-\\FIST-D} \\ \midrule
         GT & CEIP+TS+EX & \textbf{4} & 3.87 & 3.93 & 3.8 & 3.94 & 3.71 \\
         GT & CEIP+TS+EX+forward & \textbf{4} & 3.87 & 3.9 & 3.74 & 3.96 & 3.93 \\
         GT24 & CEIP+TS+EX & \textbf{4} & \textbf{4} & 3.92 & 3.97 & 3.99 & 3.87 \\
         GT24 & CEIP+TS+EX+forward & \textbf{4} & \textbf{4} & 3.99 & 3.88 & 3.95 & 3.96 \\
         KM & CEIP+TS+EX & \textbf{4} & \textbf{4} & 3.94 & 3.92 & 3.93 & 3.95 \\
         KM & CEIP+TS+EX+forward & \textbf{4} & \textbf{4} & 3.95 & 3.89 & 3.92 & 3.94 \\
         \bottomrule
    \end{tabular}}
    \caption{Ground-truth label and $k$-means label impact for CEIP+TS+EX and CEIP+TS+EX+forward after RL.}
    \label{tab:GT2}
\end{table}

The result suggests that for Kitchen-SKiLD, ground truth (both 24 flows and 33 flows) helps as CEIP with ground-truth labels works better than CEIP with $k$-means label (Table~\ref{tab:GT1} shows higher reward). For Kitchen-FIST, the reward is similar before and after RL training, and the precise label does not matter.

\textbf{Performance of behavior cloning and replaying demonstrations.} We test behavior cloning and replaying demonstrations (which is duplicating actions regardless of current state) on the kitchen and office environment to see if the task-specific dataset already provides the optimal solution for our testbeds. Table~\ref{tab:BCRP} shows the result of vanilla behavior cloning (BC), behavior cloning with explicit prior (BC+EX), with explicit prior and push-forward (BC+EX+forward), and replaying demonstrations (replay). The result shows that: 1) behavior cloning and replay are very brittle, and cannot directly solve our testbed; 2) an explicit prior significantly improves the performance of behavior cloning, which proves the validity of our design.

\begin{table}[htbp]
\setlength{\tabcolsep}{2pt}
    \centering
    {
    \begin{tabular}{cccccc}\toprule
         Environment & BC & BC+EX & BC+EX+forward & Replay & CEIP+TS+EX+forward\\ \midrule 
         Kitchen-SKiLD-A & $0.02${\scriptsize $\pm 0.04$} & $1.52${\scriptsize $\pm 1.15$} & $2.2${\scriptsize $\pm 0.62$} & $1.0${\scriptsize $\pm 0.82$}  & $\mathbf{4.0}${\scriptsize $\pm 0.00$} \\
         Kitchen-SKiLD-B & $0.03${\scriptsize$\pm 0.08$} & $1.03${\scriptsize$\pm 0.90$} & $0.8${\scriptsize$\pm 0.75$} & $0.67${\scriptsize$\pm 0.94$} & $\mathbf{3.93}${\scriptsize$\pm 0.08$}\\
         Kitchen-FIST-A & $0.67${\scriptsize$\pm 0.76$} & $2.17${\scriptsize$\pm 0.06$} & $3.03${\scriptsize$\pm 0.15$} & $2.33${\scriptsize$\pm 0.47$} & $\mathbf{3.95}${\scriptsize$\pm 0.05$}\\
         Kitchen-FIST-B & $0.4${\scriptsize$\pm 0.59$} & $2.13${\scriptsize$\pm 0.47$} & $1.87${\scriptsize$\pm 0.29$} & $0.67${\scriptsize$\pm 0.47$} & $\mathbf{3.89}${\scriptsize$\pm 0.07$}\\
         Kitchen-FIST-C & $0.5${\scriptsize$\pm 0.75$} & $2.2${\scriptsize$\pm 1.61$} & $1.9${\scriptsize$\pm 0.96$} & $2.33${\scriptsize$\pm 0.94$} & $\mathbf{3.92}${\scriptsize$\pm 0.06$}\\
         Kitchen-FIST-D & $0.67${\scriptsize$\pm 0.39$} & $1.63${\scriptsize$\pm 1.42$} & $2.17${\scriptsize$\pm 1.67$} & $2.33${\scriptsize$\pm 0.94$} & $\mathbf{3.94}${\scriptsize$\pm 0.07$} \\
         Office & $0.62${\scriptsize$\pm 0.59$} & $0.53${\scriptsize$\pm 0.42$} & $1.83${\scriptsize$\pm 0.49$} & $4.67${\scriptsize$\pm 0.83$} & $\mathbf{6.33}${\scriptsize$\pm 0.30$} \\
         \bottomrule
    \end{tabular}
    }
    \caption{Performance of behavior cloning and replaying demonstrations. For convenience, we also list CEIP+TS+EX+forward for reference.}
    \label{tab:BCRP}
\end{table}

\textbf{Robustness of CEIP with respect to the precision of task-specific demonstrations.} We test the robustness of CEIP and FIST with imprecise task-specific demonstrations in the office environment. The original office environment uses a $[-0.01, 0.01]$ uniformly random noise for the starting position of each dimension for each item in the environment. We increase this noise at test time (which the agent never sees in imitation learning) and summarize the result in Table~\ref{tab:robust}. The result shows that albeit an improvement upon FIST, CEIP is still not robust to imprecise demonstrations, which is a limitation that we discussed in the limitation section. 

\begin{table}[htbp]
    \centering{
    \begin{tabular}{cccc}\toprule
        Noise level & CEIP+TS+EX & CEIP+TS+EX+forward & FIST\\ \midrule
         $0.01$ (original) & $4.17$ & $\mathbf{6.33}$ & $5.6$ \\
         $0.02$ & $\mathbf{4.20}$ & $4.17$ & $3.8$ \\
         $0.05$ & $0.57$ & $\mathbf{0.83}$ & $0.6$ \\
         $0.1$ & $0.05$ & $\mathbf{0.1}$ & $0$ \\
         $0.2$ & $0.01$ & $\mathbf{0.02}$ & $0$ \\
         \bottomrule
    \end{tabular}}
    \caption{Comparison of the reward for CEIP and FIST when noise increases.}
    \label{tab:robust}
\end{table}



\section{Computational Resource Usage}
\label{sec:comp_resource}
All experiments are conducted on an Ubuntu 18.04 server with $72$ Intel Xeon Gold 6254 CPUs @ 3.10GHz, with a single NVIDIA RTX 2080Ti GPU. Under such settings, our method and PARROT+TA require around 
$1.5-3.5$ hours for training the implicit prior in the kitchen and office environments, depending on early stopping. FIST requires around $40$ minutes for prior training and $5$ minutes for the other parts. SKiLD requires around $9$ hours for prior training, $6-7$ hours for posterior training, and $6-7$ hours for discriminator training. PARROT+TS only needs a few minutes. As for reinforcement learning / deployment, our method on kitchen needs on average $10$ minutes for each run on fetchreach, and less than $2$ hours for the kitchen environment. For the office environment, we reach a speed of $12$ steps per second (including updates); SKiLD and PARROT can reach a speed of $20$ steps per second; FIST can reach a speed of $25-30$ steps per second as it has no RL updates.

\section{Dataset and Algorithm Licenses}
\label{sec:license}

We developed our code on the basis of multiple environment testbeds and algorithm repositories. 

\textbf{Environment testbeds.} We adopt fetchreach using the gym package from OpenAI, which has an MIT license. For the kitchen environment, we are using a forked version of the d4rl package which has an Apache-2.0 license. For the office environment, we are using a forked version of the roboverse, which has an MIT license.

\textbf{Algorithm repositories.} We implement PARROT from scratch as PARROT is not open-sourced. For SKiLD and FIST, we use their official github repositories. SKiLD has no license, but we have informed the authors and got their consent for using code academically. FIST has a BSD-3-clause license.

\end{document}